\documentclass{article}

\PassOptionsToPackage{numbers,sort&compress}{natbib}

\usepackage[preprint]{neurips_2026}

\usepackage{algorithm,algpseudocode}
\usepackage{amsmath,amssymb,amsthm,bm,mathtools}
\usepackage[english]{babel}
\usepackage{booktabs}
\usepackage[T1]{fontenc}
\usepackage{graphicx}
\usepackage[none]{hyphenat}
\usepackage[utf8]{inputenc}
\usepackage{nicefrac}
\usepackage{rotating}
\usepackage[subrefformat=parens]{subcaption}
\usepackage{tikz}
\usepackage[svgnames]{xcolor}

\usepackage[allcolors=Navy,colorlinks]{hyperref}
\usepackage[babel]{microtype}

\algnewcommand\algorithmicinput{\textbf{Input:}}
\algnewcommand\algorithmicoutput{\textbf{Output:}}
\algnewcommand\Input{\item[\algorithmicinput\ ]}
\algnewcommand\Output{\item[\algorithmicoutput\ ]}
\graphicspath{
  {./Figures/}
  {./Figures/Comparison}
  {./Figures/Downstream}
  {./Figures/Searching}
  {./Figures/Showcase}}

\title{FILLER: Feature Imputation via Latent Location Exploration and Retrieval}

\author{%
  Santu Mondal \\
  Machine Intelligence Unit \\ Indian Statistical Institute \\
  203 Barrackpore Trunk Road \\ Kolkata 700108 \\
  \texttt{santumondal002@gmail.com} \\
  \And
  Chayan Maitra \\
  Machine Intelligence Unit \\ Indian Statistical Institute \\
  203 Barrackpore Trunk Road \\ Kolkata 700108 \\
  \texttt{chayanmath25@gmail.com} \\
  \And
  Rajat K.~De \\
  Machine Intelligence Unit \\ Indian Statistical Institute \\
  203 Barrackpore Trunk Road \\ Kolkata 700108 \\
  \texttt{rajat@isical.ac.in} \\
}

\begin{document}

\maketitle

\begin{abstract}

In real-world machine learning applications, incomplete observations create a fundamental challenge. Researchers have come up with several ideas to address this crucial problem. However, current models still face challenges in balancing scalability and structural consistency. This study proposes a feature imputation method, called FILLER, that deliberately searches the two-dimensional latent space produced by a generative model and fills the missing values with appropriate entries. The generative model is trained on fully observed data to generate samples from the latent space, and FILLER uses this trained model to impute the values missing in the corrupted test samples. In this study, G-NeuroDAVIS serves the purpose of the generative model. This work also presents a mathematical proof on the convergence of the iterative search. Finally, FILLER has been evaluated on several image datasets under random and structured missingness patterns with varying levels of imputation complexities. In order to justify the efficacy of FILLER, it has been compared against existing state-of-the-art solution strategies in terms of RMSE, PSNR, and SSIM. In addition, Wilcoxon signed-rank test has been carried out to validate statistical significance. Moreover, downstream analyses (classification and clustering) have also established the quality of imputation in terms of standard metrics.

\end{abstract}

\section{Introduction}\label{sec:introduction}


Missing data poses a fundamental challenge in real-world machine learning workflows, arising from acquisition errors, data corruption, experimental limitations, privacy constraints, and sensor failures. Incomplete data can substantially degrade the performance of downstream tasks, including classification, regression, and clustering. Therefore, effective data imputation is a critical preprocessing step in many analytical pipelines \citep{Survey2023}. Consequently, the development of robust imputation techniques has attracted sustained interest across disciplines such as data science, computer vision, and computational biology.


A broad range of imputation techniques has been developed to address this problem.
%
%
Traditional data imputation methods include zero imputation, mean or median substitution, \(k\)-nearest neighbors (\(k\)-NN) and singular value decomposition (SVD) based imputation \citep{KNNimpute}. While these methods are computationally efficient and easy to implement, they rely on local similarity or simple statistical assumptions and are often inadequate in high-dimensional settings, particularly under complex missingness patterns.
%
%
More recent works have focused on deep generative models, for which data imputation is a natural application. The generative adversarial imputation network (GAIN) \citep{GAIN} extends the traditional generative adversarial network (GAN) \citep{GAN} to handle datasets containing missing values. GAIN formulates imputation as a generative adversarial learning problem in which the generator imputes missing data, while the discriminator learns to distinguish observed entries from imputed ones. To ensure that the generated samples are consistent with the true underlying data distribution, the discriminator is provided with additional information in the form of hints. A generative adversarial multiple imputation network (GAMIN) is proposed in \cite{GAMIN} for performing multiple imputation under high rates of missingness. In MIWAE \citep{MIWAE}, deep latent variable models can be trained on incomplete datasets using an objective known as the missing data importance-weighted autoencoder bound and subsequently applied to single or multiple imputation. These approaches have demonstrated improved performance over classical methods in a variety of settings, particularly when complex non-linear relationships exist among features.


Single imputation methods replace each missing entry with a single estimated value, thereby producing a complete dataset suitable for standard analysis. Techniques such as mean or median substitution, KNNimpute, SVDimpute, and matrix factorization \citep{MFI} fall into this category. In contrast, multiple imputation generates several plausible completions of the data. Multivariate Imputation by Chained Equations (MICE) \citep{MICE} and missForest \citep{MissForest} are widely used examples of this approach. Stochastic imputation methods can be applied repeatedly to generate multiple imputations \citep{GAMIN}.


Despite the success, current models face challenges in balancing scalability and structural consistency. Image datasets require models that respect local continuity and spatial correlations. The proposed method, called FILLER (Feature Imputation via Latent Location Exploration and Retrieval), in this work aims to address these gaps by learning from fully observed training data to infer missing values in test samples, combining statistical robustness with deep generative modelling. It introduces a two-dimensional latent space search technique to impute a corrupted sample in hand. Recent studies have shown the effectiveness of latent space exploration in solving several complex tasks efficiently \citep{Learning2021, GANInversion, Searching2026}.


This study proposes a data-driven, generative, single imputation method trained exclusively on fully observed samples and subsequently applied to impute missing values in unseen, incomplete samples. The approach is evaluated on several datasets with varying levels of complexity and dimensionality. A wide range of missingness percentages is considered, including both random and structured missingness \citep{StructuredMissingness} patterns. Imputation performance is assessed using standard reconstruction metrics, and the downstream utility of the imputed data is further evaluated using classification and clustering analyses.


The article is organized into four main sections. Section~\ref{sec:introduction} presents the motivation for addressing the problem of missing data, outlines the objectives of the proposed imputation method, and provides background on existing approaches. Section~\ref{sec:methodology} provides a detailed description of the proposed imputation method. In Section~\ref{sec:results}, the experimental findings are reported and analyzed under different missingness patterns and percentages, and imputation quality is evaluated based on several metrics, as well as downstream classification and clustering performance. Finally, Section~\ref{sec:conclusion} summarizes the key findings, discusses their implications, and suggests potential directions for future work on robust and scalable data imputation.

\section{Methodology}\label{sec:methodology}

This section describes the problem scenario and the motivation behind FILLER, and then the method is described in detail.

\subsection{Problem scenario and solution approach}\label{sec:problem-solution}

Many real-world datasets contain missing values for a variety of reasons, commonly represented as blanks, `NaN's, or other placeholders. Such datasets are not directly compatible with most machine learning algorithms, which require that all values in a data sample be intact, numerical, and meaningful. A simple approach to handle these datasets is to discard entire rows or columns that contain missing values. This approach may introduce bias \citep{Survey2023} and result in the loss of potentially useful information despite the data being incomplete. A more effective alternative is to perform imputation, where the missing values are estimated from the available data.\footnote{\url{https://scikit-learn.org/stable/modules/impute.html}}



In order to impute a data sample, one needs to know the inherent distribution it follows. Motivated by this fact, in this study, a generative deep neural network model, called G-NeuroDAVIS \citep{GNeuroDAVIS}, has been considered to serve the purpose of data imputation. However, one can choose any other generative model that produces a two-dimensional latent embedding to achieve the same.
G-NeuroDAVIS consists of an encoder and a decoder. The encoder projects the data into a latent space, and the decoder projects it back to the original space. The encoder is capable of producing a generalized embedding from which it can generate realistic and diverse samples using the decoder. Moreover, the latent space it produces supports smooth interpolation among data points and offers interpretability for controlling and manipulating generated samples.

Initially, the proposed imputation method explores the two-dimensional latent space deliberately and generates samples. Later, it exploits the generated samples to obtain a better match to the corrupt sample (with missing value(s)) in hand. This process of exploration and exploitation has been performed iteratively to achieve the best possible match to the corrupted data, and accordingly, the missing values get imputed. In the next section, the process of imputation has been described with all the mathematical details.

\subsection{FILLER}

FILLER consists of two major parts. The first part is about the training of the generative model (which is G-NeuroDAVIS here), and the second part performs an iterative search in the two-dimensional latent space which has been obtained from the generative method, for a close reconstruction of the corrupted data sample. It may be mentioned here that the generative model training requires complete training data, i.e., data without any missing values. Therefore, the data has been partitioned into training and test subsets using standard splitting ratios (if unavailable, using an \(80{:}20\) ratio); the training subset has been used to train the generative model, and the test subset has been used to demonstrate the effectiveness of the imputation method. A mask has been considered for each of the test samples, which decides the values being treated as NaNs. These corrupted values have been imputed using FILLER with the help of a trained generative model.

Let
\(
\bm{X} = \{\bm{x}_{i} \mid \bm{x}_{i} \in \mathbb{R}^{d}\}_{i = 1}^{n}
\)
be a dataset consisting of \(n\) samples and characterized by \(d\) features. Also, let
\(
\bm{X}^\mathrm{[train]} = \{
  \bm{x}^\mathrm{[train]}_{i} \mid \bm{x}^\mathrm{[train]}_{i} \in \mathbb{R}^{d}
\}_{i=1}^{n_1}
\)
and
\(
\bm{X}^\mathrm{[test]} = \{
  \bm{x}^\mathrm{[test]}_{i} \mid \bm{x}^\mathrm{[test]}_{i} \in \mathbb{R}^{d}
\}_{i=1}^{n_2}
\)
be a disjoint partition of the dataset \(\bm{X}\). Here, \(\bm{X}^\mathrm{[train]}\) has been utilized to train the G-NeuroDAVIS model. On the other hand, \(\bm{X}^\mathrm{[test]}\) has been corrupted intentionally with a suitable mask
\(
\bm{M} = \{\bm{m}_{i} \mid \bm{m}_{i} \in \{0,1\}^{d}\}_{i=1}^{n_2}
\),
and considered for validation of the proposed imputation method. In this section, the imputation process of a corrupted test sample (say \(\bm{x}^\mathrm{[test]}_{k} \odot \bm{m}_{k}\)) has been described. Here, \(\bm{m}_{k}\) is a binary mask that corrupts an \(k\)\textsuperscript{th} sample by making certain positions of the sample zero.

After a successful training of G-NeuroDAVIS, the two-dimensional latent embedding has been extracted using the encoder, i.e.,
\(
\bm{X}^\mathrm{[lat]} = \operatorname{ENCODER}(\bm{X}^\mathrm{[train]})
\).
In order to achieve a less complex search space, a two-dimensional latent space in the G-NeuroDAVIS architecture has been considered, i.e.,
\(
\bm{X}^\mathrm{[lat]}  = \{
  \bm{x}^\mathrm{[lat]}_{i} \mid \bm{x}^\mathrm{[lat]}_{i} \in \mathbb{R}^{2}
\}_{i=1}^{n_1}
\).
Here, \(\bm{x}^\mathrm{[lat]}_{i}\) represents the projection of the training sample \(\bm{x}^\mathrm{[train]}_{i}\) to two-dimensional space. For a better geometrical understanding, let \((x_{i}^\mathrm{[lat]}, y_{i}^\mathrm{[lat]})\) be the coordinates of \(\bm{x}^\mathrm{[lat]}_{i}\) in the two-dimensional latent space. The initial bounds of the search area have been computed from the embeddings themselves. The bounded area is demarcated by the points
\((x_{\min}^{(1)}, y_{\min}^{(1)})\),
\((x_{\max}^{(1)}, y_{\min}^{(1)})\),
\((x_{\max}^{(1)}, y_{\max}^{(1)})\), and
\((x_{\min}^{(1)}, y_{\max}^{(1)})\), where
\begin{align}\label{eq:bbox-init}
  x_{\min}^{(1)} &= \min(\{
    x_{i}^\mathrm{[lat]} \mid (x_{i}^\mathrm{[lat]}, y_{i}^\mathrm{[lat]}) \in \bm{X}^\mathrm{[lat]} \}),
& x_{\max}^{(1)} &= \max(\{
    x_{i}^\mathrm{[lat]} \mid (x_{i}^\mathrm{[lat]}, y_{i}^\mathrm{[lat]}) \in \bm{X}^\mathrm{[lat]} \}), \nonumber \\
  y_{\min}^{(1)} &= \min(\{
    y_{i}^\mathrm{[lat]} \mid (x_{i}^\mathrm{[lat]}, y_{i}^\mathrm{[lat]}) \in \bm{X}^\mathrm{[lat]} \}),
& y_{\max}^{(1)} &= \max(\{
    y_{i}^\mathrm{[lat]} \mid (x_{i}^\mathrm{[lat]}, y_{i}^\mathrm{[lat]}) \in \bm{X}^\mathrm{[lat]} \}).
\end{align}
Let the bounding box be defined as the Cartesian product of intervals, i.e.,
\[
\mathsf{BBOX}^{(1)} = [x_{\min}^{(1)}, x_{\max}^{(1)}] \times [y_{\min}^{(1)}, y_{\max}^{(1)}].
\]
Thereafter, a set
\(
\Lambda^{(1)} = \{\bm{\alpha}_{1}^{(1)}, \bm{\alpha}_{2}^{(1)}, \dots, \bm{\alpha}_{p}^{(1)}\}
\)
of \(p\) points is sampled uniformly from the bounded area in \(\mathsf{BBOX}^{(1)}\), and passed through the trained decoder to obtain their reconstructions. These reconstructions are compared with the test sample \(\bm{x}^\mathrm{[test]}_{k}\) intended for imputation. Let \(\bm{\alpha}^{(1)}_{*} = ({\alpha}^{(1)}_{*}, {\beta}^{(1)}_{*})\) be the point that produces the closest reconstruction to the corrupted version of \(\bm{x}^\mathrm{[test]}_{k}\). Mathematically,
\[
\bm{\alpha}^{(1)}_{*} = \underset{\bm{\alpha} \in \Lambda^{(1)}}{\arg\min}\: \lVert
  \bm{x}^\mathrm{[test]}_{k} \odot \bm{m}_{k} -
  \operatorname{DECODER}(\bm{\alpha}) \odot \bm{m}_{k} \rVert_2^2
\]
Then in the next iteration, another set \(\Lambda^{(2)}\) and another smaller bounding box \(\mathsf{BBOX}^{(2)}\) have been created. The set \(\Lambda^{(2)}\) consists of \(p\) points that include \(\bm{\alpha}^{(1)}_{*}\), and the other \((p - 1)\) points \((\bm{\alpha}^{(2)}_{1},\bm{\alpha}^{(2)}_{2}, \dots, \bm{\alpha}^{(2)}_{p-1})\) have been sampled uniformly from the smaller bounding box \(\mathsf{BBOX}^{(2)}\). 
Therefore, the set \(\Lambda^{(2)}\) and \(\mathsf{BBOX}^{(2)}\) look like
\begin{gather*}
\Lambda^{(2)} = \{
  \bm{\alpha}^{(1)}_{*},        \bm{\alpha}^{(2)}_{1},
  \bm{\alpha}^{(2)}_{2}, \dots, \bm{\alpha}^{(2)}_{p-1}\}; \quad
\mathsf{BBOX}^{(2)} = [x_{\min}^{(2)}, x_{\max}^{(2)}] \times [y_{\min}^{(2)}, y_{\max}^{(2)}]
\end{gather*}
where
\begin{align}\label{eq:bbox-update}
  l &= \tfrac{1}{4}(x_{\max}^{(1)} - x_{\min}^{(1)}),
& h &= \tfrac{1}{4}(y_{\max}^{(1)} - y_{\min}^{(1)}), \nonumber \\
  x_{\min}^{(2)} &= \max(x_{\min}^{(1)}, {\alpha}^{(1)}_{*} - l),
& x_{\max}^{(2)} &= \min(x_{\max}^{(1)}, {\alpha}^{(1)}_{*} + l), \nonumber \\
  y_{\min}^{(2)} &= \max(y_{\min}^{(1)}, {\beta}^{(1)}_{*} - h),
& y_{\max}^{(2)} &= \min(y_{\max}^{(1)}, {\beta}^{(1)}_{*} + h).
\end{align}
The points in \(\Lambda^{(2)}\) are again passed through the decoder to obtain their reconstructions. Again, the closest approximation is determined by observing the masked loss, and accordingly, the bounding box \(\mathsf{BBOX}^{(3)}\) and \(\Lambda^{(3)}\) are created. This process is repeated multiple times till convergence. After \(t\) iterations,
\begin{gather*}
\Lambda^{(t)} = \{
  \bm{\alpha}^{(t-1)}_{*},        \bm{\alpha}^{(t)}_{1},
  \bm{\alpha}^{(t)}_{2},   \dots, \bm{\alpha}^{(t)}_{p-1} \}; \quad
\mathsf{BBOX}^{(t)} = [x_{\min}^{(t)}, x_{\max}^{(t)}] \times [y_{\min}^{(t)}, y_{\max}^{(t)}] \\ \text{where}\
\bm{\alpha}^{(t-1)}_{*}
  = ({\alpha}^{(t-1)}_{*}, {\beta}^{(t-1)}_{*})
  = \underset{\bm{\alpha} \in \Lambda^{(t-1)}}{\arg\min} \lVert
    \bm{x}^\mathrm{[test]}_{k} \odot \bm{m}_{k} -
    \operatorname{DECODER}(\bm{\alpha}) \odot \bm{m}_{k} \rVert_2^2
\end{gather*}
\begin{align}\label{eq:bbox-update_t}
  l &= \tfrac{1}{4}(x_{\max}^{(t-1)} - x_{\min}^{(t-1)}),
& h &= \tfrac{1}{4}(y_{\max}^{(t-1)} - y_{\min}^{(t-1)}), \nonumber \\
  x_{\min}^{(t)} &= \max(x_{\min}^{(t-1)}, {\alpha}^{(t-1)}_{*} - l),
& x_{\max}^{(t)} &= \min(x_{\max}^{(t-1)}, {\alpha}^{(t-1)}_{*} + l), \nonumber \\
  y_{\min}^{(t)} &= \max(y_{\min}^{(t-1)}, {\beta}^{(t-1)}_{*} - h),
& y_{\max}^{(t)} &= \min(y_{\max}^{(t-1)}, {\beta}^{(t-1)}_{*} + h).
\end{align}
A few iterations of the entire search have been depicted in Figure~\ref{fig:MNIST-search} for a better understanding.

Without loss of generality, let the earlier process converges after \(\tau\) iterations, and therefore, the closest approximation of \(\bm{x}^\mathrm{[test]}_{k} \odot \bm{m}_{k}\) will be \(\operatorname{DECODER}(\bm{\alpha}^{(\tau)}_*) \odot \bm{m}_{k}\). Let \(\hat{\bm{x}}^\mathrm{[test]}_{k}\) be the imputed version of \(\bm{x}^\mathrm{[test]}_{k} \odot \bm{m}_{k}\), and thus, it can be represented as
\[
\hat{\bm{x}}^\mathrm{[test]}_{k} =
  \bm{x}^\mathrm{[test]}_{k} \odot \bm{m}_{k} +
  \operatorname{DECODER}(\bm{\alpha}^{(\tau)}_{*}) \odot (\bm{1} - \bm{m}_{k})
\]
Here, \((\bm{1} - \bm{m}_{k})\) represents the binary complement of \(\bm{m}_{k}\). The above equation suggests that the values that are forcefully made zero have been recovered from the closest approximation obtained by the iterative search. The entire search algorithm (Algorithm~\ref{alg:imputation}) along with its time complexity (Appendix~\ref{sec:algorithm}) and the proof of convergence (Appendix~\ref{sec:proof}) have been depicted in the Appendices.

\section{Results}\label{sec:results}

The proposed imputation method has been evaluated on several datasets with varying complexities and dimensionalities.
%
%
The root mean squared error (RMSE) has been used as the primary quantitative metric across all datasets. Additionally, peak signal-to-noise ratio (PSNR) is reported to assess reconstruction fidelity. In order to complement pixel-wise error metrics, the structural similarity index (SSIM) \citep{SSIM} is also reported for the datasets. SSIM is designed to capture perceptual image quality by comparing local patterns of luminance, contrast, and structural information between the ground-truth and reconstructed images. Unlike RMSE and PSNR, which operate purely on pixel-wise differences, SSIM correlates more closely with human visual perception. As with the other metrics, SSIM is computed with respect to the imputed regions to specifically evaluate the structural consistency of the reconstructed missing areas.

The proposed method has been benchmarked against widely used imputation baselines GAIN and MIWAE under a test-time imputation setting. Through extensive experiments\footnote{Source code to reproduce the results is available at \url{https://github.com/SantuMondal002/FILLER}.}, the strengths and limitations of latent search-based imputation have been examined relative to other imputation strategies, thereby demonstrating the conditions under which access to fully observed data can yield substantial improvements in imputation performance and downstream analysis.
%
%
The experiments have been conducted on a workstation equipped with an Intel\textregistered{} Core\texttrademark{} i9-14900K processor, 128 GB RAM, and a 24 GB NVIDIA RTX 4500 Ada Generation GPU, running 64-bit Ubuntu 22.04.5 LTS operating system.

\subsection{Data description}

In the current study, three image datasets, viz., \textit{CMU Face Images}, \textit{Fashion-MNIST}, and \textit{MNIST} have been considered. The \textit{CMU Face Images} dataset \citep{CMUFaceImages} consists of 640 grayscale face images of 20 people with varying poses and expressions. Out of these, 624 images have been used, as some images have been damaged. The images are available in full resolution \((128 \times 120)\), half resolution \((64 \times 60)\), and quarter resolution \((32 \times 30)\). For a robust model evaluation, this dataset has been subsequently partitioned into training and test subsets using an \(80{:}20\) ratio. The \textit{MNIST} dataset \citep{MNIST} contains \(28 \times 28\) grayscale images of handwritten digits (0--9). The \textit{Fashion-MNIST} dataset \citep{FMNIST} serves as a more challenging alternative to the original \textit{MNIST}, containing \(28 \times 28\) grayscale images of fashion products from 10 categories. Both datasets contain 60,000 training samples and 10,000 test samples. These datasets have undergone preprocessing steps to ensure consistency and compatibility with the proposed methodology. The pixel intensity values have been rescaled to the range \([0, 1]\). 

\subsection{Imputation results}

\begin{figure}[t]\centering 
\begin{subfigure}{0.285\linewidth}\centering
\begin{tikzpicture}
  \useasboundingbox (0,0) rectangle (\linewidth,0.75\linewidth);
  \node[anchor=south west, inner sep=0] (img)
    at (0,0) {\includegraphics[width=\linewidth]{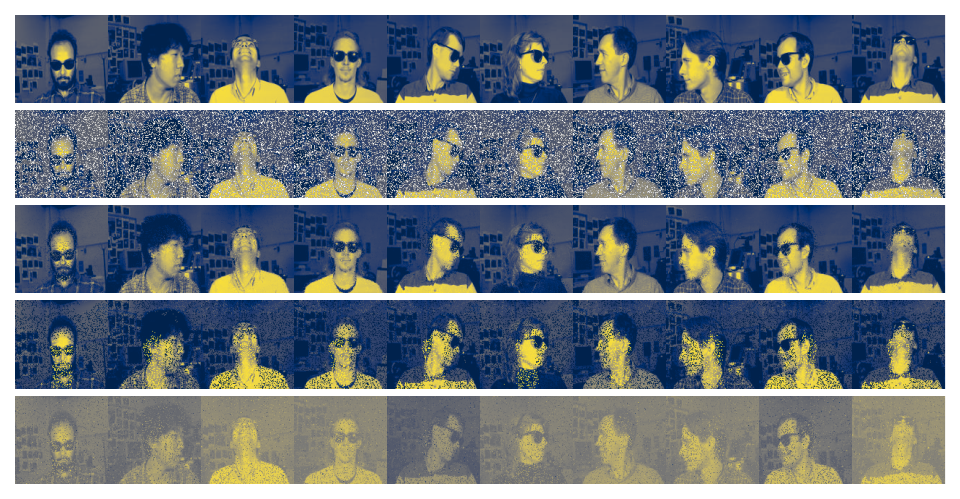}};
  \begin{scope}[x={(img.south east)}, y={(img.north west)}]
    \bfseries\scriptsize
    \node[left] at (0, 0.9) {Original};
    \node[left] at (0, 0.7) {Corrupted};
    \node[left] at (0, 0.5) {FILLER};
    \node[left] at (0, 0.3) {GAIN};
    \node[left] at (0, 0.1) {MIWAE};
  \end{scope}
\end{tikzpicture}
\caption{Random Missingness}\label{fig:showcase-Faces-random}
\end{subfigure}
\begin{subfigure}{0.285\linewidth}\centering
\includegraphics[width=\linewidth]{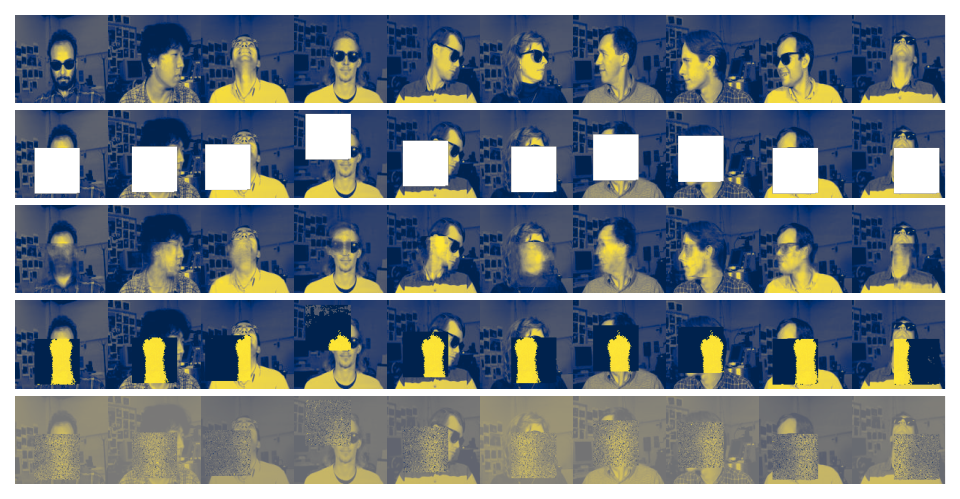}
\caption{Block-wise Missingness}\label{fig:showcase-Faces-block}
\end{subfigure}
\begin{subfigure}{0.285\linewidth}\centering
\includegraphics[width=\linewidth]{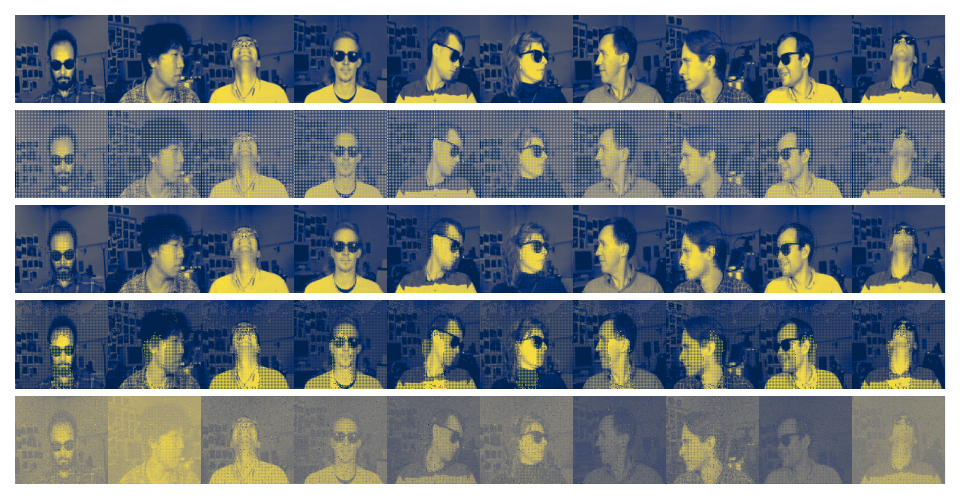}
\caption{Grid-wise Missingness}\label{fig:showcase-Faces-grid}
\end{subfigure}
\caption{A qualitative comparison of various imputation methods. The figure illustrates imputation examples from the full-resolution \textit{CMU Face Images} dataset. In each column, different types of missingness (25\%) are demonstrated. Each subfigure contains samples from different classes in the dataset.}
\label{fig:showcase-Faces}
\end{figure}
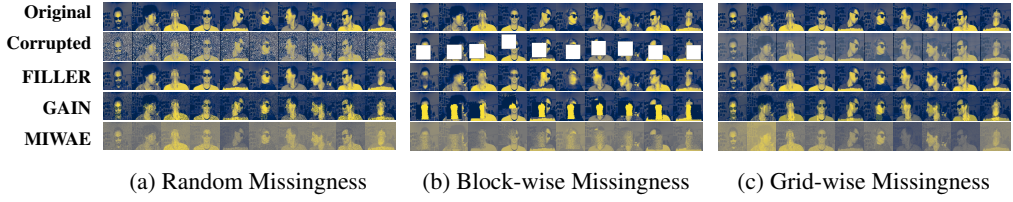

\begin{figure}[t]\centering 
\includegraphics[width=\linewidth]{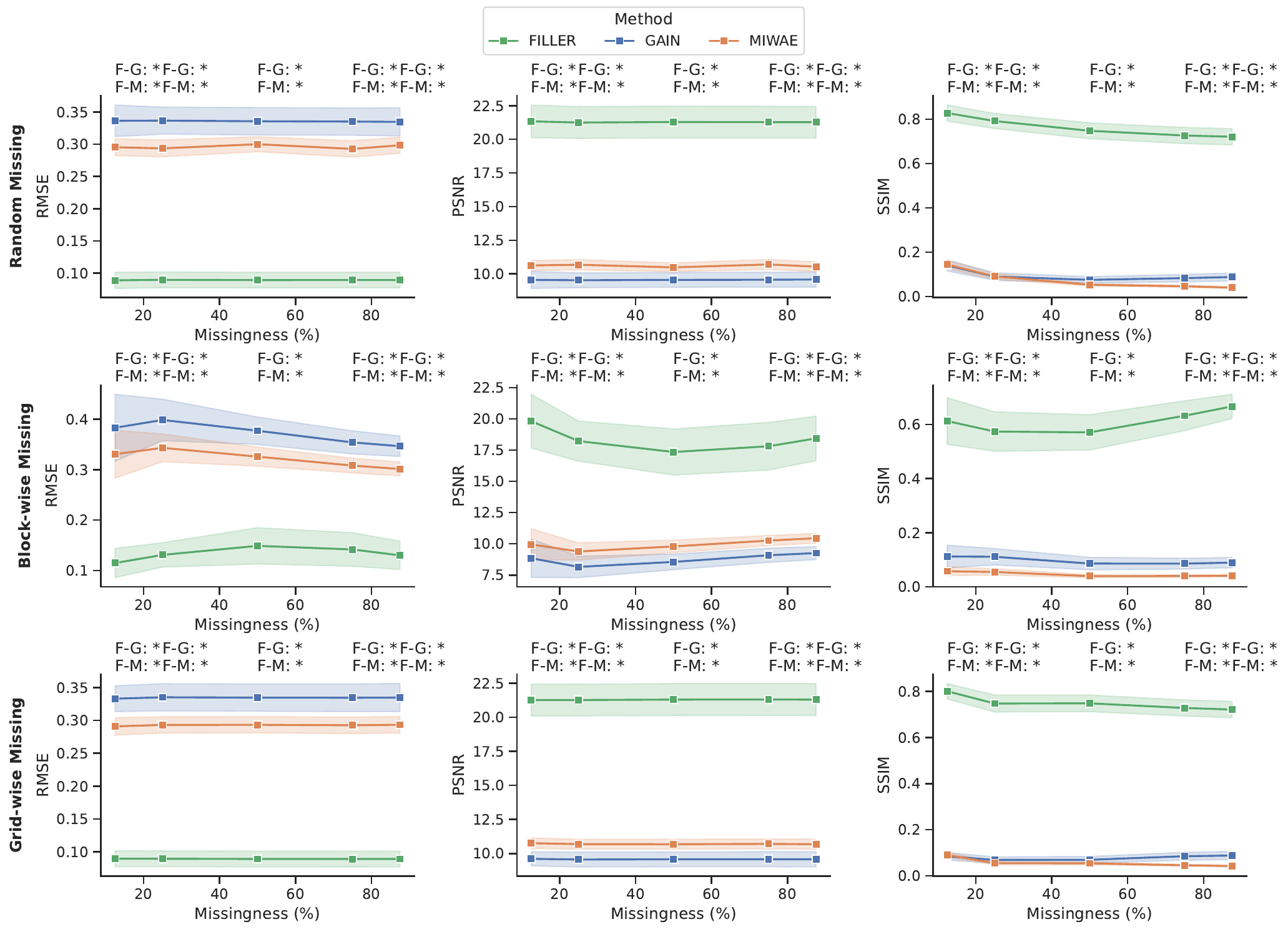}
\caption{Comparison of several imputation methods based on RMSE, PSNR, and SSIM on the full-resolution \textit{CMU Face Images} dataset with varying levels of missingness. Statistical significance has been measured with a Wilcoxon signed-rank test. F-G and F-M denote the significance levels of FILLER against GAIN and MIWAE, respectively. \(*\) indicates statistical significance at 0.05 level.}
\label{fig:Faces-comparison}
\end{figure}

The datasets have been used under random missingness as well as two structured missingness scenarios: block-like and grid-like patterns. The missingness percentages range from 12.5\% to 87.5\% to stress test imputation robustness in image data. Initially, the imputation performance of FILLER has been validated visually (Figure~\ref{fig:showcase-Faces}), then validated using standard metrics (Figure~\ref{fig:Faces-comparison}), and compared against the state-of-the-art methods, viz., GAIN and MIWAE. In addition, a Wilcoxon signed-rank test has been conducted to validate the level of significance in the performance metrics.

For the \textit{CMU Face Images} dataset, it is evident from Figure~\ref{fig:showcase-Faces-random} that GAIN and MIWAE produced reconstructions are noisy; however, FILLER fills the missing pixels precisely. In addition, the background of each image has been successfully retrieved by FILLER. In the case of block-wise missingness, both GAIN and MIWAE have shown poor performance. In contrast, FILLER has produced somewhat noisy but realistic outputs (Figure~\ref{fig:showcase-Faces-block}). Finally, in the case of grid-wise missingness, FILLER has significantly reduced visual artifacts that have been seen in the corrupted image, but the state-of-the-art methods have failed to achieve the same (Figure~\ref{fig:showcase-Faces-grid}). A similar result has been observed when comparing using standard metrics. Figure~\ref{fig:Faces-comparison} shows that for the \textit{CMU Face Images} dataset, FILLER has outperformed all other state-of-the-art methods, significantly, in terms of both imputation and perceptual quality, which has been measured using RMSE, PSNR, and SSIM. This performance trend continues across all three available resolutions, demonstrating the scalability and robustness of FILLER. In addition, comparison has also been made on the lower-resolution versions of the \textit{CMU Face Images} dataset. Visual comparison has been observed in Figure~\ref{fig:showcase-Faces2-Faces4}, and comparison with respect to the metrics has been observed in Figures \ref{fig:Faces2-comparison} and \ref{fig:Faces4-comparison}.


Figure~\ref{fig:showcase-FMNIST} reflects the visual comparison among the imputation methods for the \textit{Fashion-MNIST} dataset over the three different types of missingness. Visually, FILLER has produced clearer images compared against GAIN and MIWAE. Across all the different missingness scenarios, FILLER has not only imputed the original object but also recovered the background as well. When compared using metrics, it has been observed that, in the case of random missingness, FILLER demonstrates stable performance across all metrics throughout the range of missingness, while the performance of GAIN has declined under extreme missingness (Figure~\ref{fig:FMNIST-comparison} in the Appendices). For block-like missingness situations, FILLER scales steadily across the range of missingness, whereas the performance of GAIN is negatively impacted as missingness increases (Figure~\ref{fig:FMNIST-comparison} in the Appendices). In the case of grid-like missingness, FILLER scales steadily across the entire range of missingness and performs comparatively better than the other methods (Figure~\ref{fig:FMNIST-comparison} in the Appendices). Besides, the improvements achieved are also statistically significant, supported by a Wilcoxon signed-rank test.


Similar to \textit{Fashion-MNIST}, imputation for the \textit{MNIST} dataset by all the methods has also revealed similar visual observations (Figure~\ref{fig:showcase-MNIST} in the Appendices). Besides, Figure~\ref{fig:showcase-MNIST-block} reflects that the inputs corresponding to the digits 3, 5, 6, 7, and 8 lack readability after applying a block-wise corruption. FILLER has imputed it in such a way that it has become readable again. The state-of-the-art has failed to perform in a similar manner, which has demonstrated the effectiveness of FILLER. In the case of random missingness in the \textit{MNIST} dataset, the imputation performance of FILLER remains stable even under extremely high missingness scenarios. GAIN has performed better for low missing percentages, but the RMSE sharply increases with increasing missingness (Figure~\ref{fig:MNIST-comparison} in the Appendices). FILLER exhibits significantly high PSNR and SSIM values even under extreme missingness conditions. In the case of block-like missingness, FILLER has performed comparatively better in low missingness percentages; MIWAE has started to perform better from 50\% missingness onwards (Figure~\ref{fig:MNIST-comparison} in the Appendices). In grid-like missingness scenarios, FILLER has performed evenly across the entire range of missingness. GAIN has struggled with steadiness throughout the range of missingness.

It may be mentioned here that to achieve the above-mentioned results, G-NeuroDAVIS takes fewer parameters compared to GAIN; however, MIWAE takes even fewer parameters to perform. Table~\ref{tab:parameters} reports the number of learnable parameters used to fulfill the task of imputation across multiple datasets with varying input sizes.

\subsection{Downstream analysis}

While reconstruction-based metrics quantify the accuracy of imputed values, they do not fully capture the impact of imputation on subsequent analytical tasks. In order to assess the practical utility of the imputed data, the effect of different imputation methods has been evaluated with respect to downstream classification and clustering performance. This evaluation aims to determine whether improvements in imputation accuracy translate into meaningful gains in task-level performance. 

\begin{figure}[t]\centering 
\begin{subfigure}[t]{\linewidth}\centering 
\includegraphics[width=\linewidth]{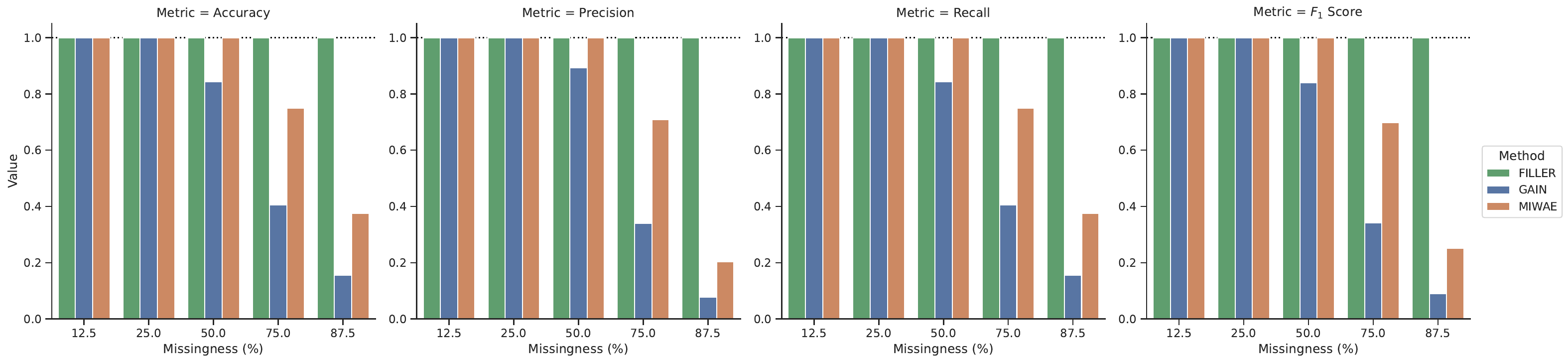}
\caption{Random Missingness}\label{fig:Faces-random-RFC}
\end{subfigure}
\begin{subfigure}[t]{\linewidth}\centering 
\includegraphics[width=\linewidth]{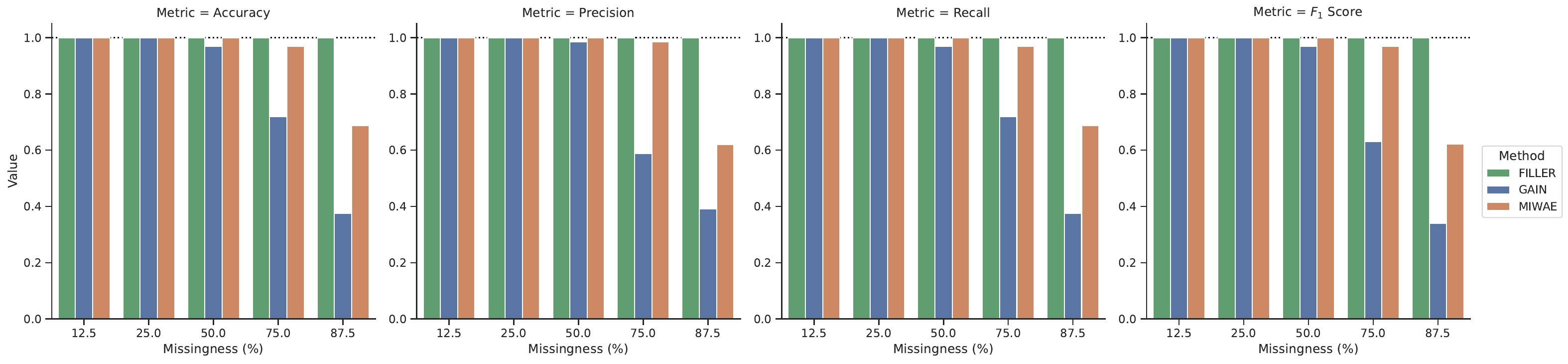}
\caption{Block-wise Missingness}\label{fig:Faces-block-RFC}
\end{subfigure}
\begin{subfigure}[t]{\linewidth}\centering 
\includegraphics[width=\linewidth]{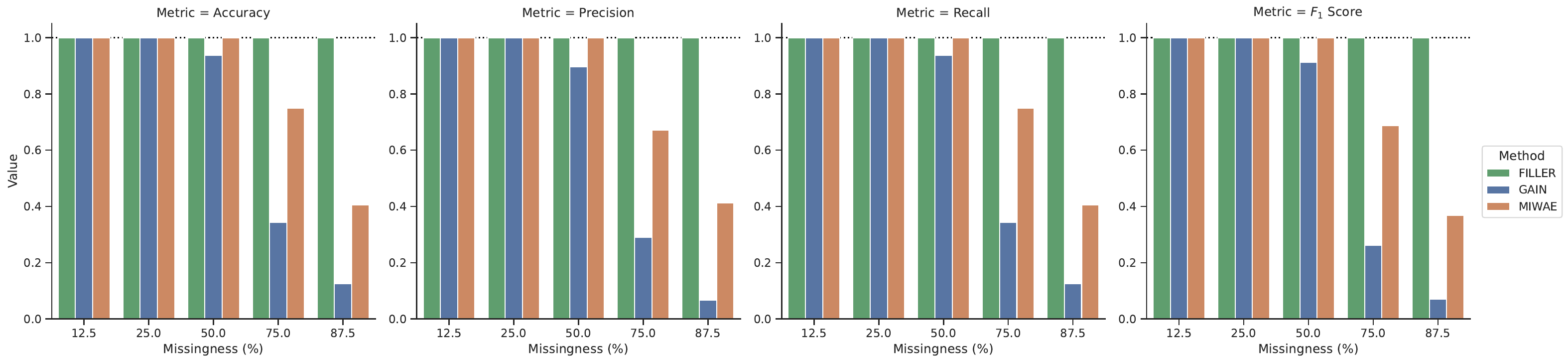}
\caption{Grid-wise Missingness}\label{fig:Faces-grid-RFC}
\end{subfigure}
\caption{Classification results of random forest classifier on the full-resolution \textit{CMU Face Images} dataset with three types of missingness. The dotted lines indicate the classification performance on the original test data.}
\label{fig:Faces-RFC}
\end{figure}

Downstream analyses have been performed on the imputed test data produced by each method. The same downstream models and evaluation protocols have been applied consistently across all imputation methods to ensure a fair comparison. Downstream analyses have been conducted independently for each missingness level and pattern. To provide context for the observed performance, downstream results obtained from imputed data have been compared against those obtained from the corresponding fully observed test data. This comparison highlights the extent to which imputation mitigates the degradation in downstream performance caused by missing data.


For classification tasks, two standard classifiers, viz., random forest and support vector classifier, and for clustering tasks, \(k\)-means and agglomerative clustering from the scikit-learn \citep{scikit-learn} library have been employed. These models have been trained on the fully observed training data and evaluated on the corresponding imputed test data using ground-truth labels. Performance has been reported using commonly adopted classification metrics, viz., accuracy, precision, recall, and \(\text{F}_1\) score. All classifiers have been trained with identical hyperparameters across all imputation methods to isolate the effect of the imputation strategy on predictive performance. Likewise, clustering has been performed independently for each imputation method using identical algorithmic settings. The resulting cluster assignments have been evaluated against available ground-truth labels using external clustering metrics, viz., Adjusted Mutual Information (AMI), Adjusted Rand Index (ARI), Fowlkes-Mallows Index (FMI), and Normalized Mutual Information (NMI). It has enabled a direct comparison of how different imputation strategies affect the preservation of intrinsic data structure.

\begin{figure}[t]\centering 
\begin{subfigure}[t]{\linewidth}\centering 
\includegraphics[width=\linewidth]{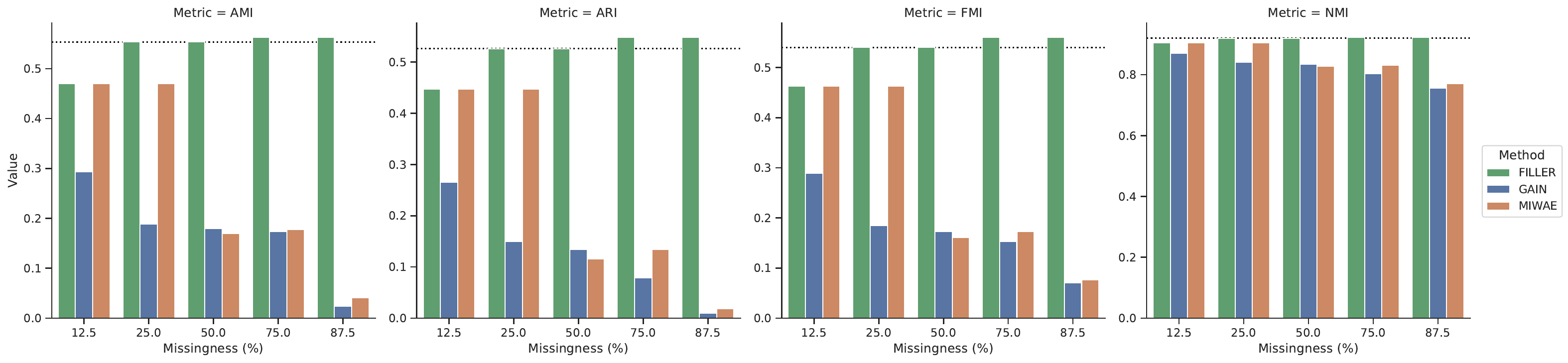}
\caption{Random Missingness}\label{fig:Faces-random-AC}
\end{subfigure}
\begin{subfigure}[t]{\linewidth}\centering 
\includegraphics[width=\linewidth]{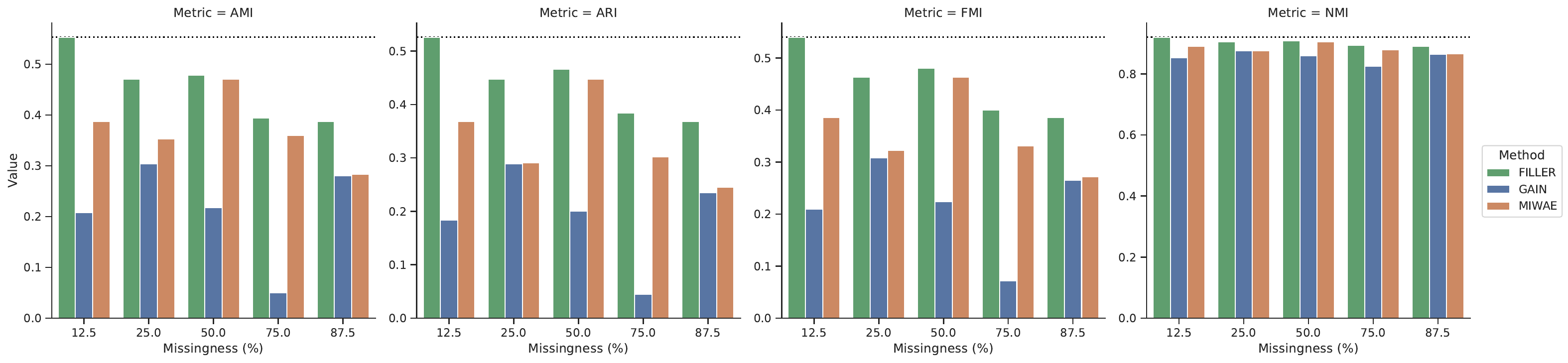}
\caption{Block-wise Missingness}\label{fig:Faces-block-AC}
\end{subfigure}
\begin{subfigure}[t]{\linewidth}\centering 
\includegraphics[width=\linewidth]{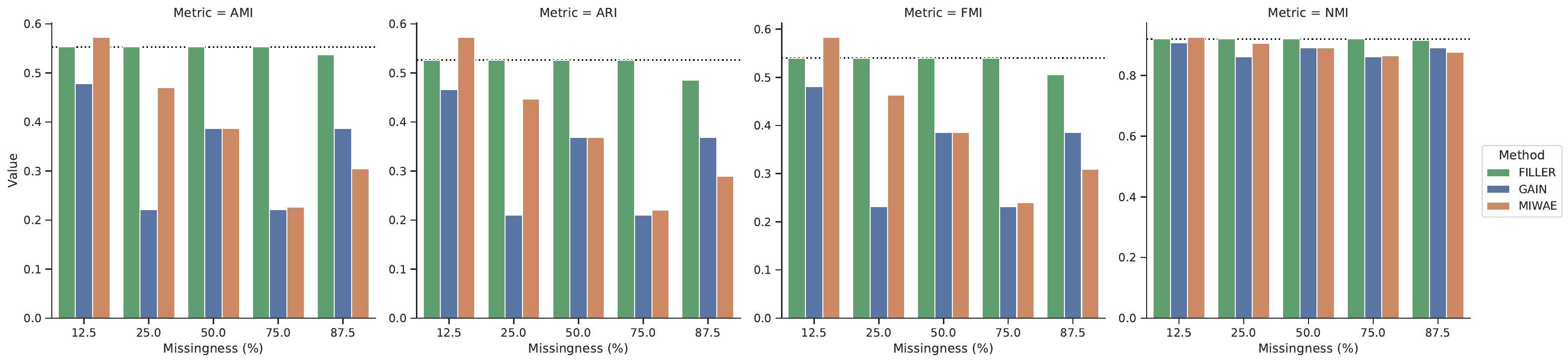}
\caption{Grid-like Missingness}\label{fig:Faces-grid-AC}
\end{subfigure}
\caption{Clustering results of agglomerative clustering on the full-resolution \textit{CMU Face Images} dataset with three types of missingness. The dotted lines indicate the clustering performance on the original test data.}
\label{fig:Faces-AC}
\end{figure}

Figure~\ref{fig:Faces-RFC} has reflected a stable classification performance of random forest classifier over the imputed data produced by FILLER across all missingness percentages on the full-resolution \textit{CMU Face Images}. The classifier has failed to achieve the same on both the imputed data produced by GAIN and MIWAE. All classification metrics drop significantly as the missingness percentage increases over 50\%. Similar results have been observed when support vector classifier is used (Figure~\ref{fig:Faces-SVC}). The clustering performance of agglomerative clustering on the imputed data produced by FILLER, GAIN, and MIWAE shows that FILLER has outperformed GAIN and MIWAE in terms of AMI, ARI, and FMI (Figure~\ref{fig:Faces-AC}). The difference in the performance has become significant as the missingness percentage rises. The usage of \(k\)-means algorithm has also revealed similar results (Figure~\ref{fig:Faces-KMC}). Figures~\ref{fig:Faces2-RFC}--\ref{fig:Faces4-SVC} and \ref{fig:Faces2-AC}--\ref{fig:Faces4-KMC} show a similar trend on both half and quarter-resolution \textit{CMU Face Images} in classification and clustering, respectively. 

Figures \ref{fig:FMNIST-random-RFC}, \ref{fig:FMNIST-random-SVC}, \ref{fig:MNIST-random-RFC}, and \ref{fig:MNIST-random-SVC} have demonstrated that the classification performance on the imputed samples produced by FILLER for the \textit{Fashion-MNIST} and the \textit{MNIST} datasets in a random missingness setup is closely comparable with that obtained by the other state-of-the-art methods. However, when the missingness percentage increases, a significant improvement has been observed for all the classification metrics for both datasets. In the block-wise missingness setup, the accuracy drops for all the methods as the missing percentage increases, but FILLER has outperformed GAIN and MIWAE in terms of all the classification metrics (Figures \ref{fig:FMNIST-block-RFC}, \ref{fig:FMNIST-block-SVC}, \ref{fig:MNIST-block-RFC} and \ref{fig:MNIST-block-SVC}). Finally, for the grid-wise missingness scenario, all results obtained with a small missingness percentage are again comparable with an accuracy close to 0.84 for the \textit{Fashion-MNIST} dataset (Figures \ref{fig:FMNIST-grid-RFC} and \ref{fig:FMNIST-grid-SVC}) and close to 0.95 for the \textit{MNIST} dataset (Figures \ref{fig:MNIST-grid-RFC} and \ref{fig:MNIST-grid-SVC}). With a higher missingness percentage, the accuracy of the FILLER imputed dataset, drops to 0.74 for the \textit{Fashion-MNIST} dataset (Figures \ref{fig:FMNIST-grid-RFC} and \ref{fig:FMNIST-grid-SVC}) and to 0.8 for the \textit{MNIST} dataset (Figures \ref{fig:MNIST-grid-RFC} and \ref{fig:MNIST-grid-SVC}), however, a significant drop in accuracy has been observed on the imputed data produced by GAIN an MIWAE.

Clustering on the imputed datasets in a random missingness setup has revealed that all these methods are closely comparable with each other. Usage of \(k\)-means shows that FILLER is slightly better than the others, sometimes even better than the original test inputs (Figure~\ref{fig:MNIST-random-KMC}); however, agglomerative clustering has produced the exact opposite results (Figure~\ref{fig:MNIST-random-AC}). Figures \ref{fig:MNIST-block-AC} and \ref{fig:MNIST-grid-AC} have shown that the \(k\)-means clustering performance has decreased as the missingness percentage increases (Figures \ref{fig:MNIST-block-KMC} and \ref{fig:MNIST-grid-KMC}); however, agglomerative clustering has shown that FILLER has performed well even with a larger missing percentage (Figures \ref{fig:MNIST-block-AC} and \ref{fig:MNIST-grid-AC}). For the \textit{Fashion-MNIST} dataset, both agglomerative and \(k\)-means clustering have shown comparative clustering performance, in both random (Figures \ref{fig:FMNIST-random-AC} and \ref{fig:FMNIST-random-KMC}) and block-wise (Figures \ref{fig:FMNIST-block-AC} and \ref{fig:FMNIST-block-KMC}) missingness scenarios. For the grid-like missingness scenario, FILLER has outperformed GAIN and MIWAE in terms of all the clustering metrics, when agglomerative clustering has been used (Figure~\ref{fig:FMNIST-grid-AC}), and the results are again comparable when \(k\)-means clustering has been used (Figure~\ref{fig:FMNIST-grid-KMC}).

\section{Conclusion}\label{sec:conclusion}

This work investigates a generative approach to data imputation, in which a model trained on fully observed data has been applied to impute missing values in unseen incomplete samples. G-NeuroDAVIS has been used as the generative model, and the proposed algorithm FILLER searches for a lower-dimensional representative in the latent space produced by the model. It has been proven mathematically that FILLER will converge to a unique point in the latent space. Moreover, FILLER can be coupled with any other generative model that produces a two-dimensional latent embedding. This quality makes FILLER more adaptive to the current scenario. However, the performance will depend on both the interpolating capability of the latent space and the generative capability of the generative model.

Through a comprehensive evaluation across several datasets, imputation performance has been examined under a wide range of missingness levels and patterns. FILLER has been compared with representative deep generative imputation baselines using consistent evaluation protocols. Experimental results demonstrate that leveraging fully observed training data can lead to improved imputation accuracy, particularly at higher levels of missingness and in settings where strong structural dependencies exist among features. These gains have been reflected not only in reconstruction-based metrics but also in downstream classification and clustering performance, indicating that improved imputation may translate into more reliable analytical outcomes. Wilcoxon signed-rank test has also established that the improvements are statistically significant. 

Several limitations demand acknowledgment. FILLER assumes access to fully observed training data --- an assumption that may not hold in all applications. Additionally, while the evaluation has considered multiple datasets and missingness patterns, other forms of missingness and larger-scale settings have remained to be explored. FILLER is designed to search only a two-dimensional latent space. Most deep generative models for complex image datasets utilize much higher-dimensional latent spaces to avoid bottlenecking the generation quality and thus, FILLER cannot be readily coupled with such models. However, in future, FILLER can be extended to a latent space search algorithm that works on higher dimensions. Future research may also extend this framework to partially observed training data, incorporate multiple imputation strategies, and examine robustness under distributional shift. In summary, this study contributes to a clearer understanding of the role of training data availability in data imputation and provides empirical evidence that generative imputation can be an effective strategy across diverse data complexities and dimensionalities when appropriate assumptions are met.

\bibliography{References.bib}

@inproceedings{Searching2026,
  title     = {{Searching Latent Program Spaces}},
  author    = {Matthew Macfarlane and Cl{\'e}ment Bonnet},
  booktitle = {{The Thirty-ninth Annual Conference on Neural Information Processing Systems}},
  year      = {2025},
  url       = {https://openreview.net/forum?id=CsXKGIqZtr}}

@article{GNeuroDAVIS,
  author  = {Maitra, Chayan and De, Rajat K.},
  title   = {{G-NeuroDAVIS: A generative model for data visualization through a generalized embedding}},
  journal = {Neural Networks},
  year    = {2026},
  volume  = {193},
  pages   = {107948},
  issn    = {0893-6080},
  doi     = {10.1016/j.neunet.2025.107948}}

@article{Survey2023,
  author  = {Miao, Xiaoye and Wu, Yangyang and Chen, Lu and Gao, Yunjun and Yin, Jianwei},
  title   = {{An Experimental Survey of Missing Data Imputation Algorithms}},
  journal = {IEEE Transactions on Knowledge and Data Engineering},
  year    = {2023},
  volume  = {35},
  number  = {7},
  pages   = {6630--6650},
  doi     = {10.1109/TKDE.2022.3186498}}

@article{StructuredMissingness,
  author = {Mitra, Robin
    and McGough, Sarah F.
    and Chakraborti, Tapabrata
    and Holmes, Chris
    and Copping, Ryan
    and Hagenbuch, Niels
    and Biedermann, Stefanie
    and Noonan, Jack
    and Lehmann, Brieuc
    and Shenvi, Aditi
    and Doan, Xuan Vinh
    and Leslie, David
    and Bianconi, Ginestra
    and Sanchez-Garcia, Ruben
    and Davies, Alisha
    and Mackintosh, Maxine
    and Andrinopoulou, Eleni-Rosalina
    and Basiri, Anahid
    and Harbron, Chris
    and MacArthur, Ben D.},
  title   = {{Learning from data with structured missingness}},
  journal = {Nature Machine Intelligence},
  year    = {2023},
  volume  = {5},
  number  = {1},
  pages   = {13--23},
  issn    = {2522-5839},
  doi     = {10.1038/s4256-022-00596-z}}

@article{GANInversion,
  author = {Xia, Weihao
    and Zhang, Yulun
    and Yang, Yujiu
    and Xue, Jing-Hao
    and Zhou, Bolei
    and Yang, Ming-Hsuan},
  journal = {{IEEE Transactions on Pattern Analysis and Machine Intelligence}},
  title   = {{GAN Inversion: A Survey}},
  year    = {2023},
  volume  = {45},
  number  = {3},
  pages   = {3121--3138},
  doi     = {10.1109/TPAMI.2022.3181070}}

@inproceedings{Learning2021,
  title     = {{Learning a Latent Search Space for Routing Problems using Variational Autoencoders}},
  author    = {Andr{\'e} Hottung and Bhanu Bhandari and Kevin Tierney},
  booktitle = {{International Conference on Learning Representations}},
  year      = {2021},
  url       = {https://openreview.net/forum?id=90JprVrJBO}}

@inproceedings{GAMIN,
  author    = {Yoon, Seongwook and Sull, Sanghoon},
  title     = {{GAMIN: Generative Adversarial Multiple Imputation Network for Highly Missing Data}},
  year      = {2020},
  booktitle = {{2020 IEEE/CVF Conference on Computer Vision and Pattern Recognition (CVPR)}}, 
  pages     = {8453--8461},
  doi       = {10.1109/CVPR42600.2020.00848}}

@inproceedings{MIWAE,
  author    = {Mattei, Pierre-Alexandre and Frellsen, Jes},
  title     = {{MIWAE: Deep Generative Modelling and Imputation of Incomplete Data Sets}},
  booktitle = {{Proceedings of the 36th International Conference on Machine Learning}},
  pages     = {4413--4423},
  year      = {2019},
  editor    = {Chaudhuri, Kamalika and Salakhutdinov, Ruslan},
  volume    = {97},
  series    = {Proceedings of Machine Learning Research},
  publisher = {PMLR},
  url       = {https://proceedings.mlr.press/v97/mattei19a.html}}

@inproceedings{GAIN,
  author    = {Yoon, Jinsung and Jordon, James and van der Schaar, Mihaela},
  title     = {{GAIN: Missing Data Imputation using Generative Adversarial Nets}},
  year      = {2018},
  booktitle = {{Proceedings of the 35th International Conference on Machine Learning}},
  pages     = {5689--5698},
  editor    = {Dy, Jennifer and Krause, Andreas},
  volume    = {80},
  series    = {Proceedings of Machine Learning Research},
  publisher = {PMLR},
  url       = {https://proceedings.mlr.press/v80/yoon18a.html}}

@misc{FMNIST,
  author = {Xiao, Han and Rasul, Kashif and Vollgraf, Roland},
  title  = {{Fashion-MNIST: a Novel Image Dataset for Benchmarking Machine Learning Algorithms}}, 
  year   = {2017},
  eprint = {1708.07747},
  archivePrefix = {arXiv},
  primaryClass  = {cs.LG}}

@inproceedings{GAN,
  author = {Goodfellow, Ian J.
    and Pouget-Abadie, Jean
    and Mirza, Mehdi
    and Xu, Bing
    and Warde-Farley, David
    and Ozair, Sherjil
    and Courville, Aaron
    and Bengio, Yoshua},
  title     = {{Generative Adversarial Nets}},
  year      = {2014},
  booktitle = {{Advances in Neural Information Processing Systems}},
  editor    = {Z. Ghahramani
    and M. Welling
    and C. Cortes
    and N. Lawrence
    and K.Q. Weinberger},
  publisher = {Curran Associates, Inc.},
  volume    = {27},
  pages     = {2672--2680}}

@article{MICE,
  author  = {van Buuren, Stef and Groothuis-Oudshoorn, Karin},
  title   = {{mice: Multivariate Imputation by Chained Equations in R}},
  journal = {Journal of Statistical Software},
  volume  = {45},
  number  = {3},
  year    = {2011},
  pages   = {1--67},
  doi     = {10.18637/jss.v045.i03}}

@article{MissForest,
  author  = {Stekhoven, Daniel J. and Bühlmann, Peter},
  title   = {{MissForest--non-parametric missing value imputation for mixed-type data}},
  journal = {Bioinformatics},
  volume  = {28},
  number  = {1},
  pages   = {112--118},
  year    = {2011},
  issn    = {1367-4803},
  doi     = {10.1093/bioinformatics/btr597}}

@article{scikit-learn,
  author = {Pedregosa, Fabian
    and Varoquaux, Ga{\"e}l
    and Gramfort, Alexandre
    and Michel, Vincent
    and Thirion, Bertrand
    and Grisel, Olivier
    and Blondel, Mathieu
    and Prettenhofer, Peter
    and Weiss, Ron
    and Dubourg, Vincent
    and Vanderplas, Jake
    and Passos, Alexandre
    and Cournapeau, David
    and Brucher, Matthieu
    and Perrot, Matthieu
    and Duchesnay, {\'E}douard},
  title   = {{Scikit-learn: Machine Learning in Python}},
  journal = {Journal of Machine Learning Research},
  year    = {2011},
  volume  = {12},
  number  = {85},
  pages   = {2825--2830},
  url     = {http://jmlr.org/papers/v12/pedregosa11a.html}}

@article{SSIM,
  author  = {Wang, Zhou and Bovik, Alan Conrad and Sheikh, Hamid Rahim and Simoncelli, Eero P.},
  title   = {{Image quality assessment: from error visibility to structural similarity}},
  journal = {IEEE Transactions on Image Processing},
  year    = {2004},
  volume  = {13},
  number  = {4},
  pages   = {600--612},
  issn    = {1941-0042},
  doi     = {10.1109/TIP.2003.819861}}

@article{KNNimpute,
  author = {Troyanskaya, Olga
    and Cantor, Michael
    and Sherlock, Gavin
    and Brown, Pat
    and Hastie, Trevor
    and Tibshirani, Robert
    and Botstein, David
    and Altman, Russ B.},
  title   = {{Missing value estimation methods for DNA microarrays}},
  journal = {Bioinformatics},
  volume  = {17},
  number  = {6},
  pages   = {520--525},
  year    = {2001},
  doi     = {10.1093/bioinformatics/17.6.520}}

@inproceedings{MFI,
 author    = {Lee, Daniel D. and Seung, H. Sebastian},
 title     = {{Algorithms for Non-negative Matrix Factorization}},
 year      = {2000},
 booktitle = {{Advances in Neural Information Processing Systems}},
 editor    = {Leen, Todd K. and Dietterich, Thomas G. and Tresp, Volker},
 publisher = {MIT Press},
 volume    = {13},
 pages     = {556--562}}

@misc{CMUFaceImages,
  author = {Mitchell, Tom},
  title  = {{CMU Face Images}},
  year   = {1997},
  howpublished = {UCI Machine Learning Repository},
  doi          = {10.24432/C5JC79}}

@misc{MNIST,
  author = {LeCun, Yann and Cortes, Corinna  and Burges, Christopher J.C.},
  title  = {{The MNIST database of handwritten digits}},
  year   = {1994},
  url    = {http://yann.lecun.com/exdb/mnist/}}

\appendix

\renewcommand{\thefigure}{\Alph{section}.\arabic{figure}}
\counterwithin*{figure}{section}

\renewcommand{\thetable}{\Alph{section}.\arabic{table}}
\counterwithin*{table}{section}

\section{Algorithm}\label{sec:algorithm}

\begin{algorithm}[H]
\caption{FILLER}\label{alg:imputation}
\begin{algorithmic}[1]
\Input A corrupted test data sample \(\bm{x}^\mathrm{[test]}_{k} \odot \bm{m}_{k}\), two-dimensional latent embedding \(\bm{X}^\mathrm{[lat]}\), \(\operatorname{DECODER}\) of a trained G-NeuroDAVIS, hyperparameter \(p\), and convergence parameter \(\epsilon\).
\Output \(\hat{\bm{x}}^\mathrm{[test]}_{k}\), the imputed version of \(\bm{x}^\mathrm{[test]}_{k} \odot \bm{m}_{k}\).
\State \(t \gets 1\)
\State Initialize \(x_{\min}^{(t)}\), \(x_{\max}^{(t)}\), \(y_{\min}^{(t)}\), and \(y_{\max}^{(t)}\) using equation~\eqref{eq:bbox-init}.
\Repeat
  \State Consider a bounding area
  \[
  \mathsf{BBOX}^{(t)} \gets [x_{\min}^{(t)}, x_{\max}^{(t)}] \times [y_{\min}^{(t)}, y_{\max}^{(t)}]
  \]
  \If{\(t = 1\)}
    \State A set \(\Lambda^{(t)}\) of \(p\) points is uniformly sampled from \(\mathsf{BBOX}^{(t)}\), i.e.,
    \[
    \Lambda^{(t)} \gets \{\bm{\alpha}_{1}^{(t)}, \bm{\alpha}_{2}^{(t)}, \dots, \bm{\alpha}_{p}^{(t)} \}.
    \]
  \Else
    \State A set \(\Lambda^{(t)}\) of \((p-1)\) points is uniformly sampled from \(\mathsf{BBOX}^{(t)}\), and \(\bm{\alpha}^{(t-1)}_{*}\) is prepended to the set, i.e.,
    \[
    \Lambda^{(t)} \gets \{\bm{\alpha}^{(t-1)}_{*}, \bm{\alpha}^{(t)}_{1}, \bm{\alpha}^{(t)}_{2}, \dots, \bm{\alpha}^{(t)}_{p-1}\}.
    \]
  \EndIf
  \State Find
  \(
  \bm{\alpha}^{(t)}_{*} = \underset{\bm{\alpha} \in \Lambda^{(t)}}{\arg\min}\: \lVert
    \bm{x}^\mathrm{[test]}_{k} \odot \bm{m}_{k} -
    \operatorname{DECODER}(\bm{\alpha}) \odot \bm{m}_{k} \rVert_2^2
  \)
  \State \(t \gets t + 1\)
  \State Update \(x_{\min}^{(t)}\), \(x_{\max}^{(t)}\), \(y_{\min}^{(t)}\), and \(y_{\max}^{(t)}\) using equation~\eqref{eq:bbox-update_t}.
  \Until{%
  \(
  \lVert \bm{x}^\mathrm{[test]}_{k} \odot \bm{m}_{k} -
    \operatorname{DECODER}(\bm{\alpha}^{(t)}_{*}) \odot \bm{m}_{k} \rVert_2^2 < \epsilon
  \)} 
\State
\(
\hat{\bm{x}}^\mathrm{[test]}_{k} \gets
  \bm{x}^\mathrm{[test]}_{k} \odot \bm{m}_{k} +
  \operatorname{DECODER}(\bm{\alpha}^{(t-1)}_{*}) \odot (1 - \bm{m}_{k})
\)
\State \Return \(\hat{\bm{x}}^\mathrm{[test]}_{k}\)
\end{algorithmic}\end{algorithm}


In order to analyze the time complexity of the proposed algorithm, let the number of training samples, features, points uniformly sampled from a bounding box, and iterations be \(n\), \(d\), \(p\), and \(c\), respectively. The initialization step has a complexity of \(\mathcal{O}(n)\), as the calculation of the maximum and the minimum from an unsorted array demands the same. After initialization, the iterative search starts. Creating a bounding box and drawing samples from it has a complexity of \(\mathcal{O}(1)\) and \(\mathcal{O}(p)\), respectively. These samples are then passed through the trained decoder to obtain \(d\)-dimensional reconstructions. Let the decoding has a complexity of \(\mathcal{O}(w)\). Therefore, the calculation of the best latent space sample requires a complexity of \(\mathcal{O}(p \cdot w \cdot d)\), as it finds the minimum of \(p\) numbers, each of which comes from a \(d\)-dimensional vector. Finally, the update step has a complexity of \(\mathcal{O}(1)\). After convergence, to impute the corrupted sample an additional \(\mathcal{O}(d)\) complexity is present.

Therefore, the total time complexity of FILLER is
\[
\mathcal{O}(n) + c \cdot [ \mathcal{O}(1) + \mathcal{O}(p) + \mathcal{O}(p \cdot w \cdot d) + \mathcal{O}(1)] + \mathcal{O}(d)
= \mathcal{O}(n + c \cdot p \cdot w \cdot d).
\]

FILLER consists of two hyperparameters, viz., \(c\) and \(p\). The values of these hyperparameters have been set empirically. In the present study, \(c = 25\) and \(p = 256\) have been considered.

\section{Proof of convergence}\label{sec:proof}

\begin{sidewaysfigure}\centering
\begin{subfigure}{0.375\linewidth}\centering
\includegraphics[width=\linewidth]{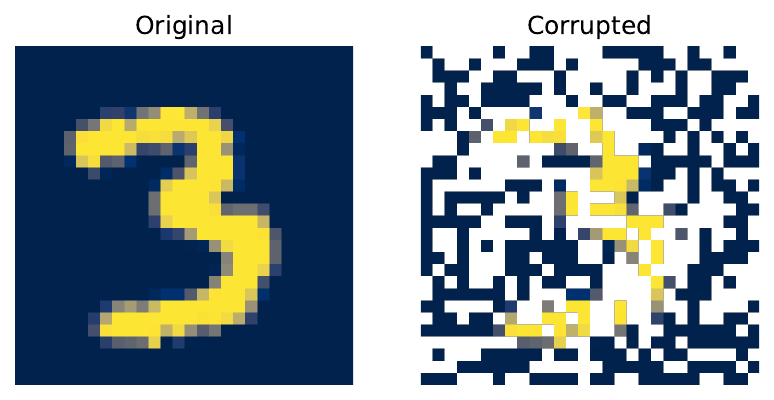}
\caption{The original and the corrupted image (50\% missing)}
\label{fig:MNIST-search-target}
\end{subfigure}

\begin{subfigure}{0.1875\linewidth}\centering
\includegraphics[width=\linewidth]{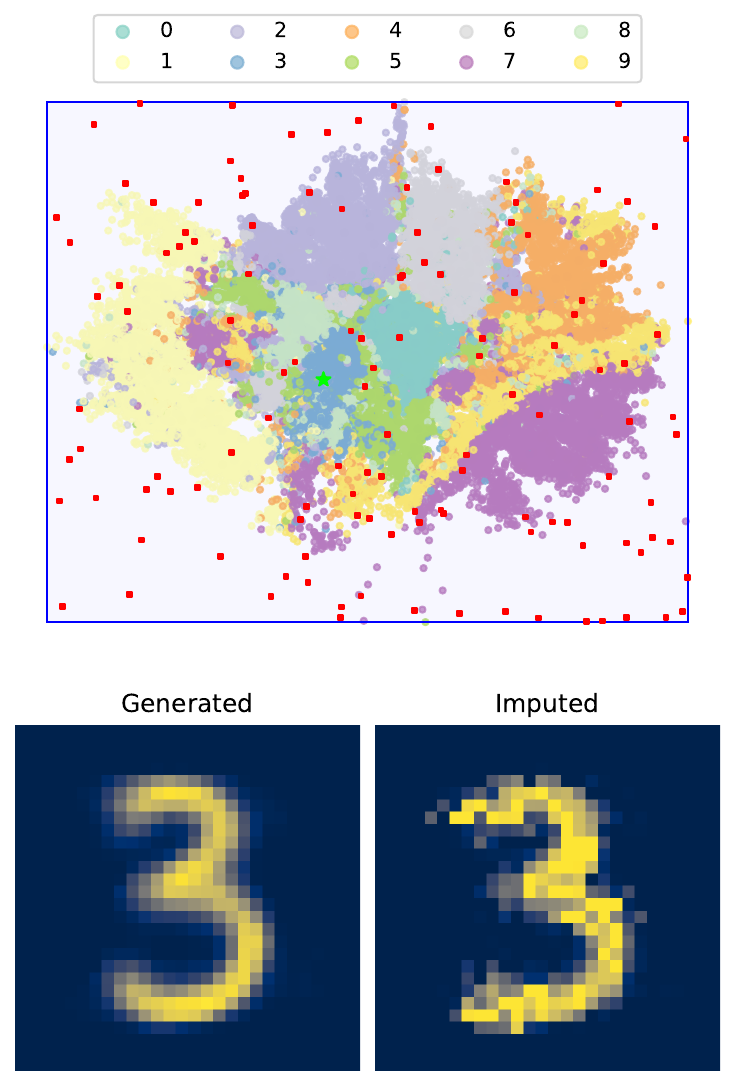}
\caption{Step 1}\label{fig:MNIST-search-0}
\end{subfigure}
\begin{subfigure}{0.1875\linewidth}\centering
\includegraphics[width=\linewidth]{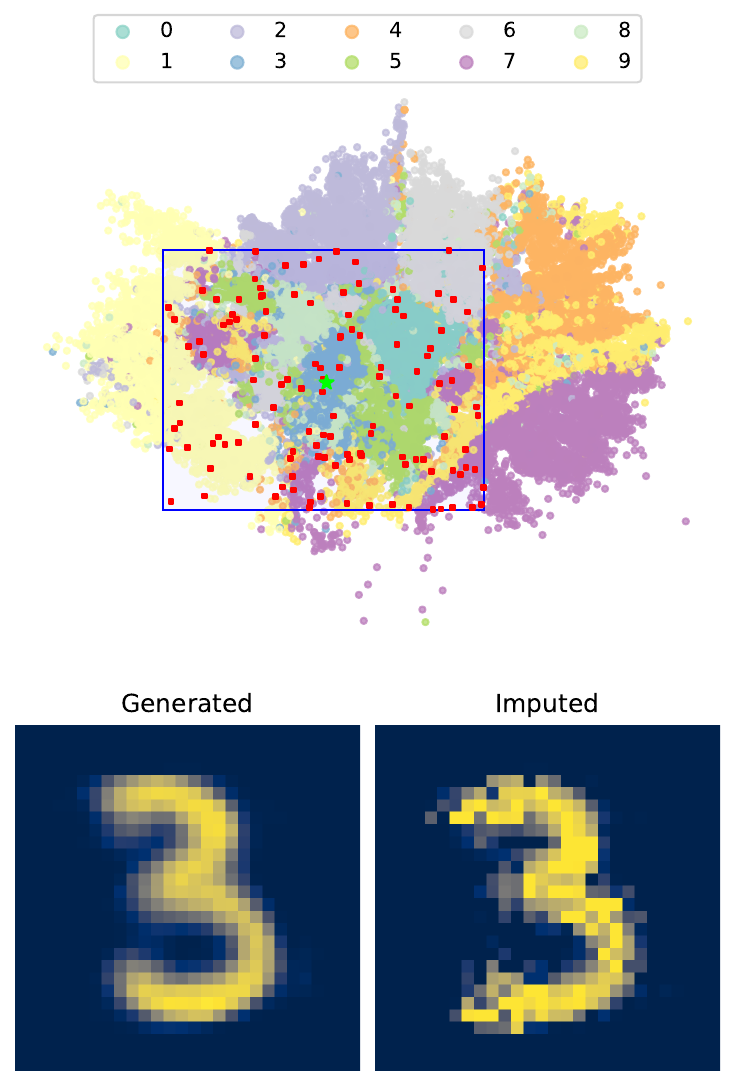}
\caption{Step 2}\label{fig:MNIST-search-1}
\end{subfigure}
\begin{subfigure}{0.1875\linewidth}\centering
\includegraphics[width=\linewidth]{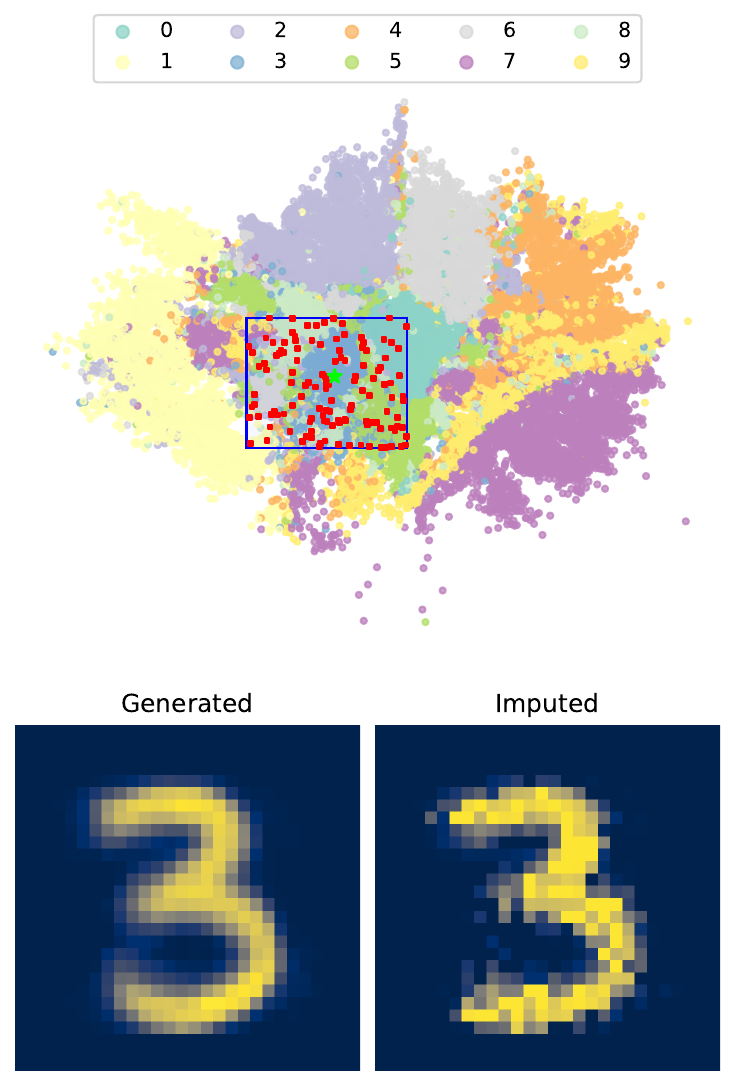}
\caption{Step 3}\label{fig:MNIST-search-2}
\end{subfigure}
\begin{subfigure}{0.1875\linewidth}\centering
\includegraphics[width=\linewidth]{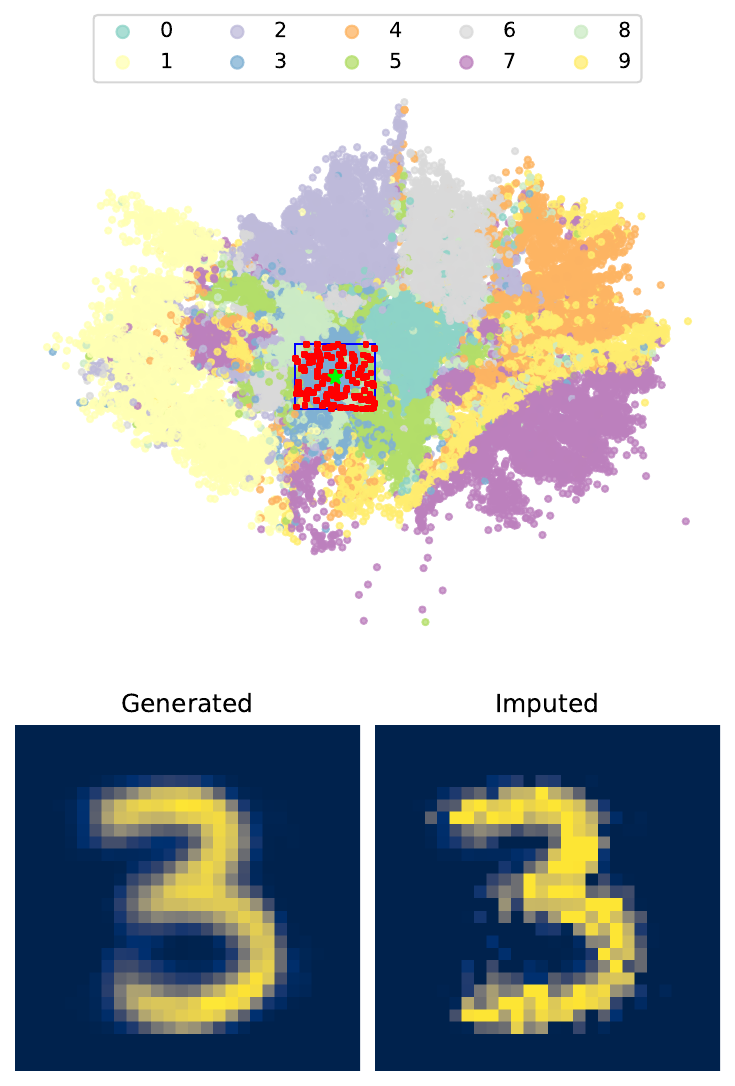}
\caption{Step 4}\label{fig:MNIST-search-3}
\end{subfigure}
\begin{subfigure}{0.1875\linewidth}\centering
\includegraphics[width=\linewidth]{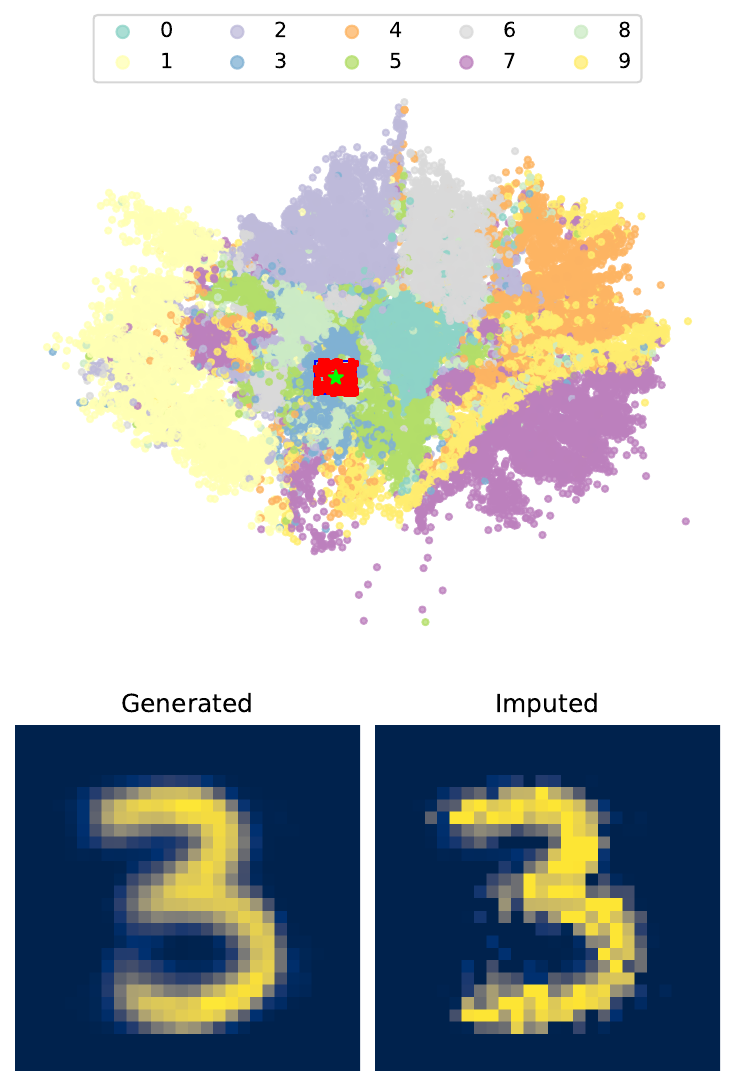}
\caption{Step 5}\label{fig:MNIST-search-4}
\end{subfigure}
\caption{A graphical representation of the FILLER method. Subfigure~\subref{fig:MNIST-search-target} represents an \(i\)\textsuperscript{th} test sample (left) and its corrupted version that needs to be imputed (right). Subfigures \subref{fig:MNIST-search-0} to \subref{fig:MNIST-search-4} demonstrate the iterative steps. The plot on the top in each step shows the embeddings produced by G-NeuroDAVIS, with a blue rectangle showing a bounding box, and the points in red are uniformly sampled from that bounding box. These points pass through the decoder, and the best match is shown as a green star. Below these, the plot on the left shows the decoded or generated image corresponding to the green star, and the plot on the right reflects the quality of imputation at the mentioned step.}
\label{fig:MNIST-search}
\end{sidewaysfigure}

In this section, a mathematical formulation of the convergence of the proposed imputation method has been discussed. The proposed method takes a corrupted image as input and searches for a representative in the generalized two-dimensional latent space obtained from G-NeuroDAVIS. The proposed method does the same by observing several reconstructions from the latent space to the original space. In this section, it is shown that the search will eventually converge to a unique point in the latent space.

As discussed in Algorithm~\ref{alg:imputation}, initially a bounded box \(\mathsf{BBOX}^{(1)}\) has been considered. A set of \(p\) points is uniformly sampled from this bounded box, and based on the reconstructions, the best point from the set is selected. Using this best point, the bounded box is updated and \(\mathsf{BBOX}^{(2)}\) is produced. This process is repeated multiple times to find the optimal representative corresponding to the corrupted input in the latent space. The construction of any of the bounded boxes, viz., \(\mathsf{BBOX}^{(t)}\), has some nice mathematical properties as follows:
\begin{itemize}
  \item Each \(\mathsf{BBOX}^{(t)}\) is a Cartesian product of two closed and bounded intervals and thus compact by construction.
  \item \(\mathsf{BBOX}^{(t+1)} \subset \mathsf{BBOX}^{(t)}\ \forall\ t \in \mathbb{N} \)
  \item \(\lim_{t \to \infty} \operatorname{Diam}(\mathsf{BBOX}^{(t)}) = 0\), where
  \(
  \operatorname{Diam}(S) = \sup\{\lVert \bm{x}_1 - \bm{x}_2 \rVert \mid \bm{x}_1, \bm{x}_2 \in S\}
  \).
\end{itemize}
Thus, by Cantor's Intersection theorem, there exists a unique point \(\bm{c} \in \mathbb{R}^{2}\) such that
\[
\bigcap_{t = 1}^{\infty} \mathsf{BBOX}^{(t)} = \{\bm{c}\}.
\]
This proves that the proposed method will converge to a unique point in the latent space.

\section{Imputation results}

\begin{figure}[H]\centering
\begin{subfigure}{0.285\linewidth}\centering
\begin{tikzpicture}
  \useasboundingbox (0,0) rectangle (\linewidth,0.75\linewidth);
  \node[anchor=south west, inner sep=0] (img)
    at (0,0) {\includegraphics[width=\linewidth]{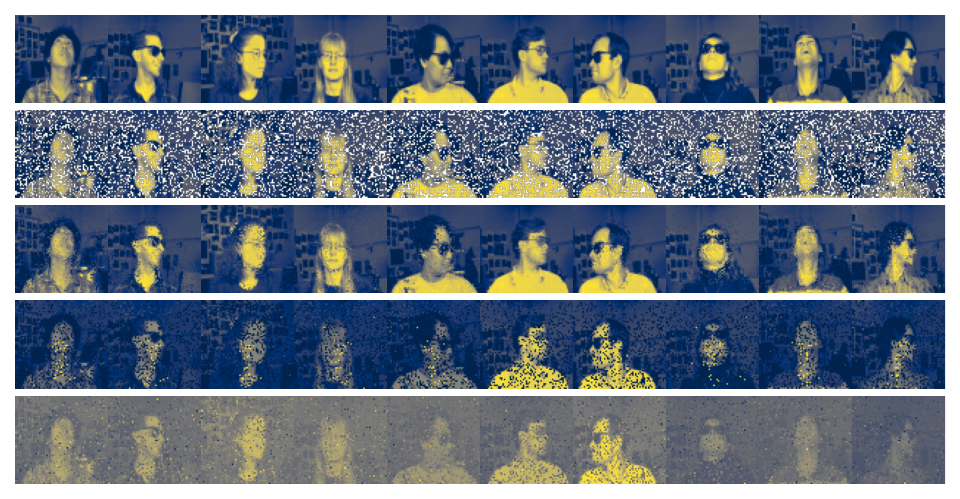}};
  \begin{scope}[x={(img.south east)}, y={(img.north west)}]
    \bfseries\scriptsize
    \node[left] at (0, 0.9) {Original};
    \node[left] at (0, 0.7) {Corrupted};
    \node[left] at (0, 0.5) {FILLER};
    \node[left] at (0, 0.3) {GAIN};
    \node[left] at (0, 0.1) {MIWAE};
  \end{scope}
\end{tikzpicture}
\caption{Random Missingness}\label{fig:showcase-Faces2-random}
\end{subfigure}
\begin{subfigure}{0.285\linewidth}\centering
\includegraphics[width=\linewidth]{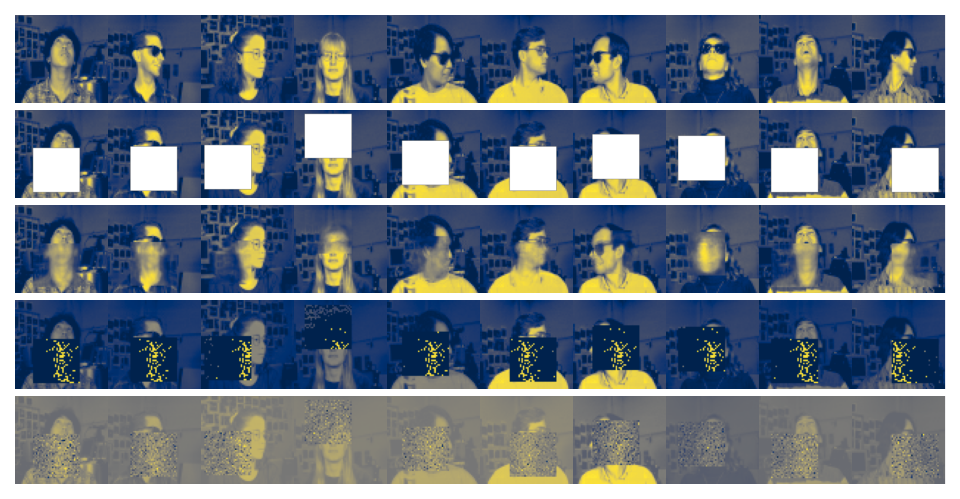}
\caption{Block-wise Missingness}\label{fig:showcase-Faces2-block}
\end{subfigure}
\begin{subfigure}{0.285\linewidth}\centering
\includegraphics[width=\linewidth]{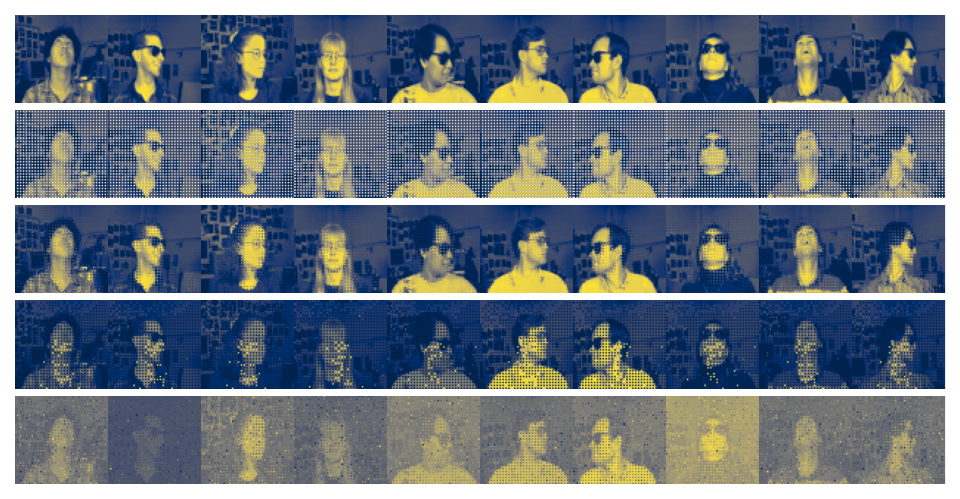}
\caption{Grid-wise Missingness}\label{fig:showcase-Faces2-grid}
\end{subfigure}

\begin{subfigure}{0.285\linewidth}\centering
\begin{tikzpicture}
  \useasboundingbox (0,0) rectangle (\linewidth,0.75\linewidth);
  \node[anchor=south west, inner sep=0] (img)
    at (0,0) {\includegraphics[width=\linewidth]{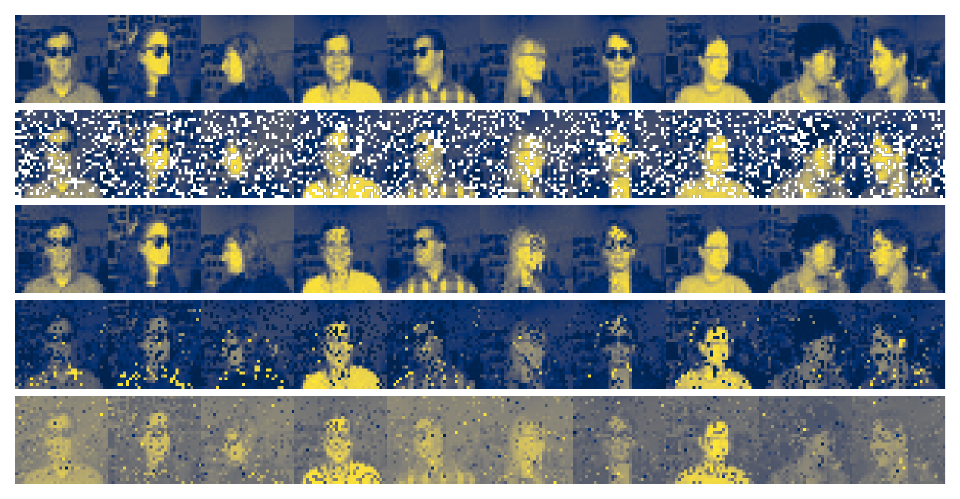}};
  \begin{scope}[x={(img.south east)}, y={(img.north west)}]
    \bfseries\scriptsize
    \node[left] at (0, 0.9) {Original};
    \node[left] at (0, 0.7) {Corrupted};
    \node[left] at (0, 0.5) {FILLER};
    \node[left] at (0, 0.3) {GAIN};
    \node[left] at (0, 0.1) {MIWAE};
  \end{scope}
\end{tikzpicture}
\caption{Random Missingness}\label{fig:showcase-Faces4-random}
\end{subfigure}
\begin{subfigure}{0.285\linewidth}\centering
\includegraphics[width=\linewidth]{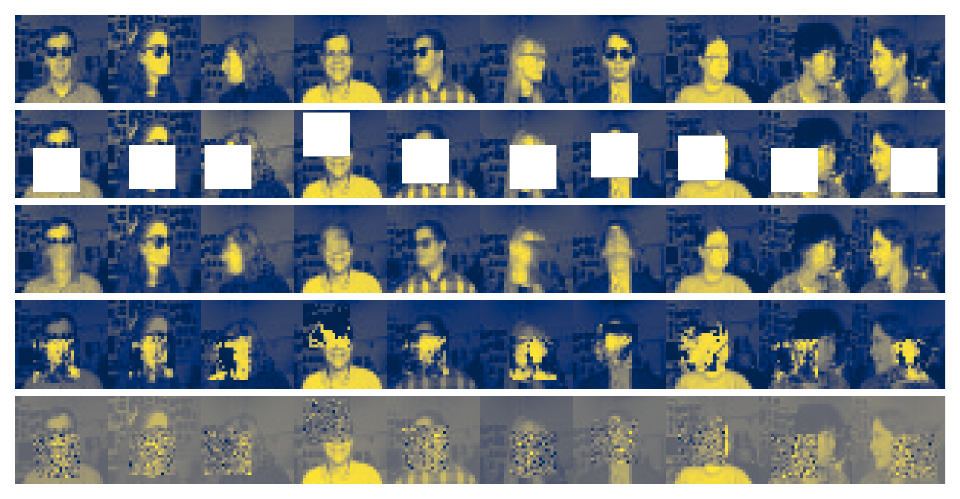}
\caption{Block-wise Missingness}\label{fig:showcase-Faces4-block}
\end{subfigure}
\begin{subfigure}{0.285\linewidth}\centering
\includegraphics[width=\linewidth]{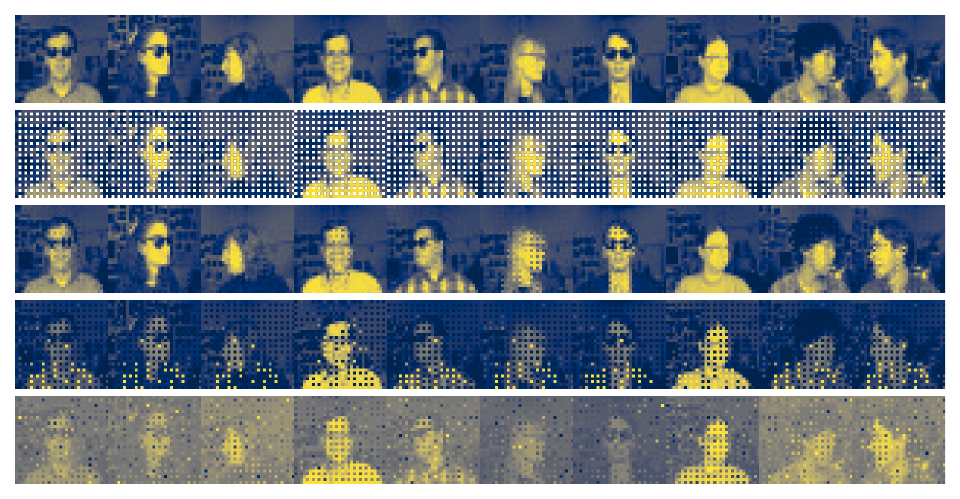}
\caption{Grid-wise Missingness}\label{fig:showcase-Faces4-grid}
\end{subfigure}
\caption{A qualitative comparison of various imputation methods. The two rows illustrate imputation examples from the half-resolution and the quarter-resolution versions of the \textit{CMU Face Images} dataset respectively. In each column, different types of missingness (25\%) is demonstrated. Each subfigure contains samples from different classes in the dataset.}
\label{fig:showcase-Faces2-Faces4}
\end{figure}
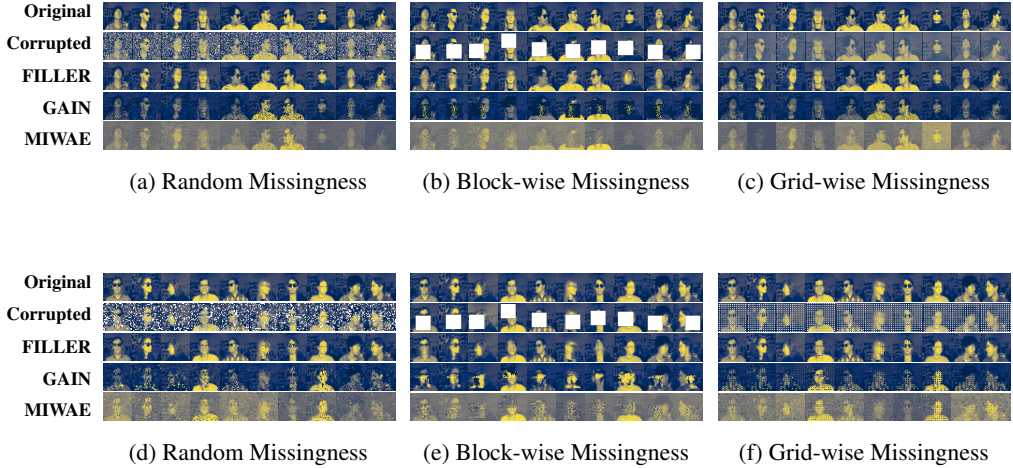

\begin{table}[H]\centering
\caption{Number of Parameters}\label{tab:parameters}
\begin{tabular}{lrrr}\toprule
\textbf{Dataset}                            & \textbf{G-NeuroDAVIS} & \textbf{GAIN} & \textbf{MIWAE} \\ \midrule
\textit{CMU Face Images} \((120\times128)\) &                 3.9 M &         1.8 B &        107.5 K \\
\textit{CMU Face Images} \(( 60\times 64)\) &                 1.0 M &       117.9 M &         26.8 K \\
\textit{CMU Face Images} \(( 30\times 32)\) &               264.7 K &         7.3 M &          6.7 K \\
\textit{Fashion-MNIST}                      &               338.3 K &         4.9 M &          5.4 K \\
\textit{MNIST}                              &               338.3 K &         4.9 M &          5.4 K \\ \bottomrule
\end{tabular}\end{table}

\begin{figure}[p]\centering 
\includegraphics[width=\linewidth]{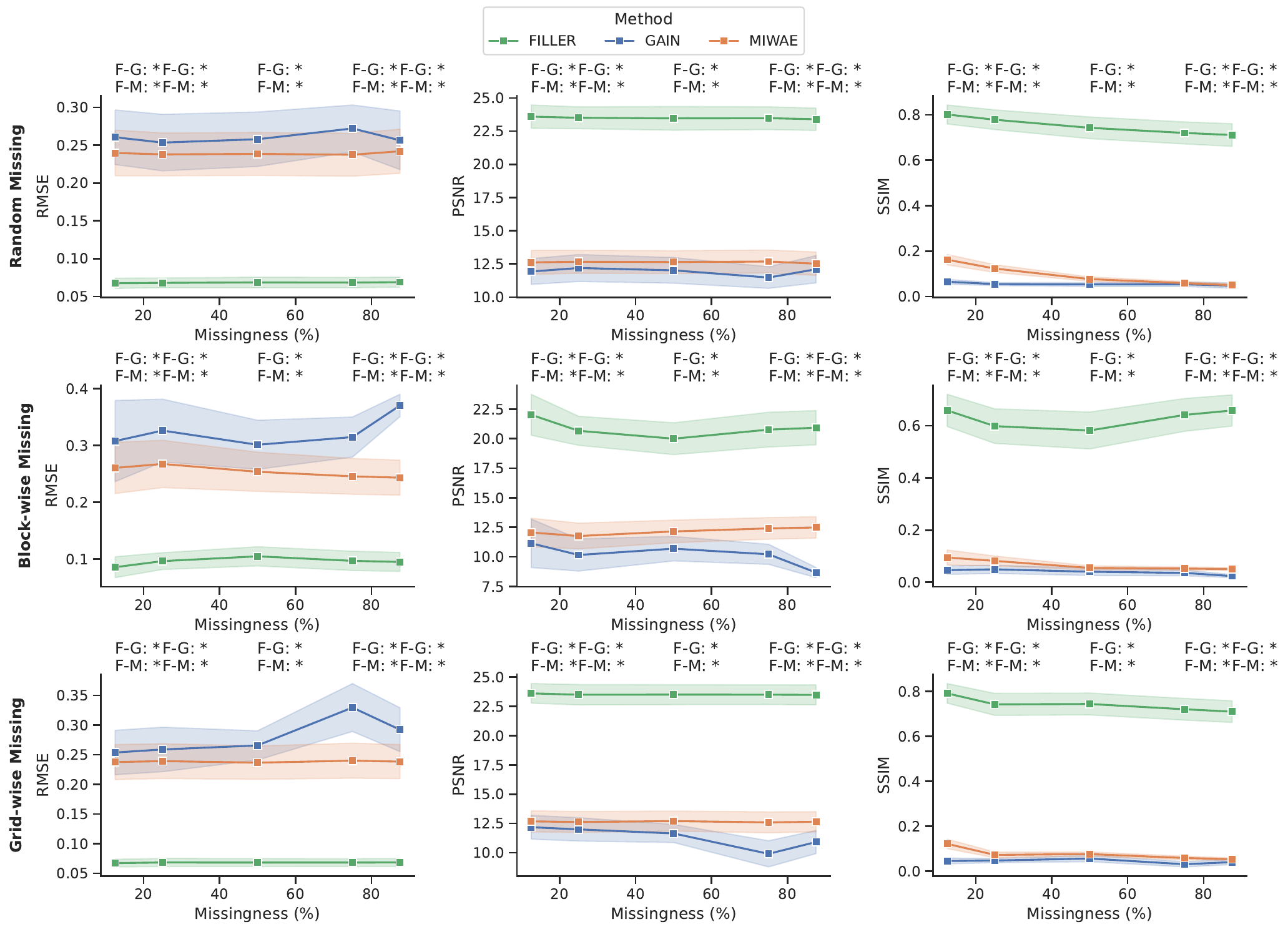}
\caption{Comparison of several imputation methods based on RMSE, PSNR, and SSIM on the half-resolution \textit{CMU Face Images} with varying levels of missingness. Statistical significance has been measured with a Wilcoxon signed-rank test. F-G and F-M denote the significance levels of FILLER against GAIN and MIWAE, respectively. \(*\) indicates statistical significance at 0.05 level.}
\label{fig:Faces2-comparison}
\end{figure}

\begin{figure}[p]\centering 
\includegraphics[width=\linewidth]{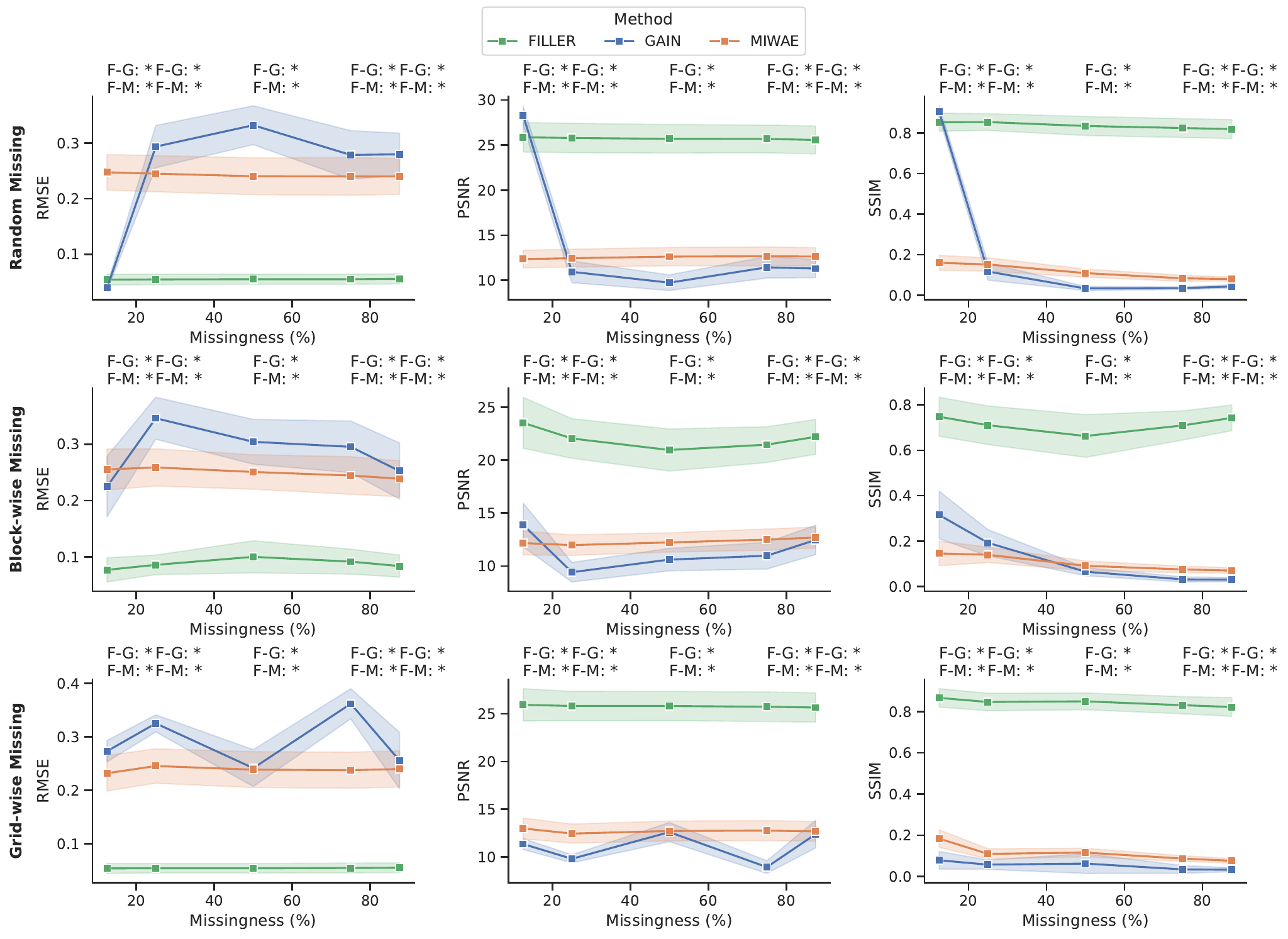}
\caption{Comparison of several imputation methods based on RMSE, PSNR, and SSIM on the quarter-resolution \textit{CMU Face Images} with varying levels of missingness. Statistical significance has been measured with a Wilcoxon signed-rank test. F-G and F-M denote the significance levels of FILLER against GAIN and MIWAE, respectively. \(*\) indicates statistical significance at 0.05 level.}
\label{fig:Faces4-comparison}
\end{figure}

\clearpage

\begin{figure}[t]\centering

\begin{subfigure}{0.285\linewidth}\centering
\begin{tikzpicture}
  \useasboundingbox (0,0) rectangle (\linewidth,0.75\linewidth);
  \node[anchor=south west, inner sep=0] (img)
    at (0,0) {\includegraphics[width=\linewidth]{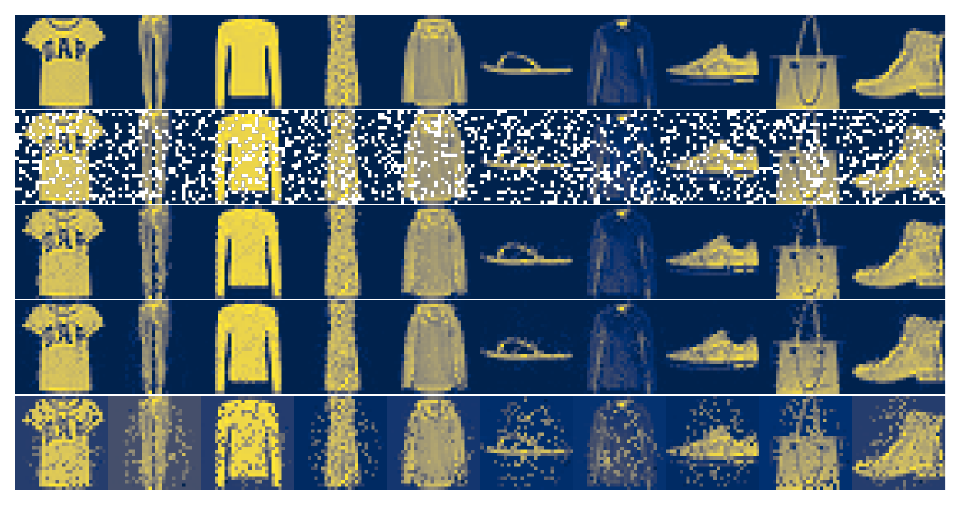}};
  \begin{scope}[x={(img.south east)}, y={(img.north west)}]
    \bfseries\scriptsize
    \node[left] at (0, 0.9) {Original};
    \node[left] at (0, 0.7) {Corrupted};
    \node[left] at (0, 0.5) {FILLER};
    \node[left] at (0, 0.3) {GAIN};
    \node[left] at (0, 0.1) {MIWAE};
  \end{scope}
\end{tikzpicture}
\caption{Random Missingness}\label{fig:showcase-FMNIST-random}
\end{subfigure}
\begin{subfigure}{0.285\linewidth}\centering
\includegraphics[width=\linewidth]{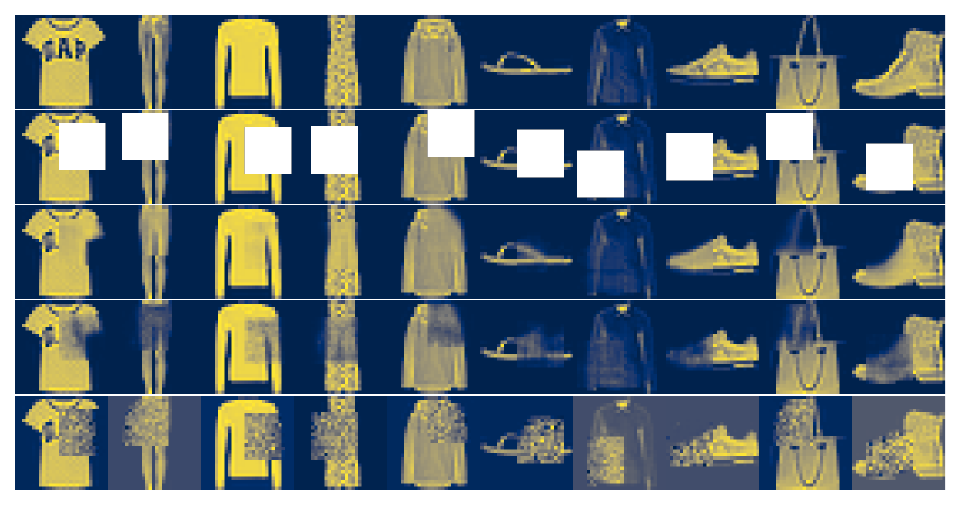}
\caption{Block-wise Missingness}\label{fig:showcase-FMNIST-block}
\end{subfigure}
\begin{subfigure}{0.285\linewidth}\centering
\includegraphics[width=\linewidth]{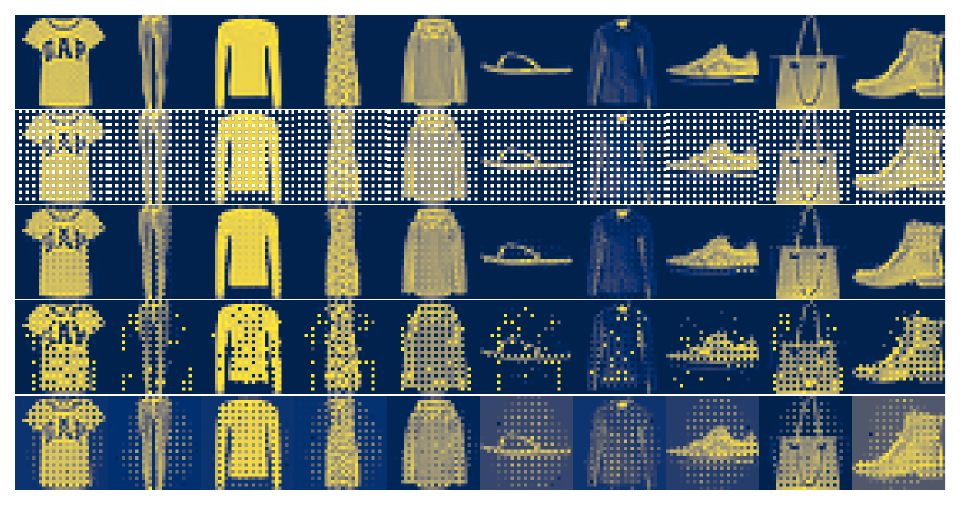}
\caption{Grid-wise Missingness}\label{fig:showcase-FMNIST-grid}
\end{subfigure}
\caption{A qualitative comparison of various imputation methods. The figure illustrates imputation examples from the \textit{Fashion-MNIST} dataset. In each column, different types of missingness (25\%) are demonstrated. Each subfigure contains samples from different classes in the dataset.}
\label{fig:showcase-FMNIST}
\end{figure}

\begin{figure}[ht]\centering 
\includegraphics[width=\linewidth]{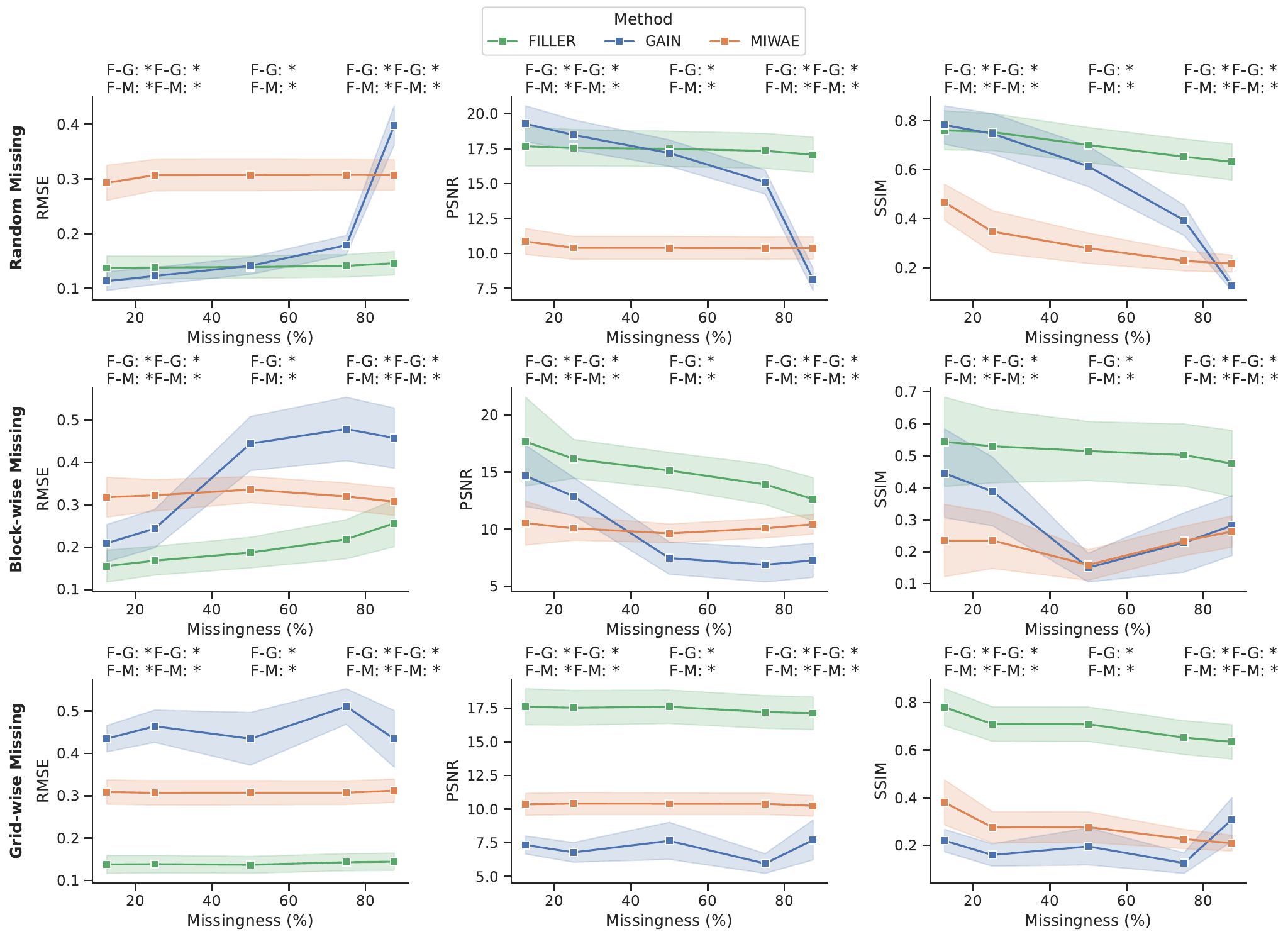}
\caption{Comparison of several imputation methods based on RMSE, PSNR, and SSIM on the \textit{Fashion-MNIST} dataset with varying levels of missingness. Statistical significance has been measured with a Wilcoxon signed-rank test. F-G and F-M denote the significance levels of FILLER against GAIN and MIWAE, respectively. \(*\) indicates statistical significance at 0.05 level.}
\label{fig:FMNIST-comparison}
\end{figure}

\clearpage

\begin{figure}[t]\centering
\begin{subfigure}{0.285\linewidth}\centering
\begin{tikzpicture}
  \useasboundingbox (0,0) rectangle (\linewidth,0.75\linewidth);
  \node[anchor=south west, inner sep=0] (img)
    at (0,0) {\includegraphics[width=\linewidth]{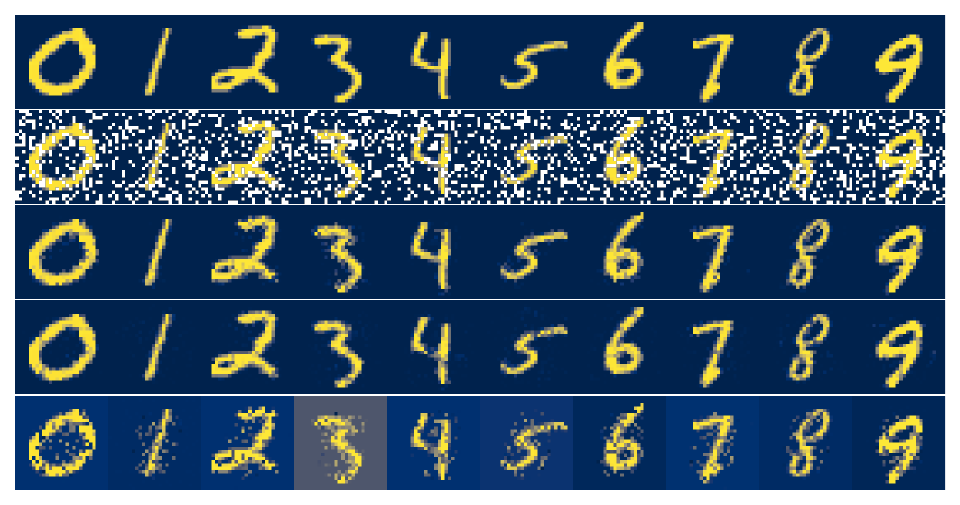}};
  \begin{scope}[x={(img.south east)}, y={(img.north west)}]
    \bfseries\scriptsize
    \node[left] at (0, 0.9) {Original};
    \node[left] at (0, 0.7) {Corrupted};
    \node[left] at (0, 0.5) {FILLER};
    \node[left] at (0, 0.3) {GAIN};
    \node[left] at (0, 0.1) {MIWAE};
  \end{scope}
\end{tikzpicture}
\caption{Random Missingness}\label{fig:showcase-MNIST-random}
\end{subfigure}
\begin{subfigure}{0.285\linewidth}\centering
\includegraphics[width=\linewidth]{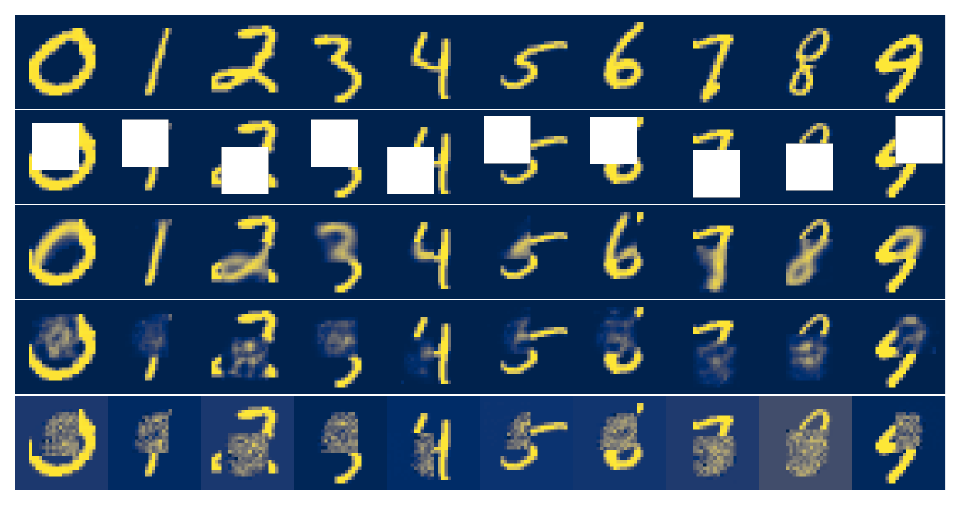}
\caption{Block-wise Missingness}\label{fig:showcase-MNIST-block}
\end{subfigure}
\begin{subfigure}{0.285\linewidth}\centering
\includegraphics[width=\linewidth]{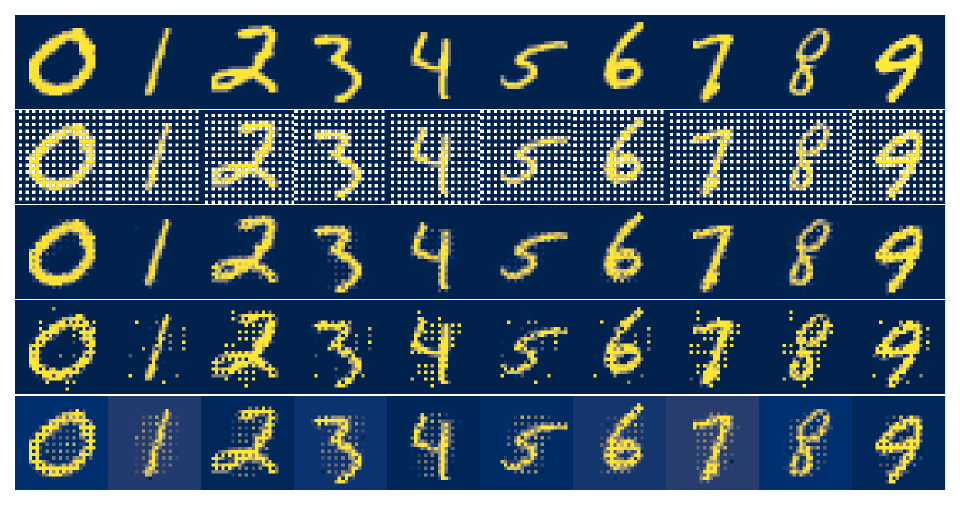}
\caption{Grid-wise Missingness}\label{fig:showcase-MNIST-grid}
\end{subfigure}
\caption{A qualitative comparison of various imputation methods. The figure illustrates imputation examples from the \textit{MNIST} dataset. In each column, different types of missingness (25\%) are demonstrated. Each subfigure contains samples from different classes in the dataset.}
\label{fig:showcase-MNIST}
\end{figure}

\begin{figure}[ht]\centering 
\includegraphics[width=\linewidth]{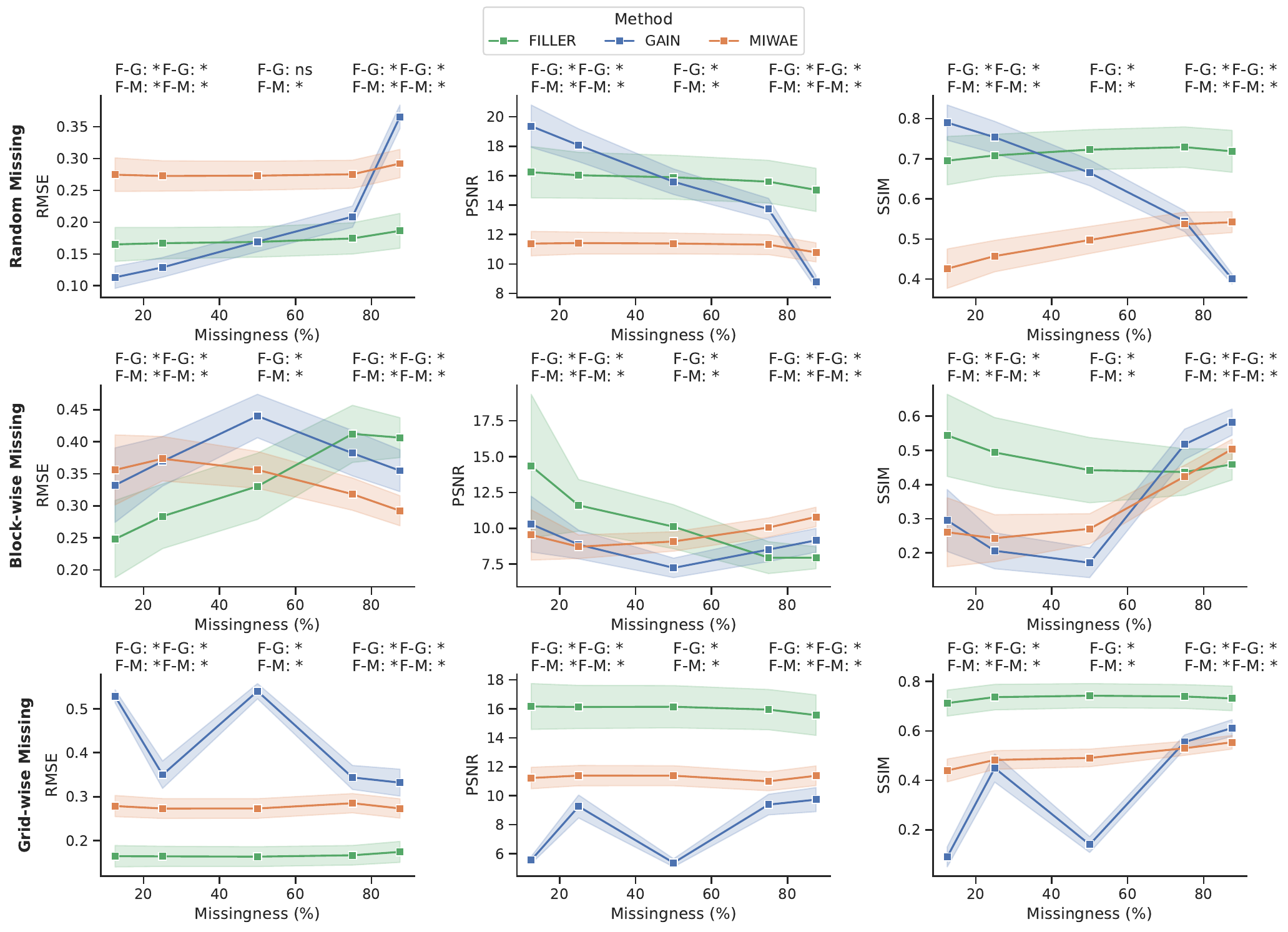}
\caption{Comparison of several imputation methods based on RMSE, PSNR, and SSIM on the \textit{MNIST} dataset with varying levels of missingness. Statistical significance has been measured with a Wilcoxon signed-rank test. F-G and F-M denote the significance levels of FILLER against GAIN and MIWAE, respectively. \(*\) indicates statistical significance at 0.05 level.}
\label{fig:MNIST-comparison}
\end{figure}

\clearpage

\section{Downstream results}


\begin{figure}[H]\centering 
\begin{subfigure}[t]{\linewidth}\centering 
\includegraphics[width=\linewidth]{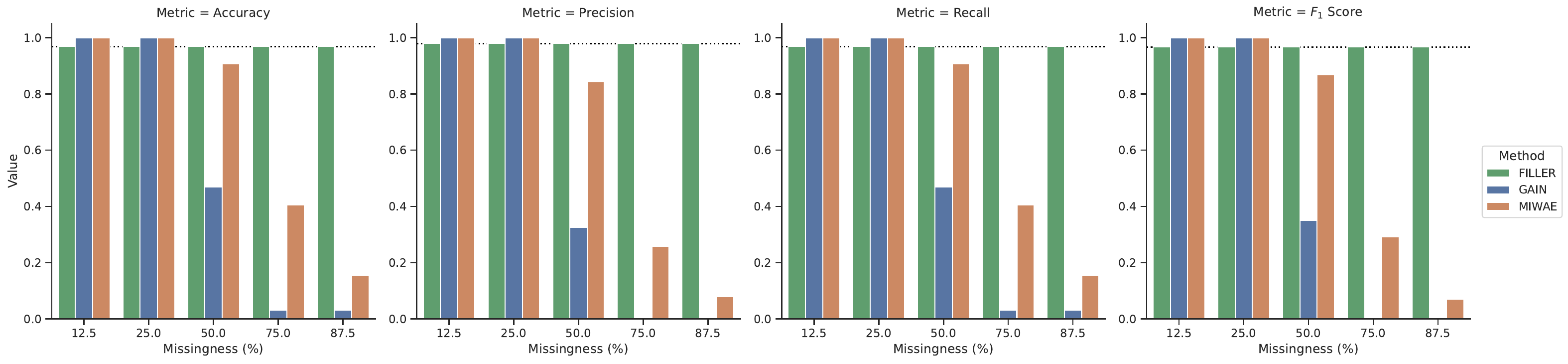}
\caption{Random Missingness}\label{fig:Faces-random-SVC}
\end{subfigure}

\begin{subfigure}[t]{\linewidth}\centering 
\includegraphics[width=\linewidth]{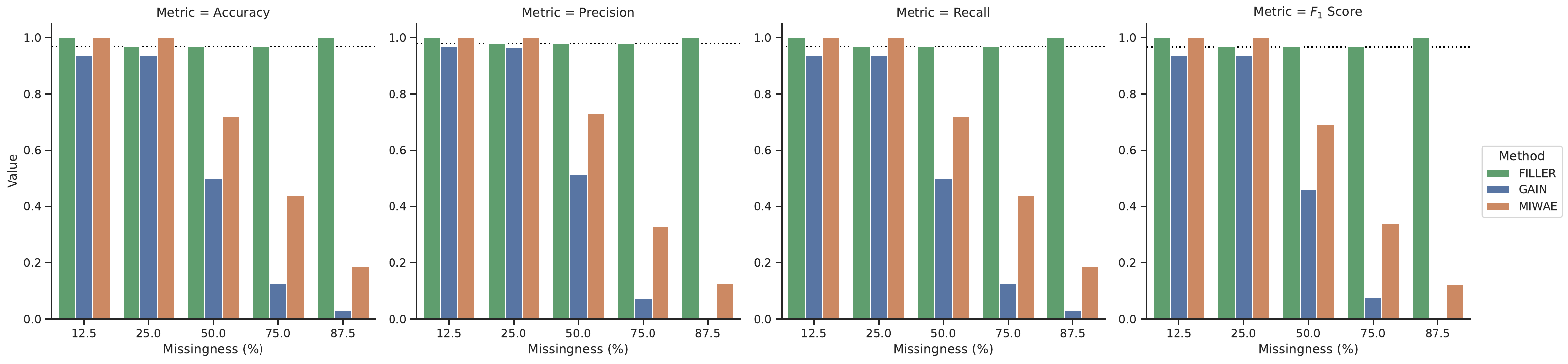}
\caption{Block-wise Missingness}\label{fig:Faces-block-SVC}
\end{subfigure}

\begin{subfigure}[t]{\linewidth}\centering 
\includegraphics[width=\linewidth]{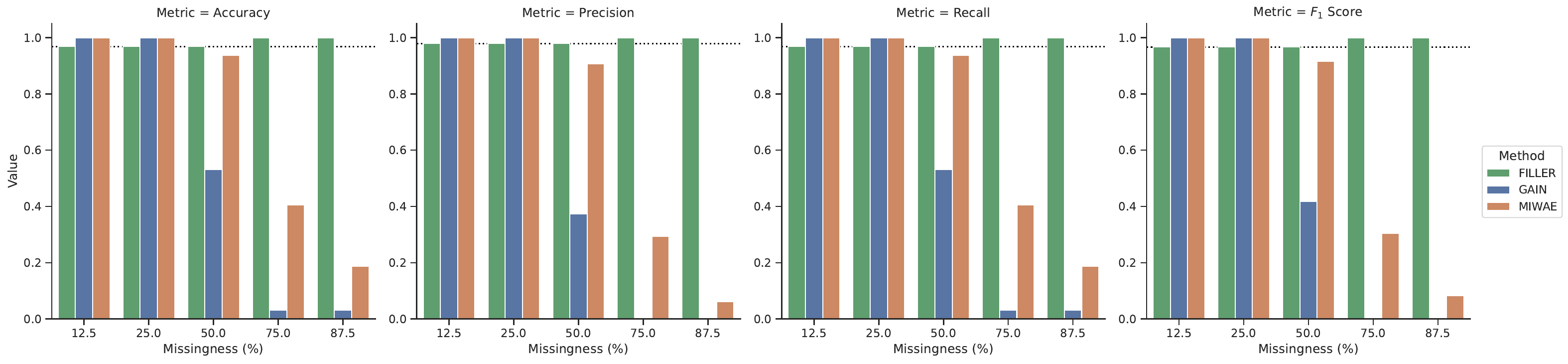}
\caption{Grid-wise Missingness}\label{fig:Faces-grid-SVC}
\end{subfigure}
\caption{Classification results of support vector classifier on the full-resolution \textit{CMU Face Images} dataset with three types of missingness. The dotted lines indicate the classification performance on the original test data.}
\label{fig:Faces-SVC}
\end{figure}

\newpage


\begin{figure}[p]\centering 
\begin{subfigure}[t]{\linewidth}\centering 
\includegraphics[width=\linewidth]{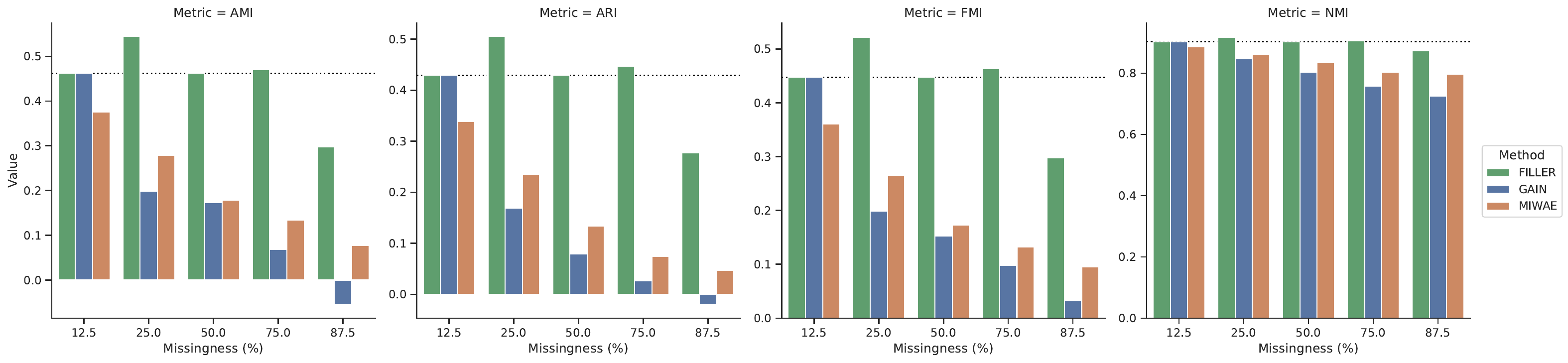}
\caption{Random Missingness}\label{fig:Faces-random-KMC}
\end{subfigure}
\begin{subfigure}[t]{\linewidth}\centering 
\includegraphics[width=\linewidth]{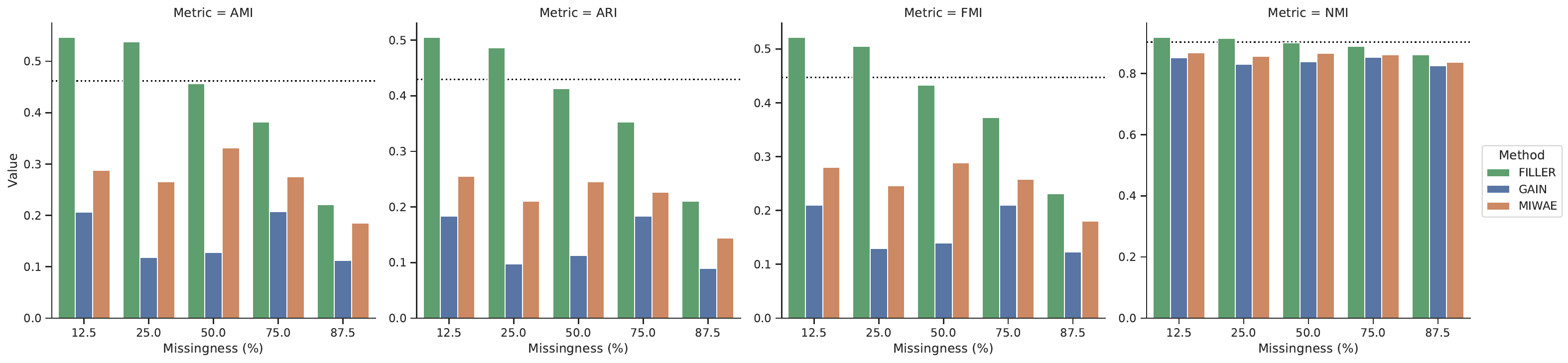}
\caption{Block-wise Missingness}\label{fig:Faces-block-KMC}
\end{subfigure}
\begin{subfigure}[t]{\linewidth}\centering 
\includegraphics[width=\linewidth]{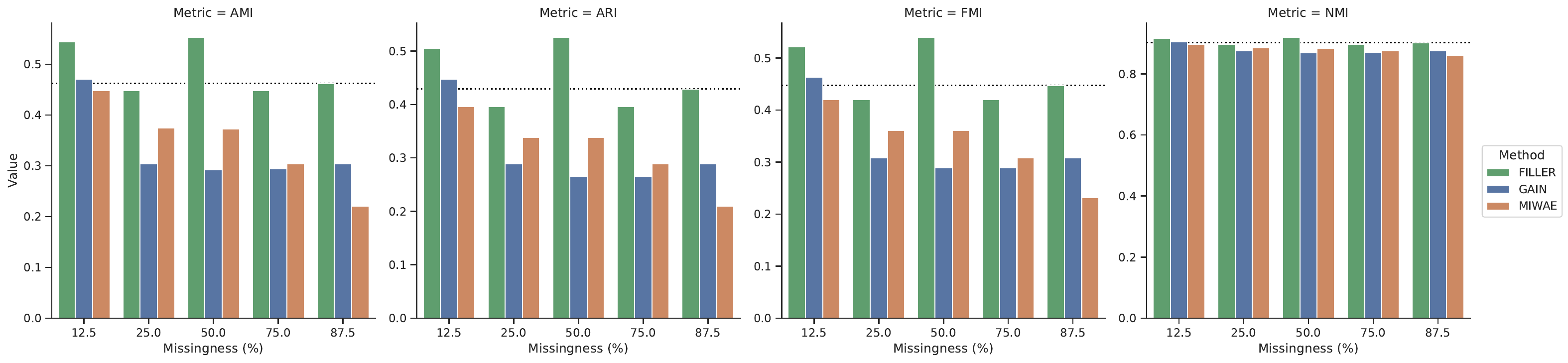}
\caption{Grid-like Missingness}\label{fig:Faces-grid-KMC}
\end{subfigure}
\caption{Clustering results of \(k\)-means clustering on the full-resolution \textit{CMU Face Images} dataset with three types of missingness. The dotted lines indicate the clustering performance on the original test data.}
\label{fig:Faces-KMC}
\end{figure}


\begin{figure}[p]\centering 
\begin{subfigure}[t]{\linewidth}\centering 
\includegraphics[width=\linewidth]{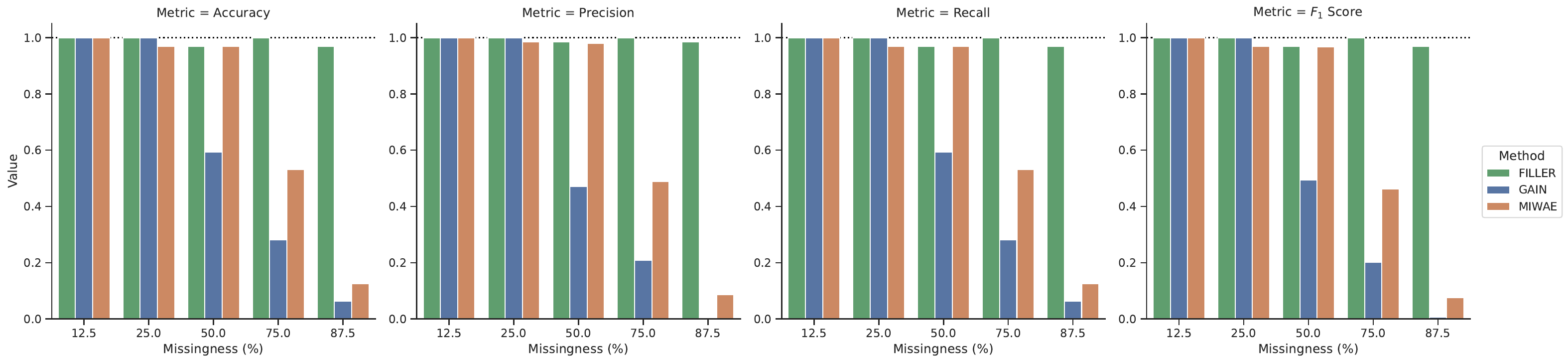}
\caption{Random Missingness}\label{fig:Faces2-random-RFC}
\end{subfigure}

\begin{subfigure}[t]{\linewidth}\centering 
\includegraphics[width=\linewidth]{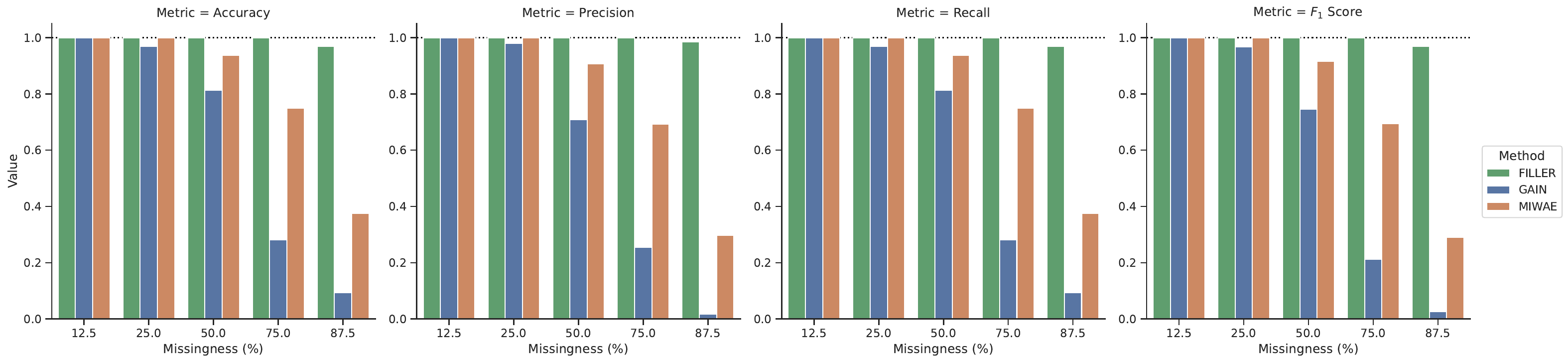}
\caption{Block-wise Missingness}\label{fig:Faces2-block-RFC}
\end{subfigure}

\begin{subfigure}[t]{\linewidth}\centering 
\includegraphics[width=\linewidth]{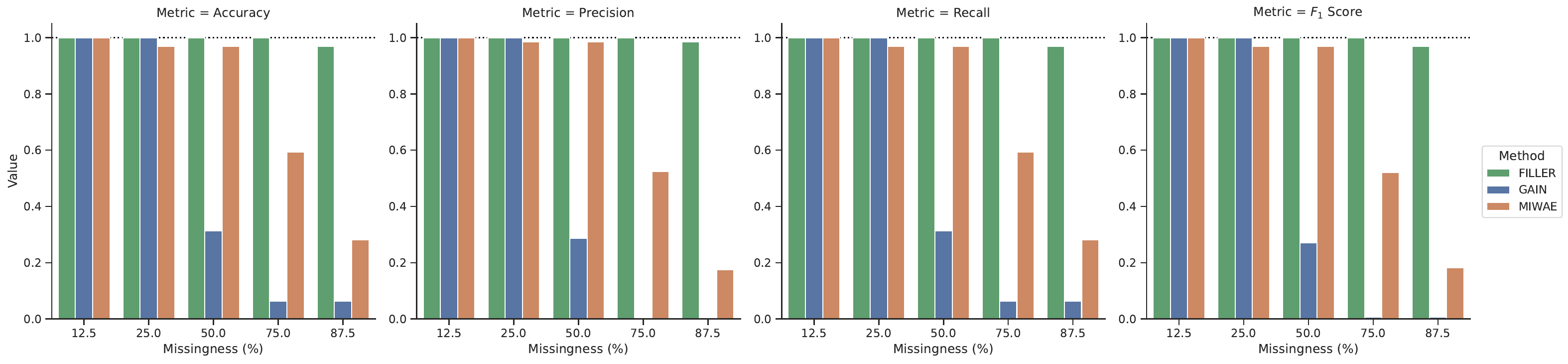}
\caption{Grid-wise Missingness}\label{fig:Faces2-grid-RFC}
\end{subfigure}
\caption{Classification results of random forest classifier on the half-resolution \textit{CMU Face Images} dataset with three types of missingness. The dotted lines indicate the classification performance on the original test data.}
\label{fig:Faces2-RFC}
\end{figure}

\begin{figure}[p]\centering 
\begin{subfigure}[t]{\linewidth}\centering 
\includegraphics[width=\linewidth]{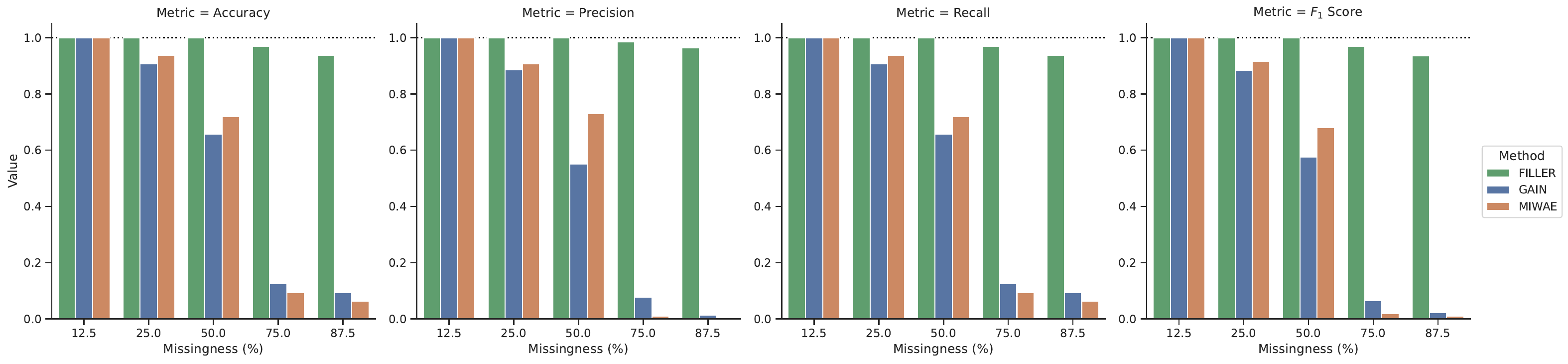}
\caption{Random Missingness}\label{fig:Faces2-random-SVC}
\end{subfigure}

\begin{subfigure}[t]{\linewidth}\centering 
\includegraphics[width=\linewidth]{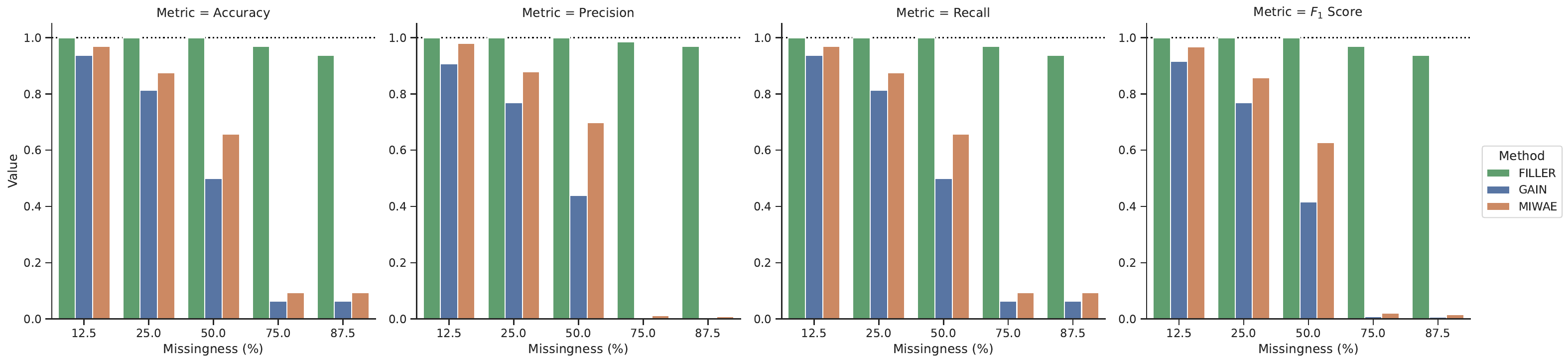}
\caption{Block-wise Missingness}\label{fig:Faces2-block-SVC}
\end{subfigure}

\begin{subfigure}[t]{\linewidth}\centering 
\includegraphics[width=\linewidth]{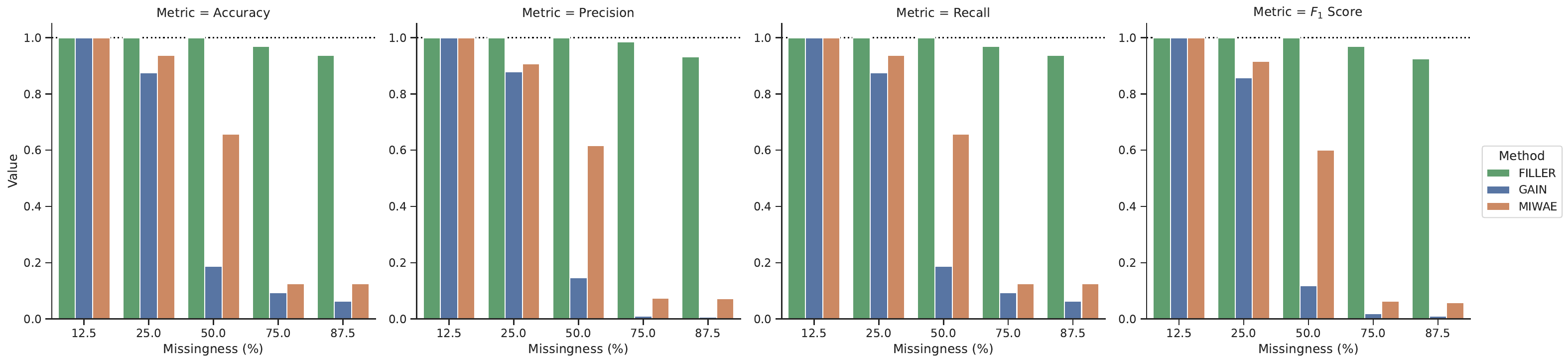}
\caption{Grid-wise Missingness}\label{fig:Faces2-grid-SVC}
\end{subfigure}
\caption{Classification results of support vector classifier on the half-resolution \textit{CMU Face Images} dataset with three types of missingness. The dotted lines indicate the classification performance on the original test data.}
\label{fig:Faces2-SVC}
\end{figure}


\begin{figure}[p]\centering 
\begin{subfigure}[t]{\linewidth}\centering 
\includegraphics[width=\linewidth]{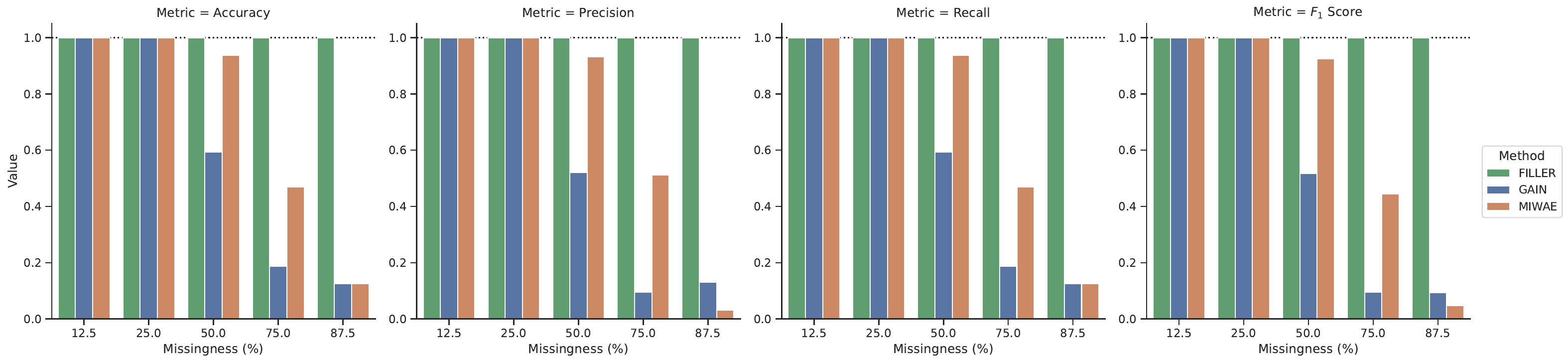}
\caption{Random Missingness}\label{fig:Faces4-random-RFC}
\end{subfigure}

\begin{subfigure}[t]{\linewidth}\centering 
\includegraphics[width=\linewidth]{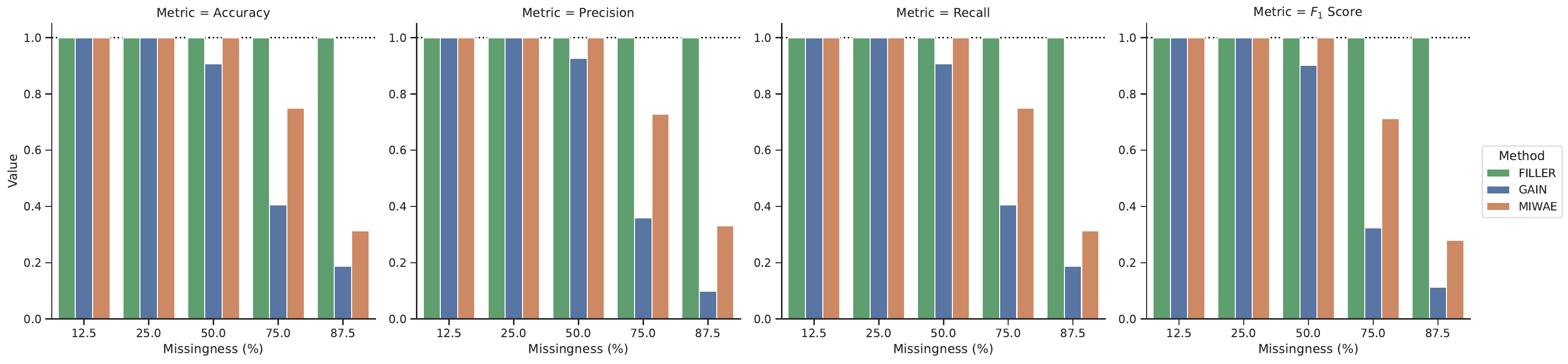}
\caption{Block-wise Missingness}\label{fig:Faces4-block-RFC}
\end{subfigure}

\begin{subfigure}[t]{\linewidth}\centering 
\includegraphics[width=\linewidth]{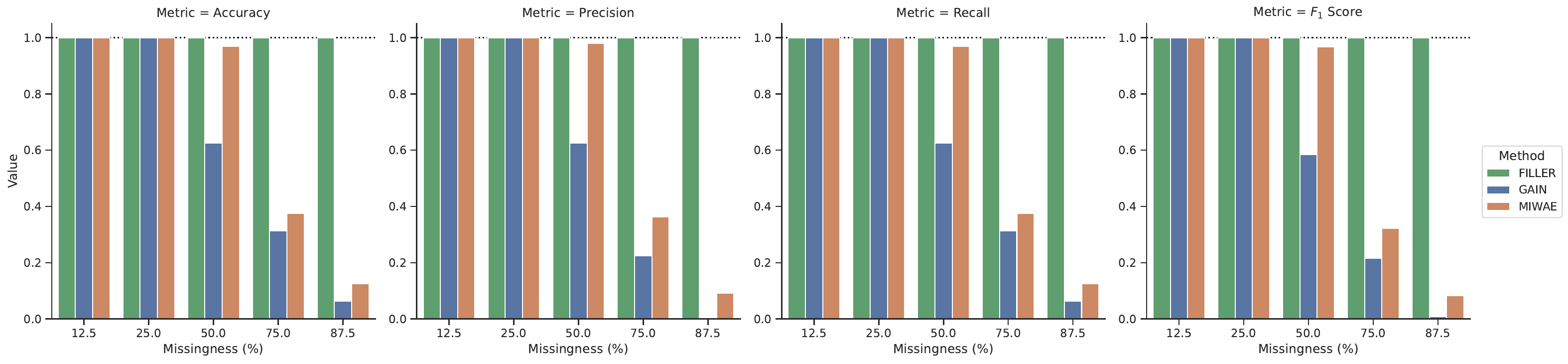}
\caption{Grid-wise Missingness}\label{fig:Faces4-grid-RFC}
\end{subfigure}
\caption{Classification results of random forest classifier on the quarter-resolution \textit{CMU Face Images} dataset with three types of missingness. The dotted lines indicate the classification performance on the original test data.}
\label{fig:Faces4-RFC}
\end{figure}

\begin{figure}[p]\centering 
\begin{subfigure}[t]{\linewidth}\centering 
\includegraphics[width=\linewidth]{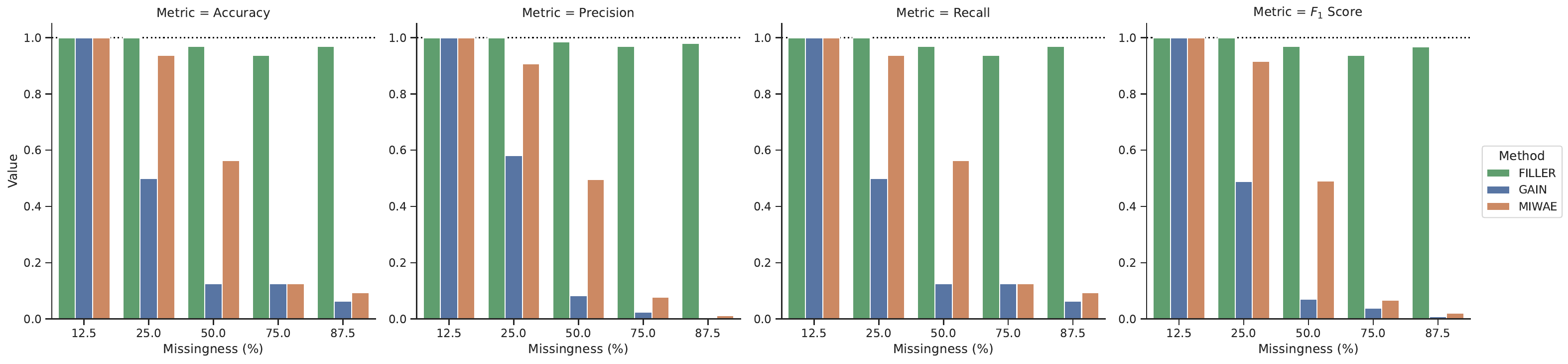}
\caption{Random Missingness}\label{fig:Faces4-random-SVC}
\end{subfigure}

\begin{subfigure}[t]{\linewidth}\centering 
\includegraphics[width=\linewidth]{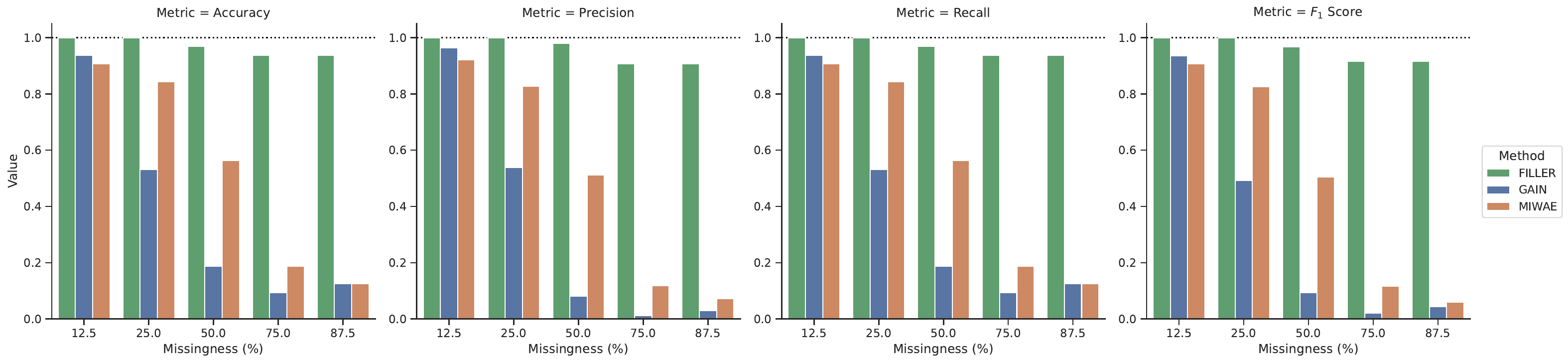}
\caption{Block-wise Missingness}\label{fig:Faces4-block-SVC}
\end{subfigure}

\begin{subfigure}[t]{\linewidth}\centering 
\includegraphics[width=\linewidth]{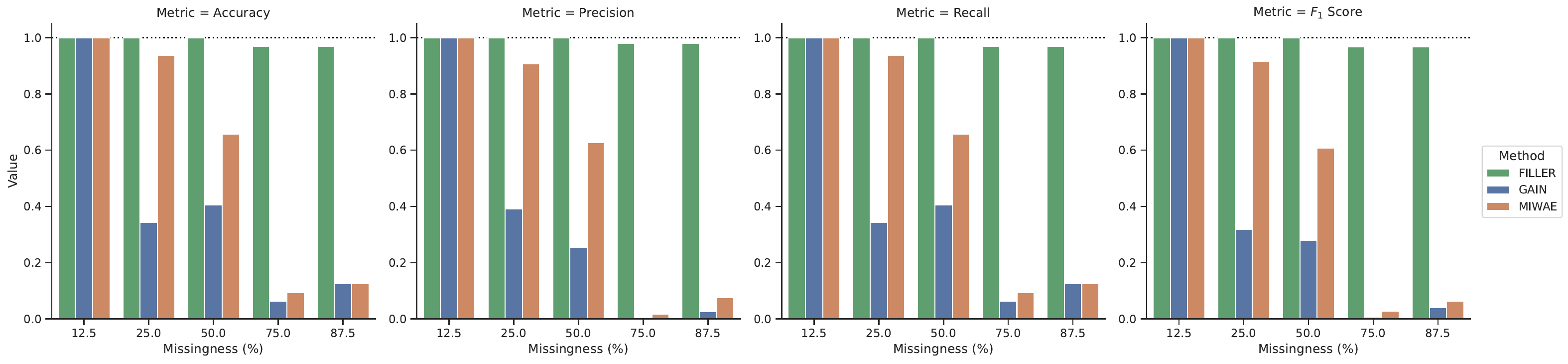}
\caption{Grid-wise Missingness}\label{fig:Faces4-grid-SVC}
\end{subfigure}
\caption{Classification results of support vector classifier on the quarter-resolution \textit{CMU Face Images} dataset with three types of missingness. The dotted lines indicate the classification performance on the original test data.}
\label{fig:Faces4-SVC}
\end{figure}


\begin{figure}[p]\centering 
\begin{subfigure}[t]{\linewidth}\centering 
\includegraphics[width=\linewidth]{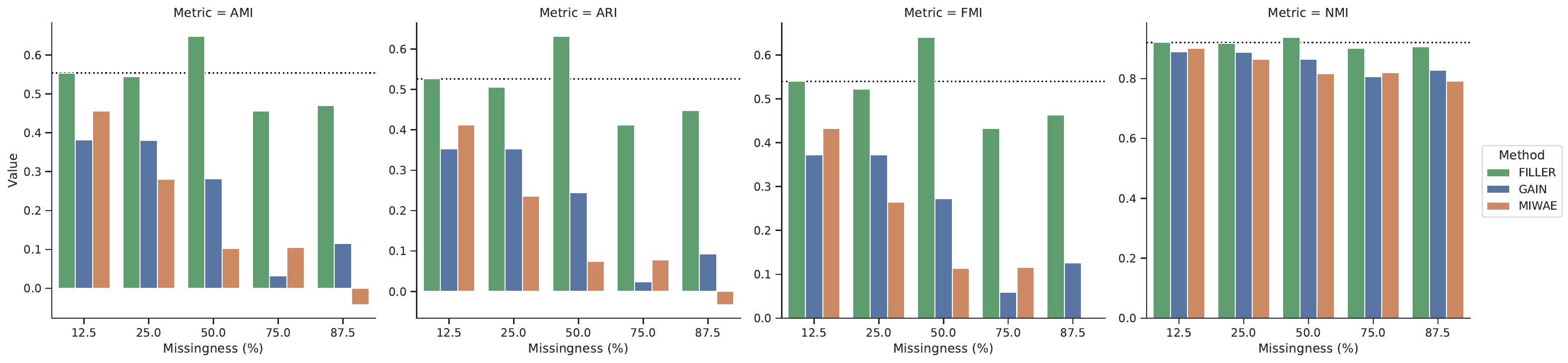}
\caption{Random Missingness}\label{fig:Faces2-random-AC}
\end{subfigure}
\begin{subfigure}[t]{\linewidth}\centering 
\includegraphics[width=\linewidth]{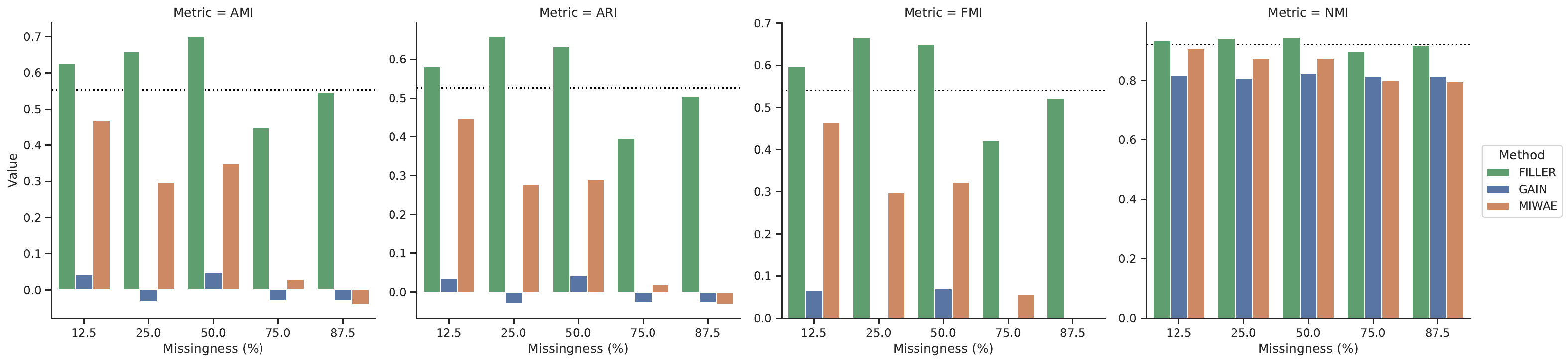}
\caption{Block-wise Missingness}\label{fig:Faces2-block-AC}
\end{subfigure}
\begin{subfigure}[t]{\linewidth}\centering 
\includegraphics[width=\linewidth]{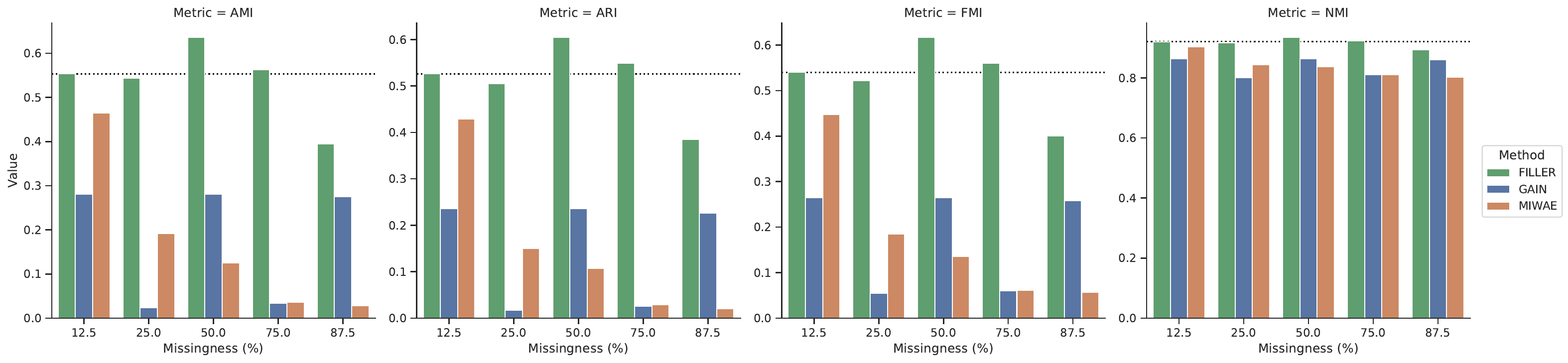}
\caption{Grid-like Missingness}\label{fig:Faces2-grid-AC}
\end{subfigure}
\caption{Clustering results of agglomerative clustering on the half-resolution \textit{CMU Face Images} dataset with three types of missingness. The dotted lines indicate the clustering performance on the original test data.}
\label{fig:Faces2-AC}
\end{figure}

\begin{figure}[p]\centering 
\begin{subfigure}[t]{\linewidth}\centering 
\includegraphics[width=\linewidth]{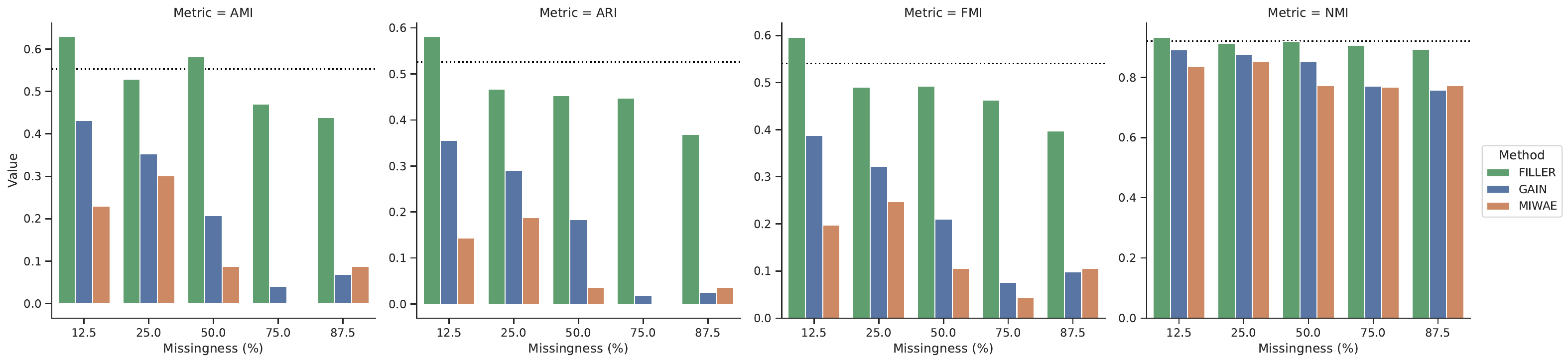}
\caption{Random Missingness}\label{fig:Faces2-random-KMC}
\end{subfigure}
\begin{subfigure}[t]{\linewidth}\centering 
\includegraphics[width=\linewidth]{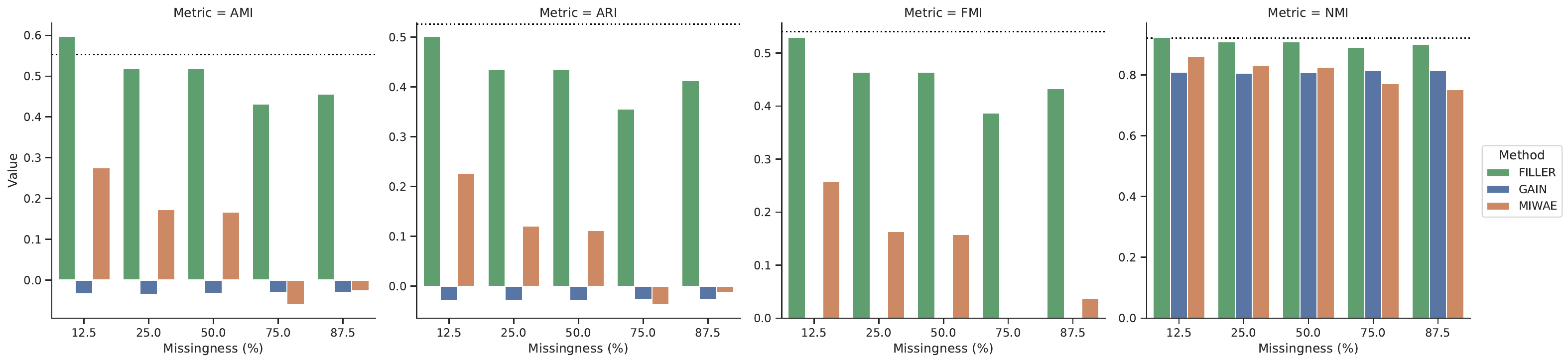}
\caption{Block-wise Missingness}\label{fig:Faces2-block-KMC}
\end{subfigure}
\begin{subfigure}[t]{\linewidth}\centering 
\includegraphics[width=\linewidth]{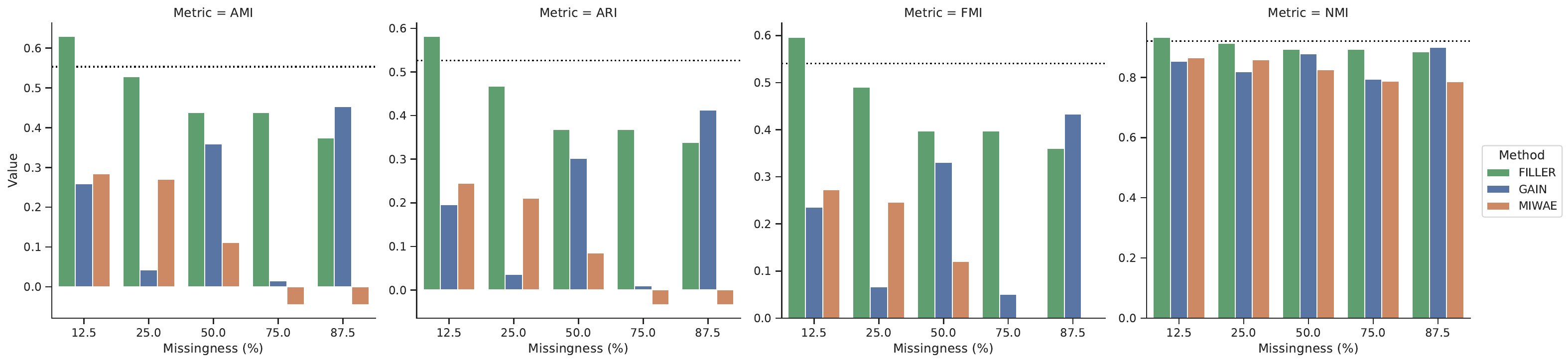}
\caption{Grid-like Missingness}\label{fig:Faces2-grid-KMC}
\end{subfigure}
\caption{Clustering results of \(k\)-means clustering on the half-resolution \textit{CMU Face Images} dataset with three types of missingness. The dotted lines indicate the clustering performance on the original test data.}
\label{fig:Faces2-KMC}
\end{figure}


\begin{figure}[p]\centering 
\begin{subfigure}[t]{\linewidth}\centering 
\includegraphics[width=\linewidth]{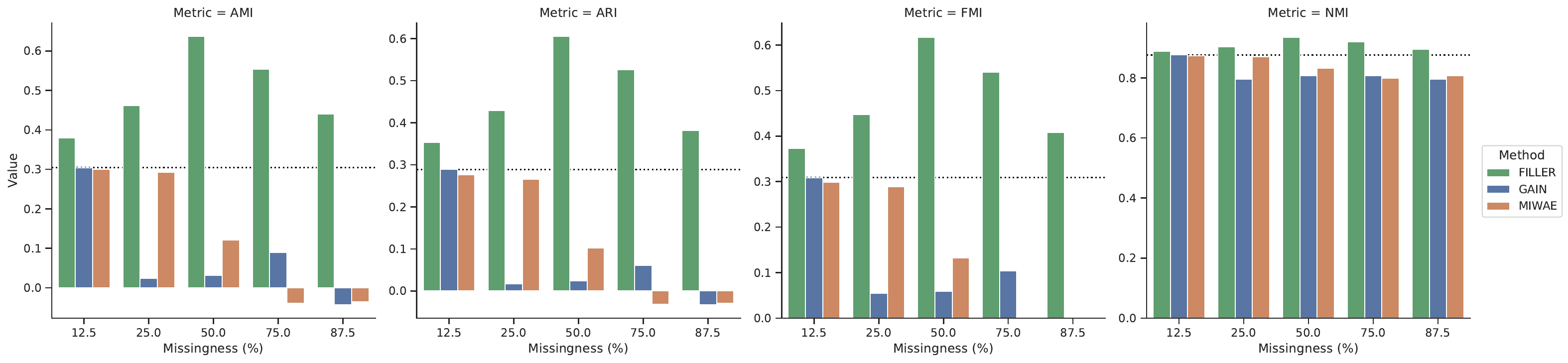}
\caption{Random Missingness}\label{fig:Faces4-random-AC}
\end{subfigure}
\begin{subfigure}[t]{\linewidth}\centering 
\includegraphics[width=\linewidth]{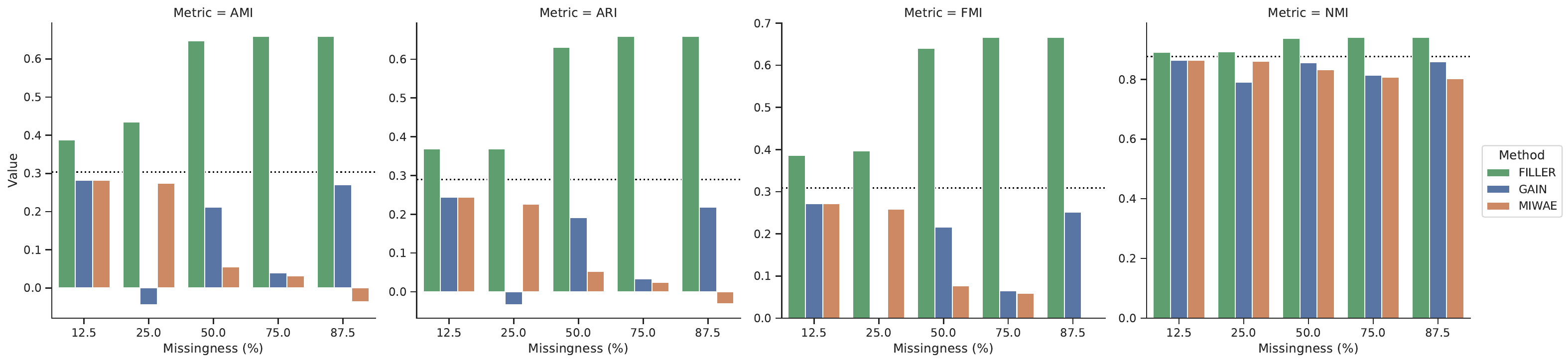}
\caption{Block-wise Missingness}\label{fig:Faces4-block-AC}
\end{subfigure}
\begin{subfigure}[t]{\linewidth}\centering 
\includegraphics[width=\linewidth]{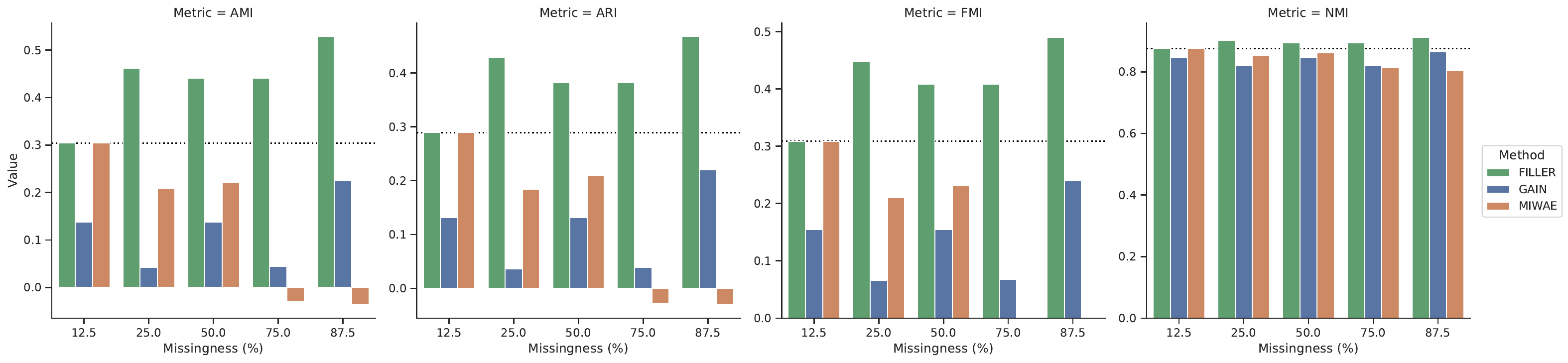}
\caption{Grid-like Missingness}\label{fig:Faces4-grid-AC}
\end{subfigure}
\caption{Clustering results of agglomerative clustering on the quarter-resolution \textit{CMU Face Images} dataset with three types of missingness. The dotted lines indicate the clustering performance on the original test data.}
\label{fig:Faces4-AC}
\end{figure}

\begin{figure}[p]\centering 
\begin{subfigure}[t]{\linewidth}\centering 
\includegraphics[width=\linewidth]{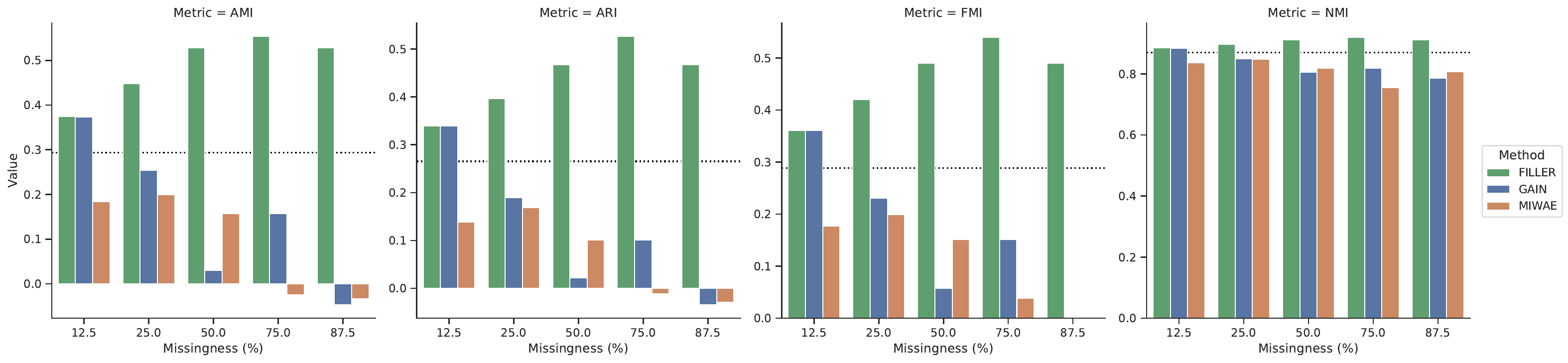}
\caption{Random Missingness}\label{fig:Faces4-random-KMC}
\end{subfigure}
\begin{subfigure}[t]{\linewidth}\centering 
\includegraphics[width=\linewidth]{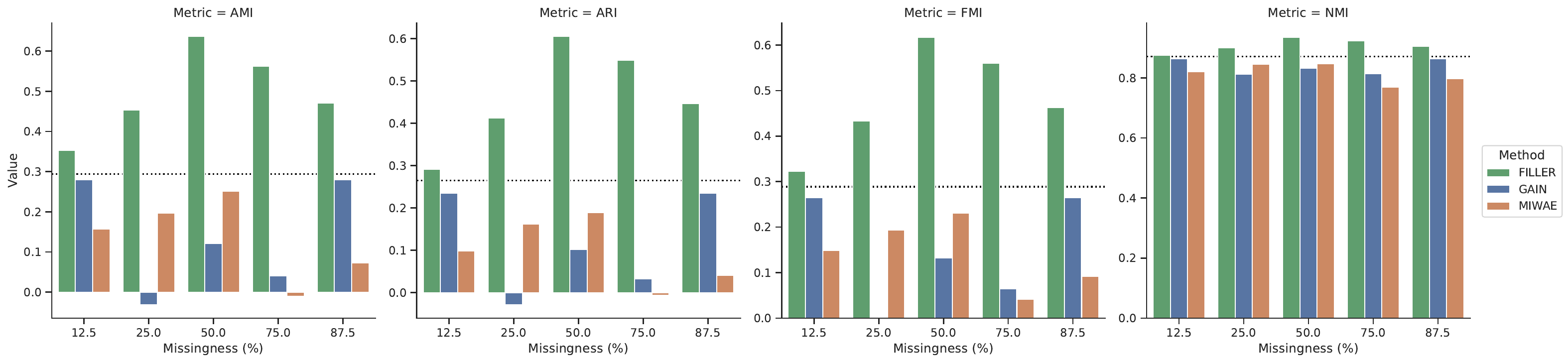}
\caption{Block-wise Missingness}\label{fig:Faces4-block-KMC}
\end{subfigure}
\begin{subfigure}[t]{\linewidth}\centering 
\includegraphics[width=\linewidth]{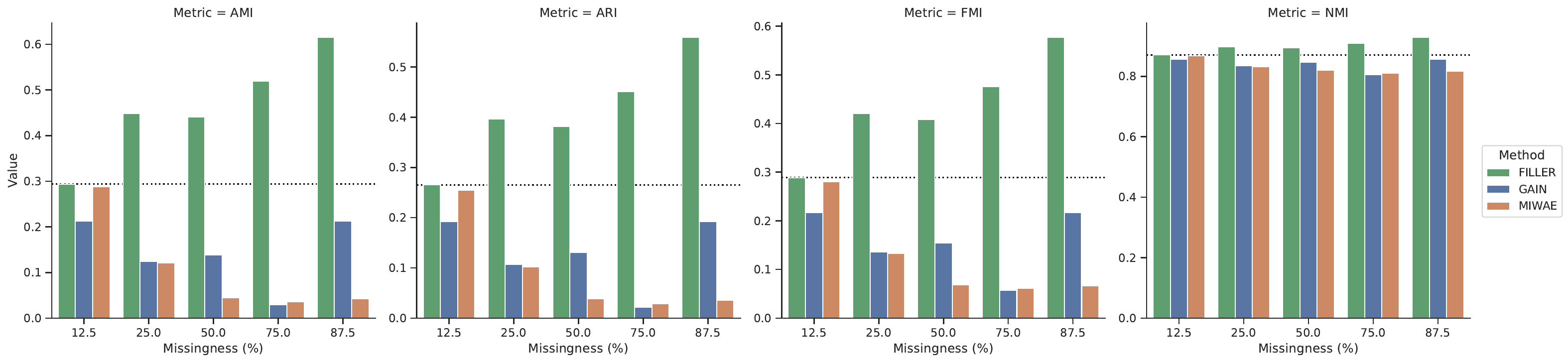}
\caption{Grid-like Missingness}\label{fig:Faces4-grid-KMC}
\end{subfigure}
\caption{Clustering results of \(k\)-means clustering on the quarter-resolution \textit{CMU Face Images} dataset with three types of missingness. The dotted lines indicate the clustering performance on the original test data.}
\label{fig:Faces4-KMC}
\end{figure}


\begin{figure}[p]\centering 
\begin{subfigure}[t]{\linewidth}\centering 
\includegraphics[width=\linewidth]{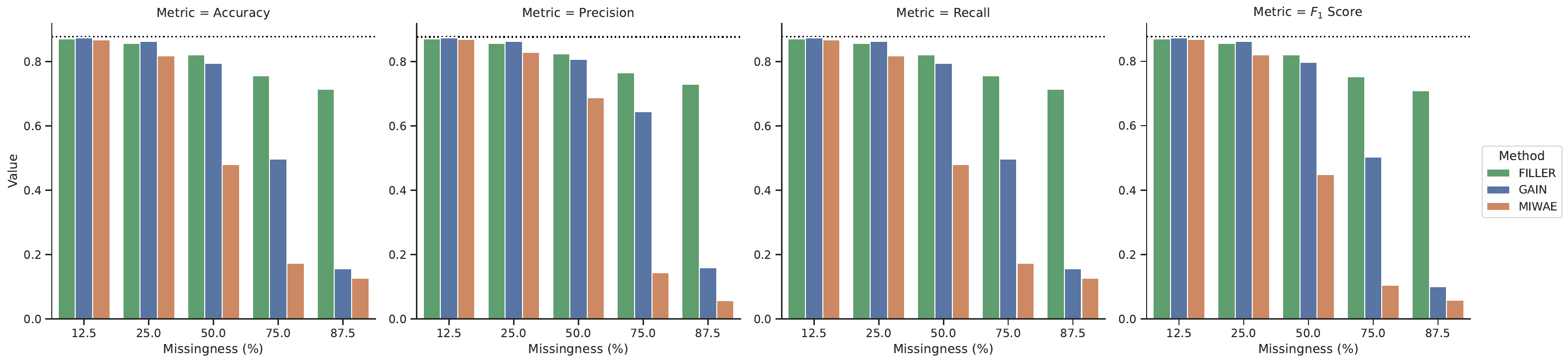}
\caption{Random Missingness}\label{fig:FMNIST-random-RFC}
\end{subfigure}

\begin{subfigure}[t]{\linewidth}\centering 
\includegraphics[width=\linewidth]{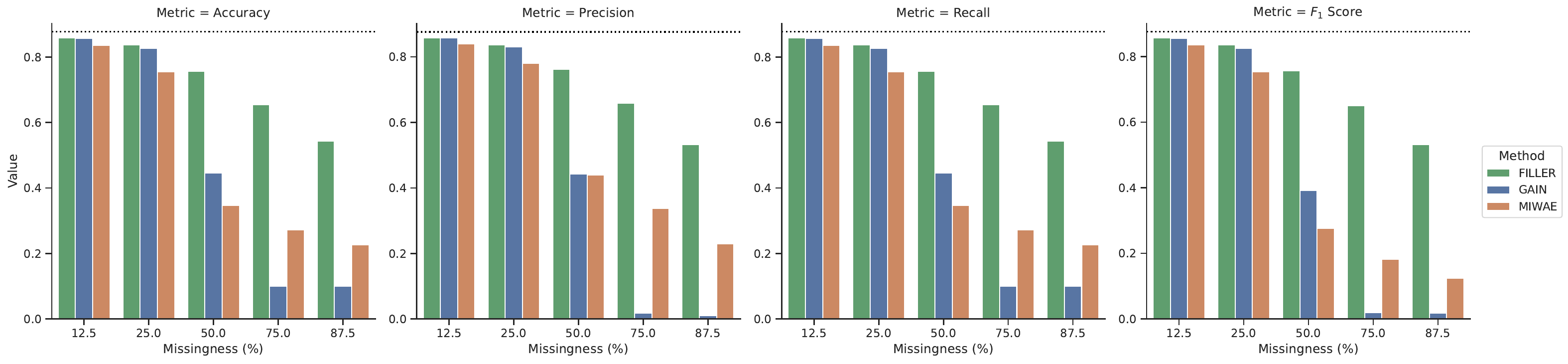}
\caption{Block-wise Missingness}\label{fig:FMNIST-block-RFC}
\end{subfigure}

\begin{subfigure}[t]{\linewidth}\centering 
\includegraphics[width=\linewidth]{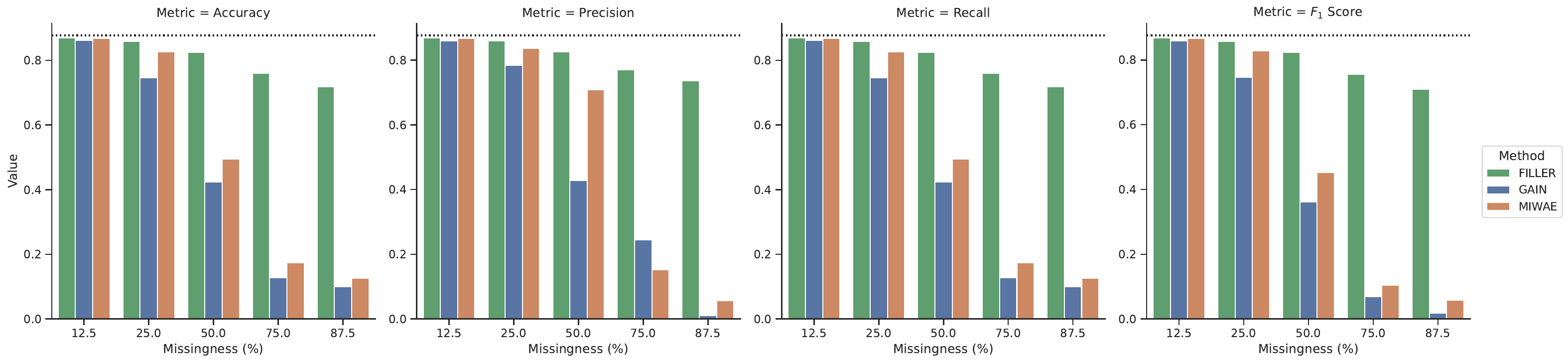}
\caption{Grid-wise Missingness}\label{fig:FMNIST-grid-RFC}
\end{subfigure}
\caption{Classification results of random forest classifier on the \textit{Fashion-MNIST} dataset with three types of missingness. The dotted lines indicate the classification performance on the original test data.}
\label{fig:FMNIST-RFC}
\end{figure}

\begin{figure}[p]\centering 
\begin{subfigure}[t]{\linewidth}\centering 
\includegraphics[width=\linewidth]{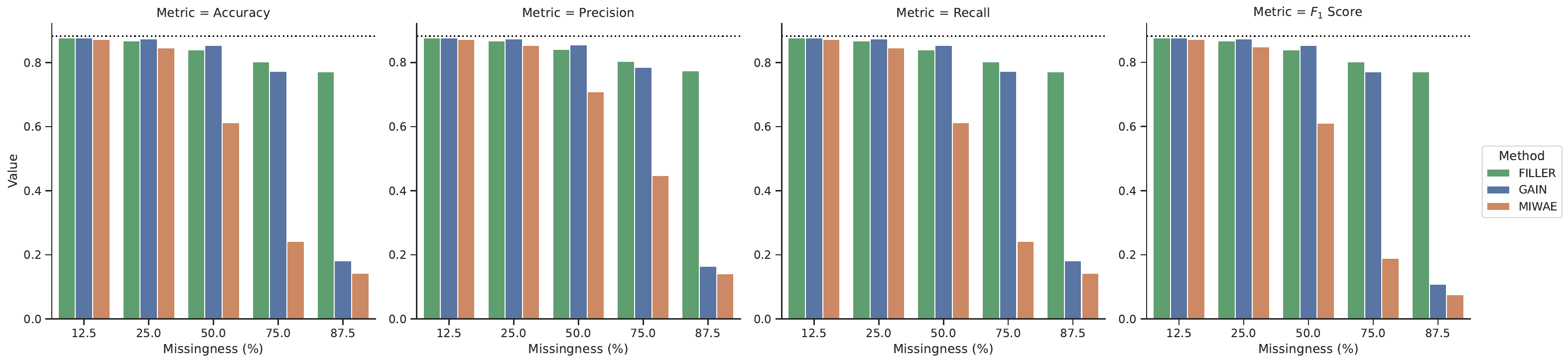}
\caption{Random Missingness}\label{fig:FMNIST-random-SVC}
\end{subfigure}

\begin{subfigure}[t]{\linewidth}\centering 
\includegraphics[width=\linewidth]{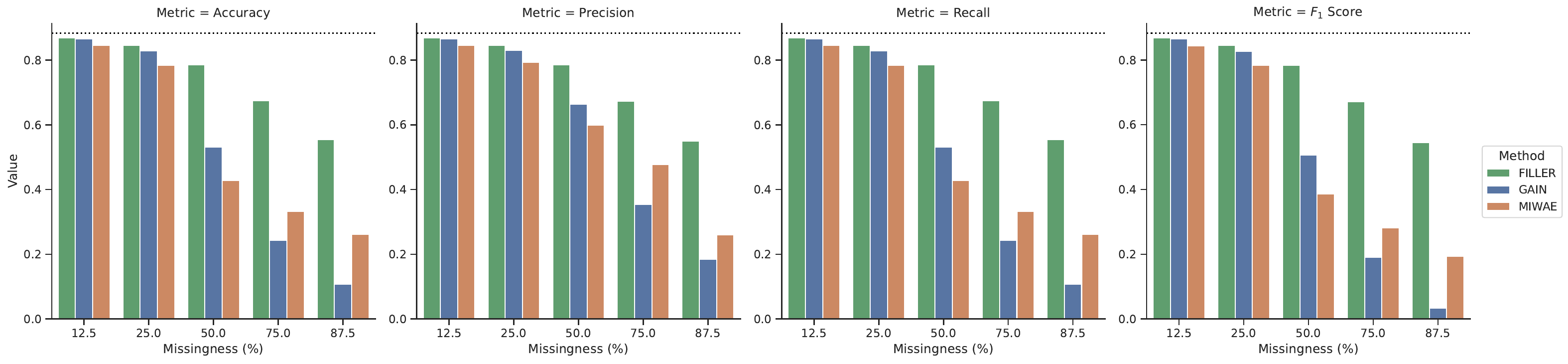}
\caption{Block-wise Missingness}\label{fig:FMNIST-block-SVC}
\end{subfigure}

\begin{subfigure}[t]{\linewidth}\centering 
\includegraphics[width=\linewidth]{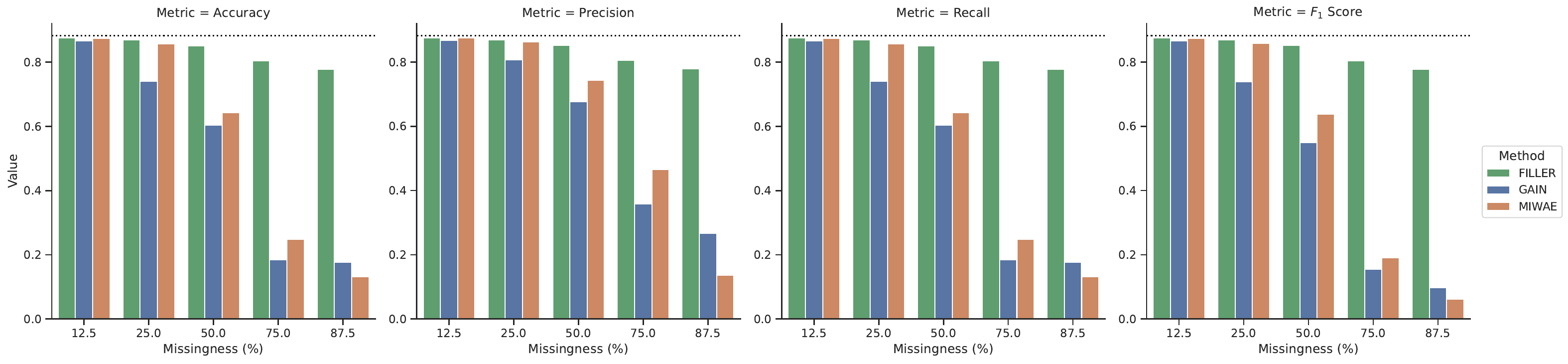}
\caption{Grid-wise Missingness}\label{fig:FMNIST-grid-SVC}
\end{subfigure}
\caption{Classification results of support vector classifier on the \textit{Fashion-MNIST} dataset with three types of missingness. The dotted lines indicate the classification performance on the original test data.}
\label{fig:FMNIST-SVC}
\end{figure}


\begin{figure}[p]\centering 
\begin{subfigure}[t]{\linewidth}\centering 
\includegraphics[width=\linewidth]{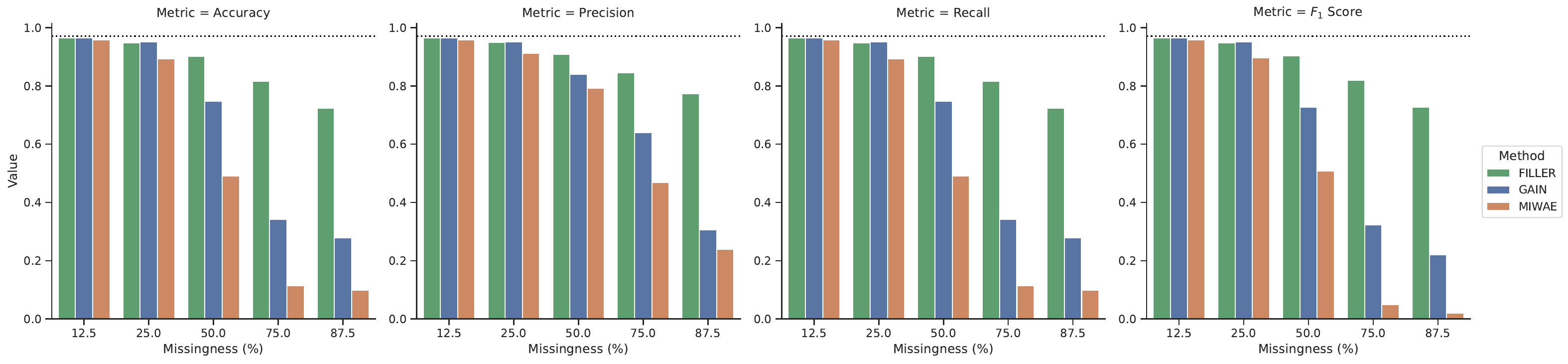}
\caption{Random Missingness}\label{fig:MNIST-random-RFC}
\end{subfigure}

\begin{subfigure}[t]{\linewidth}\centering 
\includegraphics[width=\linewidth]{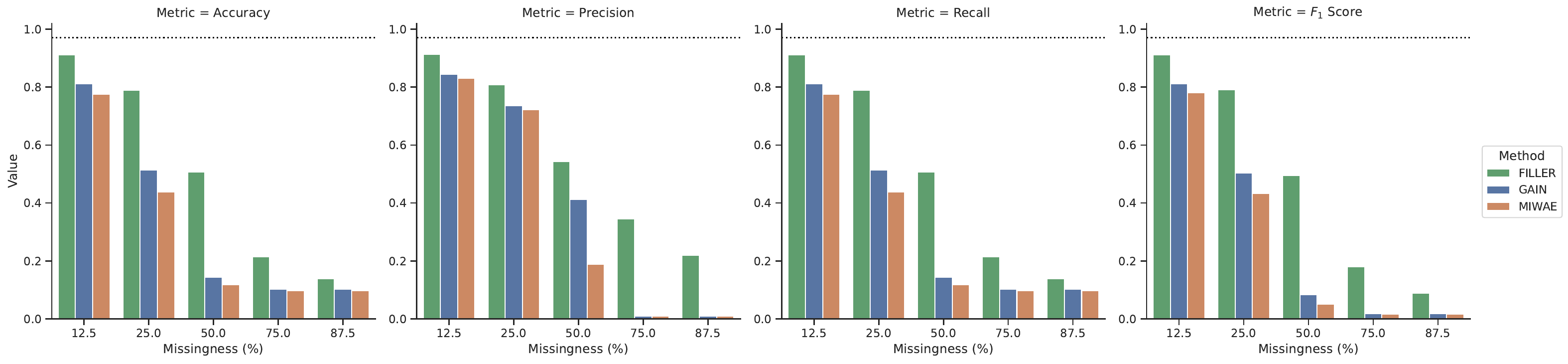}
\caption{Block-wise Missingness}\label{fig:MNIST-block-RFC}
\end{subfigure}

\begin{subfigure}[t]{\linewidth}\centering 
\includegraphics[width=\linewidth]{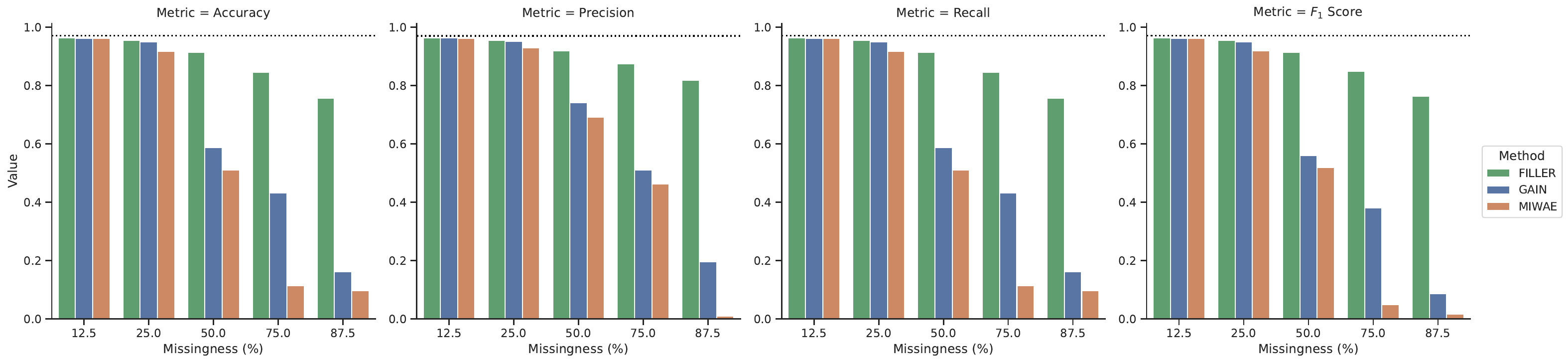}
\caption{Grid-wise Missingness}\label{fig:MNIST-grid-RFC}
\end{subfigure}
\caption{Classification results of random forest classifier on the \textit{MNIST} dataset with three types of missingness. The dotted lines indicate the classification performance on the original test data.}
\label{fig:MNIST-RFC}
\end{figure}

\begin{figure}[p]\centering 
\begin{subfigure}[t]{\linewidth}\centering 
\includegraphics[width=\linewidth]{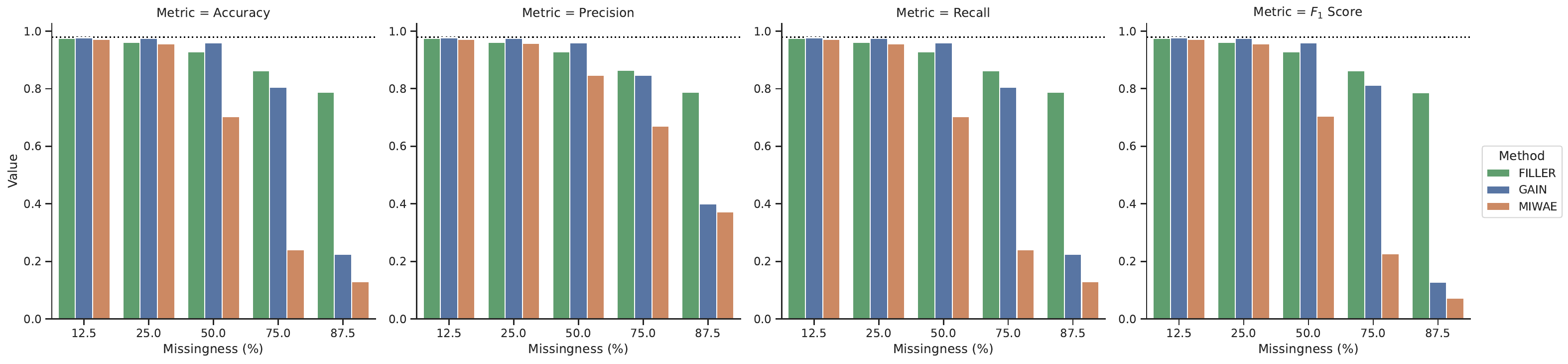}
\caption{Random Missingness}\label{fig:MNIST-random-SVC}
\end{subfigure}

\begin{subfigure}[t]{\linewidth}\centering 
\includegraphics[width=\linewidth]{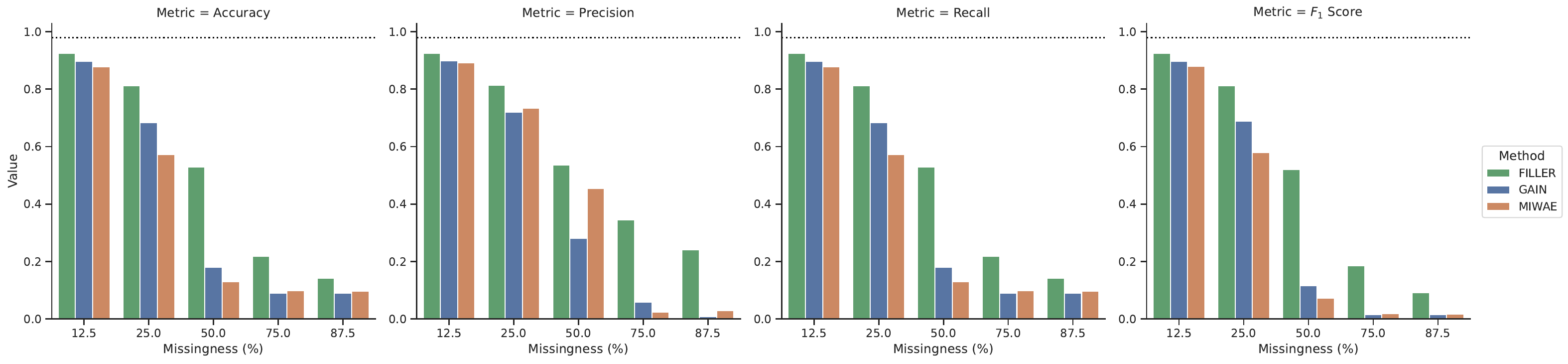}
\caption{Block-wise Missingness}\label{fig:MNIST-block-SVC}
\end{subfigure}

\begin{subfigure}[t]{\linewidth}\centering 
\includegraphics[width=\linewidth]{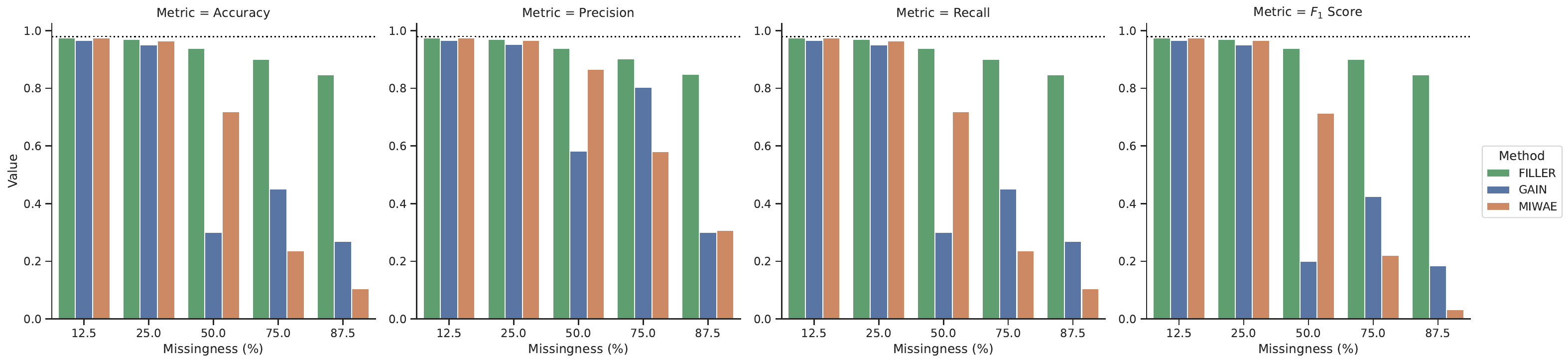}
\caption{Grid-wise Missingness}\label{fig:MNIST-grid-SVC}
\end{subfigure}
\caption{Classification results of support vector classifier on the \textit{MNIST} dataset with three types of missingness. The dotted lines indicate the classification performance on the original test data.}
\label{fig:MNIST-SVC}
\end{figure}


\begin{figure}[p]\centering 
\begin{subfigure}[t]{\linewidth}\centering 
\includegraphics[width=\linewidth]{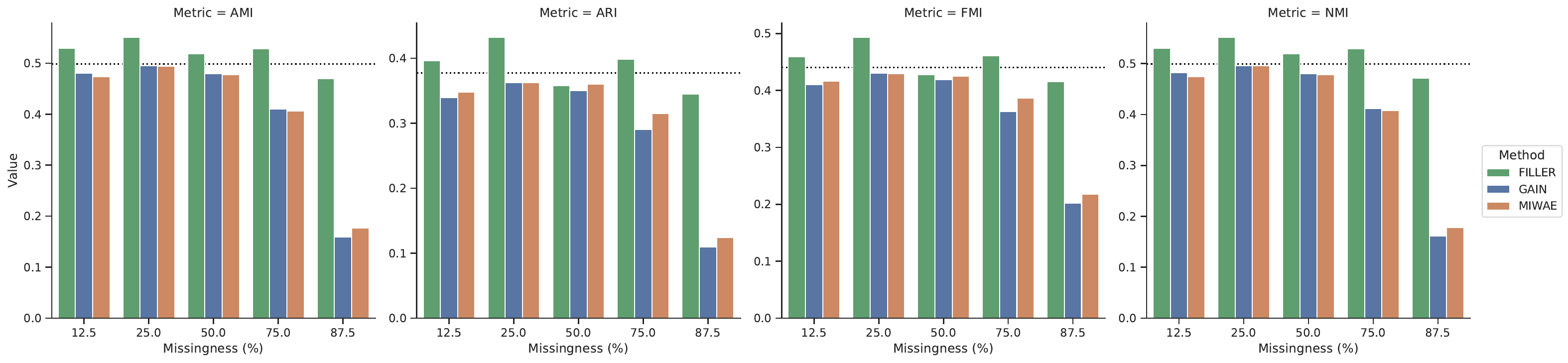}
\caption{Random Missingness}\label{fig:MNIST-random-KMC}
\end{subfigure}
\begin{subfigure}[t]{\linewidth}\centering 
\includegraphics[width=\linewidth]{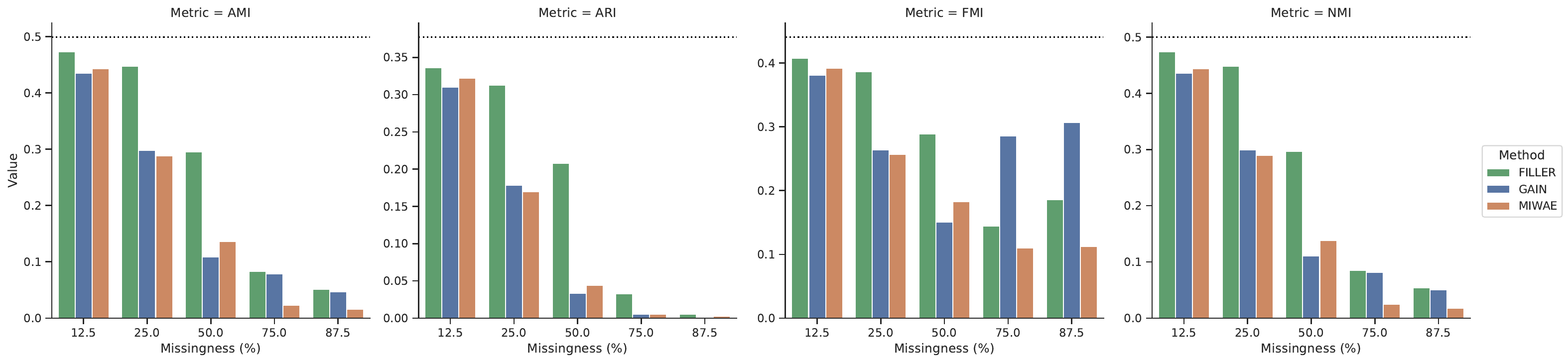}
\caption{Block-wise Missingness}\label{fig:MNIST-block-KMC}
\end{subfigure}
\begin{subfigure}[t]{\linewidth}\centering 
\includegraphics[width=\linewidth]{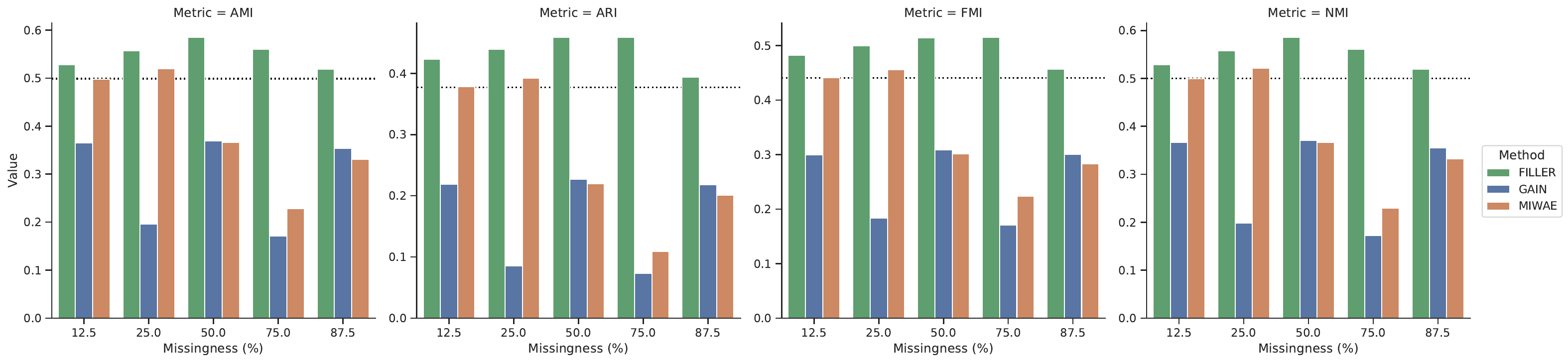}
\caption{Grid-like Missingness}\label{fig:MNIST-grid-KMC}
\end{subfigure}
\caption{Clustering results of \(k\)-means clustering on the \textit{MNIST} dataset with three types of missingness. The dotted lines indicate the clustering performance on the original test data.}
\label{fig:MNIST-KMC}
\end{figure}

\begin{figure}[p]\centering 
\begin{subfigure}[t]{\linewidth}\centering 
\includegraphics[width=\linewidth]{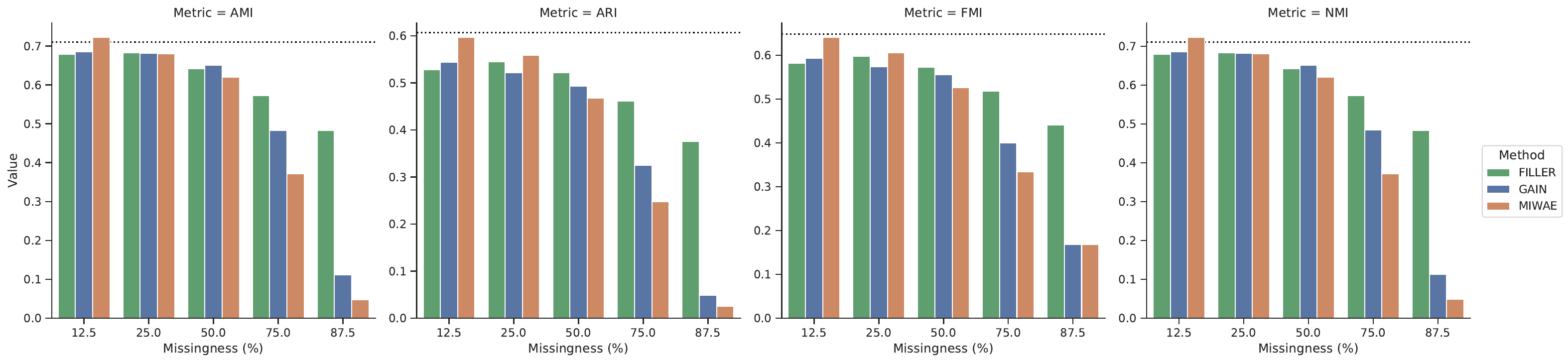}
\caption{Random Missingness}\label{fig:MNIST-random-AC}
\end{subfigure}
\begin{subfigure}[t]{\linewidth}\centering 
\includegraphics[width=\linewidth]{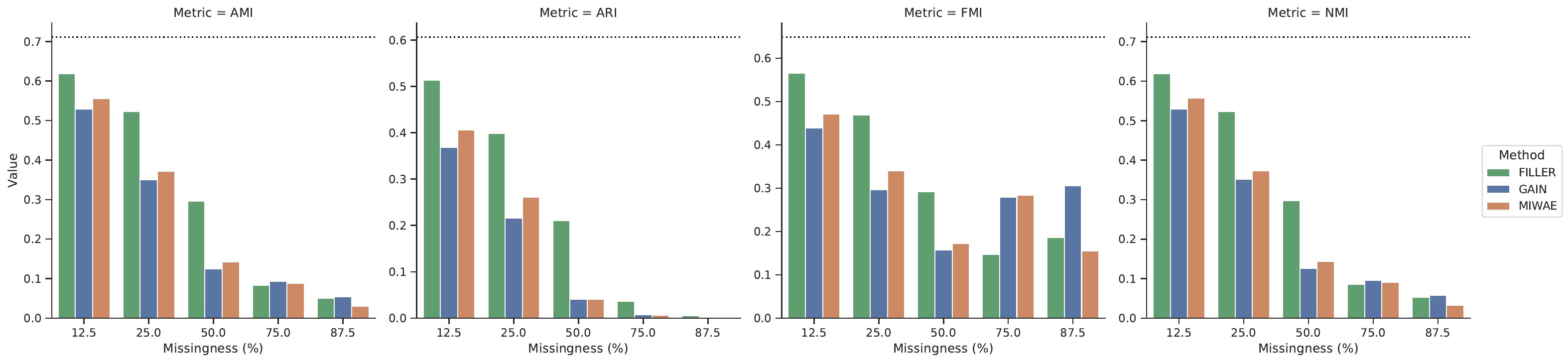}
\caption{Block-wise Missingness}\label{fig:MNIST-block-AC}
\end{subfigure}
\begin{subfigure}[t]{\linewidth}\centering 
\includegraphics[width=\linewidth]{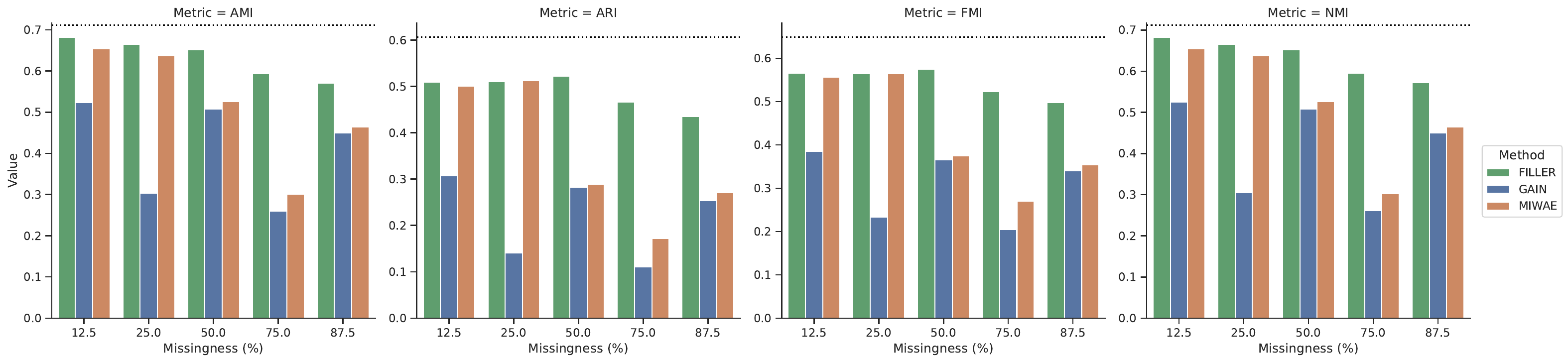}
\caption{Grid-like Missingness}\label{fig:MNIST-grid-AC}
\end{subfigure}
\caption{Clustering results of agglomerative clustering on the \textit{MNIST} dataset with three types of missingness. The dotted lines indicate the clustering performance on the original test data.}
\label{fig:MNIST-AC}
\end{figure}


\begin{figure}[p]\centering 
\begin{subfigure}[t]{\linewidth}\centering 
\includegraphics[width=\linewidth]{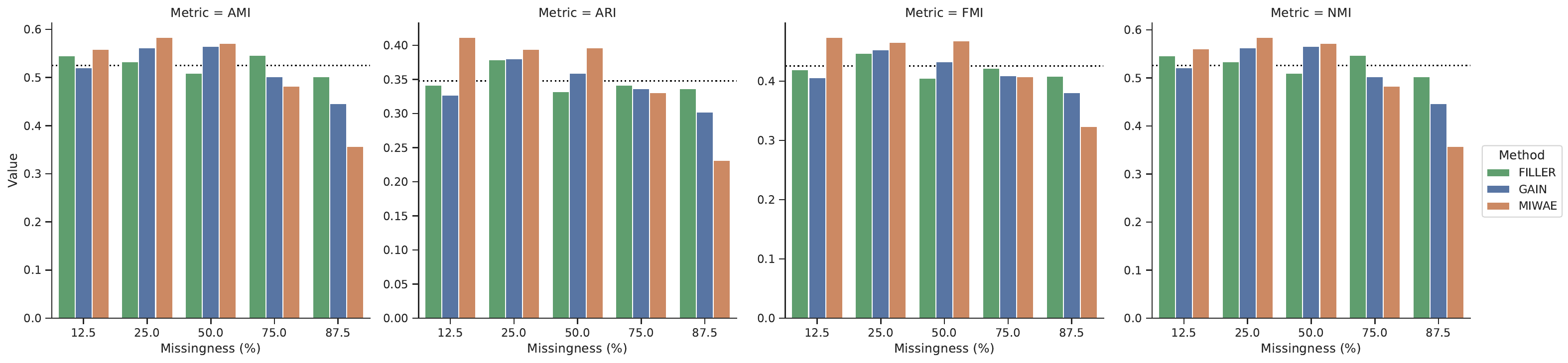}
\caption{Random Missingness}\label{fig:FMNIST-random-AC}
\end{subfigure}
\begin{subfigure}[t]{\linewidth}\centering 
\includegraphics[width=\linewidth]{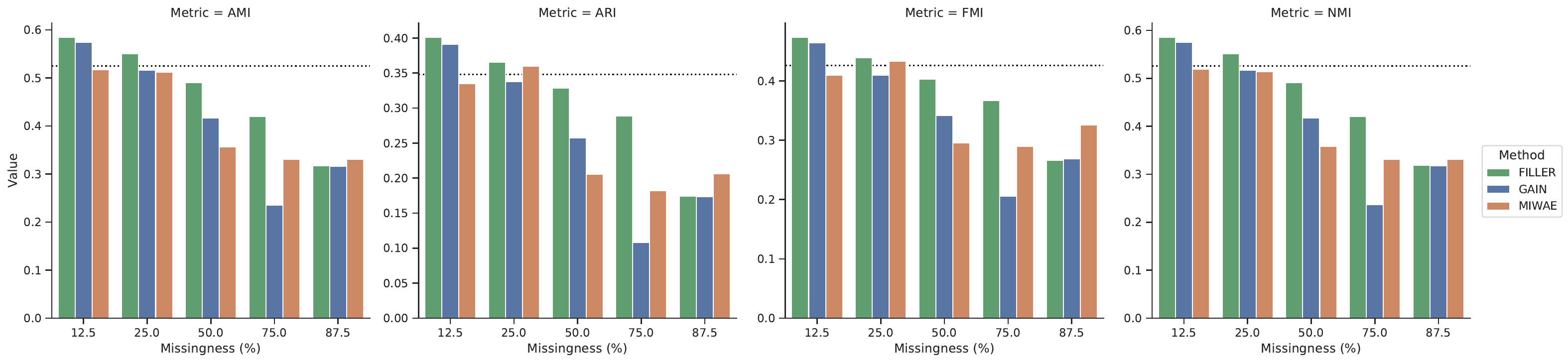}
\caption{Block-wise Missingness}\label{fig:FMNIST-block-AC}
\end{subfigure}
\begin{subfigure}[t]{\linewidth}\centering 
\includegraphics[width=\linewidth]{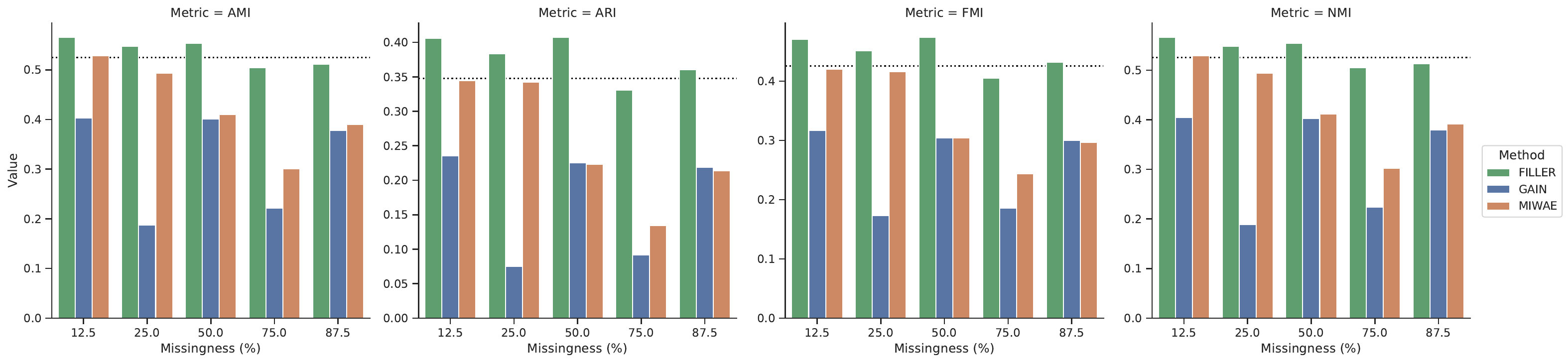}
\caption{Grid-like Missingness}\label{fig:FMNIST-grid-AC}
\end{subfigure}
\caption{Clustering results of agglomerative clustering on the \textit{Fashion-MNIST} dataset with three types of missingness. The dotted lines indicate the clustering performance on the original test data.}
\label{fig:FMNIST-AC}
\end{figure}

\begin{figure}[p]\centering 
\begin{subfigure}[t]{\linewidth}\centering 
\includegraphics[width=\linewidth]{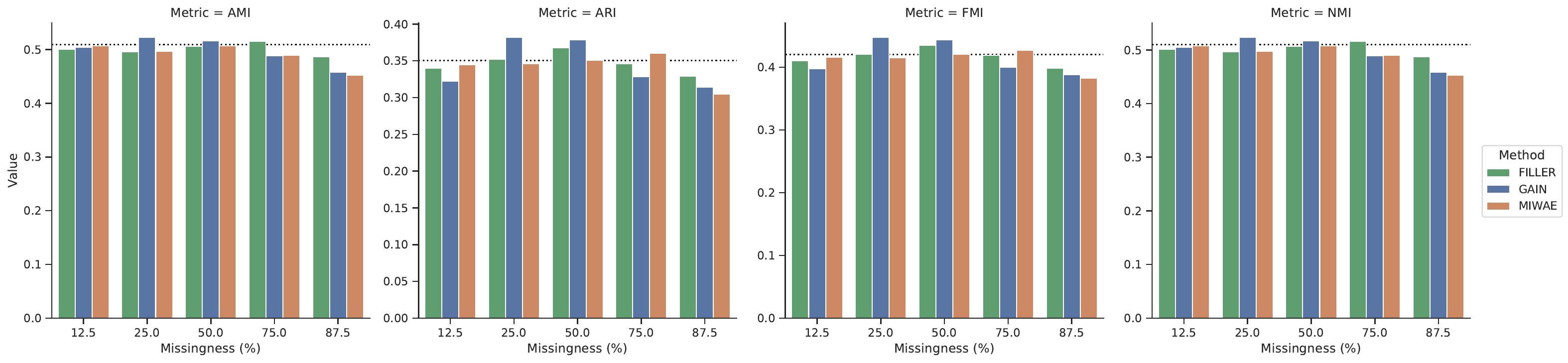}
\caption{Random Missingness}\label{fig:FMNIST-random-KMC}
\end{subfigure}
\begin{subfigure}[t]{\linewidth}\centering 
\includegraphics[width=\linewidth]{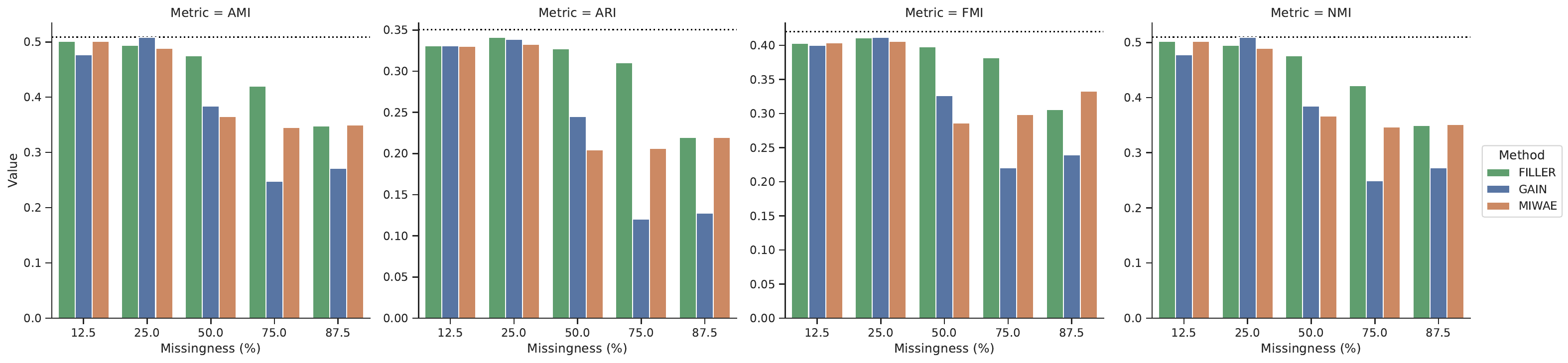}
\caption{Block-wise Missingness}\label{fig:FMNIST-block-KMC}
\end{subfigure}
\begin{subfigure}[t]{\linewidth}\centering 
\includegraphics[width=\linewidth]{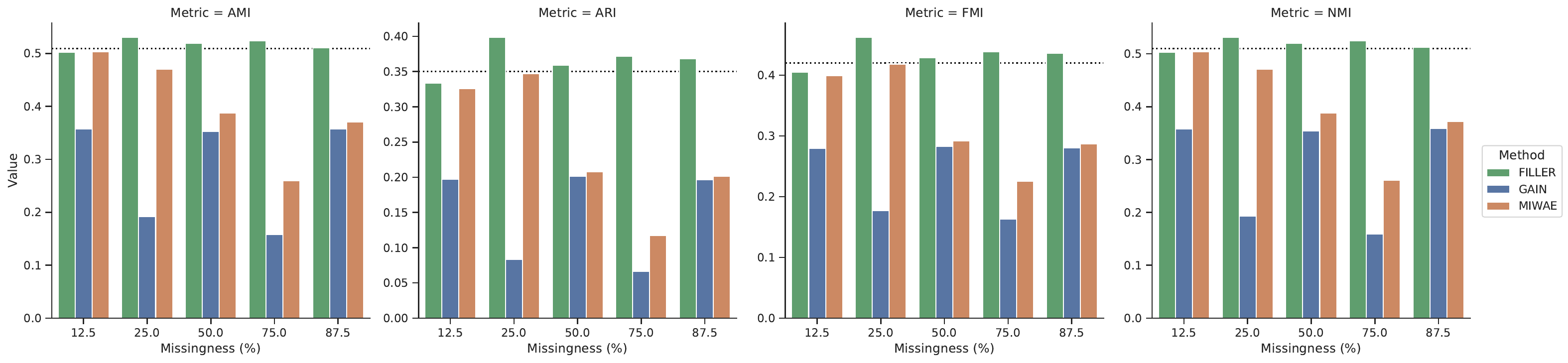}
\caption{Grid-like Missingness}\label{fig:FMNIST-grid-KMC}
\end{subfigure}
\caption{Clustering results of \(k\)-means clustering on the \textit{Fashion-MNIST} dataset with three types of missingness. The dotted lines indicate the clustering performance on the original test data.}
\label{fig:FMNIST-KMC}
\end{figure}

\end{document}